\documentclass{article} % For LaTeX2e
\usepackage{iclr2022_conference,times}
\iclrfinalcopy
\usepackage{amsmath,amsfonts,bm}

\def\eqref#1{equation~\ref{#1}}
\def\Eqref#1{Equation~\ref{#1}}
\def\1{\bm{1}}

\def\mC{{\bm{C}}}

\DeclareMathAlphabet{\mathsfit}{\encodingdefault}{\sfdefault}{m}{sl}
\SetMathAlphabet{\mathsfit}{bold}{\encodingdefault}{\sfdefault}{bx}{n}

\usepackage{hyperref}
\usepackage{url}

\usepackage[utf8]{inputenc} % allow utf-8 input
\usepackage[T1]{fontenc}    % use 8-bit T1 fonts
\usepackage{hyperref}       % hyperlinks
\usepackage{url}            % simple URL typesetting
\usepackage{booktabs}       % professional-quality tables
\usepackage{amsfonts}       % blackboard math symbols
\usepackage{amssymb}
\usepackage{amsmath}
\usepackage{nicefrac}       % compact symbols for 1/2, etc.
\usepackage{microtype}      % microtypography
\usepackage{xcolor}         % colors
\usepackage{graphicx}
\usepackage{subcaption}
\usepackage{mathrsfs}
\usepackage{microtype}
\usepackage{array}
\usepackage{lipsum}
\usepackage{multirow}
\usepackage{placeins}
\usepackage{wrapfig}
\usepackage{tikz}

\DeclareMathOperator{\EX}{\mathbb{E}}% expected value

\DeclareMathOperator*{\argmaxB}{argmax}

\newif\ifcomments
\commentstrue % Comment or set \commentsfalse to disable comments.

\newcommand{\db}{bit transition}

\newcommand{\BEL}{BEL}
\newcolumntype{L}[1]{>{\raggedright\let\newline\\\arraybackslash\hspace{0pt}}m{#1}}
\newcolumntype{C}[1]{>{\centering\let\newline\\\arraybackslash\hspace{0pt}}m{#1}}
\newcolumntype{R}[1]{>{\raggedleft\let\newline\\\arraybackslash\hspace{0pt}}m{#1}}
\newcommand{\mytitle}{Label Encoding for Regression Networks}

\title{\mytitle}

\author{Deval Shah, Zi Yu Xue \& Tor M. Aamodt  \\
Department of Electrical and Computer Engineering\\
University of British Columbia, Vancouver, BC, Canada \\
\texttt{\{devalshah,fzyxue,aamodt\}@ece.ubc.ca} \\
}

\begin{document}

\maketitle

\begin{abstract}
  Deep neural networks are used for a wide range of regression problems.
  However, there exists a significant gap in accuracy between specialized approaches and generic direct regression in which a network is trained by minimizing the squared or absolute error of output labels. 
 %solution
 Prior work has shown that solving a regression problem with a set of binary classifiers can improve accuracy by utilizing well-studied binary classification algorithms. 
  %We observe that the design space of regression by binary classification has not been fully explored.
We introduce \emph{binary-encoded labels (\BEL)}, which
  generalizes the application of binary classification to regression by providing a framework for considering 
%considers 
  arbitrary multi-bit values when encoding target values.  
  %We provide a taxonomy identifying key design aspects of this formulation.
 %%{This work introduces a novel approach using  \emph{binary-encoded labels (\BEL)} to reformulate regression as a set of binary classification sub-problems to improve accuracy.
 %%We introduce a taxonomy identifying the key design aspects of regression by binary classification.}
 % key insights
 %The proposed formulation opens up vast design space, and we demonstrate the potential of studying this design space to improve accuracy for a regression problem.
 %Of these design aspects, we primarily explore the encoding and decoding functions used for the conversion between real-valued and binary-encoded labels. 
 %Of these design aspects, we primarily explore the encoding and decoding functions used for the conversion between real-valued and binary-encoded labels and identify desirable properties of suitable encoding and decoding functions for regression. 
We identify desirable properties of suitable encoding and decoding functions used for the conversion between real-valued and binary-encoded labels {based on theoretical and empirical study}.
These properties highlight a tradeoff between classification error probability and error-correction capabilities of label encodings.
 %%%%We further derive the bounds on absolute label error in terms of binary classifiers' misclassification probability distribution for sample encoding/decoding functions, and show how and to what extent this distribution, which depends upon several factors, including the task and network architecture, impacts the suitability of encoding/decoding functions.
 %%%%We show that the error distribution of binary classifiers, which depends upon several factors, including the task and network architecture, impacts the suitability of encoding/decoding functions. \deval{this sentence does not give any new information and states the obvious}
BEL can be combined with off-the-shelf task-specific feature extractors and trained end-to-end.
%%We propose a series of sample encoding, decoding, and training loss functions for BEL and demonstrate they result in lower error than direct regression and specialized approaches while being suitable for a diverse set of regression problems, network architectures, and evaluation metrics.
We propose a series of sample encoding, decoding, and training loss functions for BEL and demonstrate they result in lower error than direct regression and specialized approaches while being suitable for a diverse set of regression problems, network architectures, and evaluation metrics. 
BEL achieves state-of-the-art accuracies for several regression benchmarks. 
%%Our analytical and empirical results suggest that the optimal encoding/decoding varies with the task, network architecture, and dataset.
%%%%In this work, we aim to introduce the design space of regression by binary classification, demonstrate that these design aspects can be explored to improve accuracy for several tasks, and provide an introductory study of suitable encoding, decoding, and loss functions to motivate further study of this design space.
 %%%We explore multiple encoding/decoding functions and evaluate them on challenging computer vision tasks.
 % using Convolutional Neural Networks (CNNs).
 Code is available at \url{https://github.com/ubc-aamodt-group/BEL_regression}. 
 %Results
  \end{abstract}

\section{Introduction}
%\blue{What is the problem \\
%What are the 2-3 most closely related SOTA - just brief selection. 2-3 closely related work- \\
%what they have not achieved?\\
% summary of what we do, how well it works, and then outline.\\
%}
{Deep regression networks, in which a continuous output is predicted for a given input, are traditionally trained by minimizing squared/absolute error of output labels, which we refer to as \emph{direct regression}.} 
However, there is a significant gap in accuracy between direct regression and recent task-specialized approaches for regression problems including head pose estimation, age estimation, and facial landmark estimation. 
Given the increasing importance of deep regression networks, developing generic approaches to improving their accuracy is desirable. 

%%%%The use of binary classification to solve ordinal regression and multiclass classification problems have been explored by prior works~\cite{ordext, ecoc}. 
{A regression problem can be posed as a set of binary classification problems. 
A similar approach has been applied to other domains such as ordinal regression~\citep{ordext} and multiclass classification~\citep{ecoc}.}
Such a formulation allows the use of well-studied binary classification approaches.
Further, new generalization bounds for ordinal regression or multiclass classification can be derived from the known generalization bounds of binary classification.  
This reduces the efforts for design, implementation, and theoretical analysis significantly~\citep{ordext}. 
{~\cite{ecoc} demonstrated that posing multiclass classification as a set of binary classification problems can increase error tolerance and improve accuracy. }
%%%%For multiclass classification, prior works have focused on the use of error-correcting output codes (ECOC) and shown superior performance to typically used one-hot encoding (i.e., output logit is $1$ for the correct class and $0$ for the rest). 
%However, in contrast to multiclass classification, not all incorrect classes are equally incorrect in regression. 
However, the proposed approaches for multiclass classification do not apply to regression due to the differences in task objective and properties of the classifiers' error probability distribution (Section~\ref{sec:related}). 
{On the other hand, prior works on ordinal regression have explored the application of binary classifiers in a more restricted way which limits its application to a wide range of complex regression problems (Section~\ref{sec:related}).
\emph{There exists a lack of a generic framework that unifies possible formulations for using binary classification to solve regression. }  
}
%%%On the other hand, prior works on ordinal regression by binary classification tackles the problem in a more restricted way which limits its application to a wide range of complex regression problems (Section~\ref{sec:related}). 

In this work, we propose \emph{binary-encoded labels (\BEL)} which improves accuracy by generalizing application of binary classification to regression. 
In BEL, a target label is quantized and converted to a binary code of length $M$, and $M$ binary classifiers are then used to learn these binary-encoded labels. 
%Each classifier's target output is ``$0$'' or ``$1$'' and it learns one or more decision boundaries over the numeric range of target labels. Here each decision boundary corresponds to a transition between $0$ and $1$ in the classifier's target output over the numeric range of target labels. 
%%%Each classifier's target output is $0$ or $1$, and it learns one or more decision boundaries for transitions between $0$ and $1$ over the numeric range of target labels. 
%%For ordinal labels in range $[1,N]$, a binary classifier's target output is ``0'' or ``1''. 
%%We define the decision boundaries for a binary classifier as the switching between ``0'' and ``1'' in the range of ordinal labels. 
An encoding function is introduced to convert the target label to a binary code, and a decoding function is introduced to decode the output of binary classifiers to a real-valued prediction. 
BEL allows using an adjustable number of binary classifiers depending upon the quantization, encoding, and decoding functions. 
BEL opens possible avenues to improve the accuracy of regression problems with a large design space spanning quantization, encoding, decoding, and loss functions. 
%BEL highlights a potentially large design space that can be explored to further improve the accuracy of deep networks for regression problems. % of a deep regression network.  

%We propose a new taxonomy for the design space of regression by binary classification and introduce three design aspects: quantization/dequantization, encoding function, and decoding function. 
We focus on the encoding and decoding functions and theoretically study the relations between the absolute error of label and binary classifiers' errors for sample encoding and decoding functions. 
%We first derive the relations between the expected absolute error of labels and classification errors for two encoding/decoding functions. 
This analysis demonstrates the impact of binary classifiers’ error distribution over the numeric range of target labels on the suitability of different encoding and decoding functions. 
Based on our analysis and empirically observed binary classifiers’ error distribution, we propose properties of suitable encoding functions for regression and explore various encoding functions on a wide range of tasks.  
%In this work, we primarily aim to introduce this design space, provide an approach for comparing different encoding and decoding functions, and \blue{demonstrate the importance} of choosing them for a given task. 
%%We also propose an expected correlation-based decoding function for regression that can effectively reduce the quantization error introduced by the encoding. 
We also propose an expected correlation-based decoding function for regression that can effectively reduce the quantization error introduced by the use of classification. 

A deep regression network consists of a feature extractor and a regressor and is trained end-to-end. 
{A regressor is typically the last fully connected layer with one output logit for direct regression. }
Our proposed regression approach (\BEL) can be combined with off-the-shelf task-specific feature extractors by increasing the regressor's output logits. % and trained end-to-end. 
Further, we find that the correlation between multiple binary classifiers' outputs can be exploited to reduce the size of the feature vector and consequently reduce the number of parameters in the regressor. 
We explore the use of different decoding functions for training loss formulation and evaluate binary cross-entropy, cross-entropy, and squared/absolute error loss functions for BEL. 
We evaluate \BEL{ }on four complex regression problems: head pose estimation, facial landmark detection, age estimation, and end-to-end autonomous driving. 
We make the following contributions in this work:
\begin{itemize}
    %%%\item We propose binary-encoded labels for regression and introduce a taxonomy for the design aspects of regression by binary classification. We theoretically study the relation between the mean absolute error and classification error for sample encoding/decoding functions. % and explore six encoding and four decoding functions.
    \item We propose binary-encoded labels for regression and introduce a general framework and a taxonomy for the design aspects of regression by binary classification. We propose desirable properties of encoding and decoding functions suitable for regression problems.  
    \item We present a series of suitable encoding, decoding, and loss functions for regression with \BEL. 
    We present an end-to-end learning approach and regression layer architecture for \BEL. We combine \BEL{ }with task-specific feature extractors for four tasks and evaluate multiple encoding, decoding, and loss functions. \BEL{ }outperforms direct regression for all the problems and specialized approaches for several tasks. % \red{achieving new state-of-the-art accuracies}. %\deval{Added this}
    \item We theoretically and empirically demonstrate the effect of different design parameters on the accuracy, how it varies across different tasks, datasets, and network architectures, and provide preliminary insights and motivation for further study.   
\end{itemize}

\section{Related Work}
\label{sec:related}
\textbf{Binary classification for regression: }
Prior works have proposed binary classification-based approaches for ordinal regression ~\citep{prank,svmo,ordext}. 
Ordinal regression is a class of supervised learning problems, where the samples are labeled by a rank that belongs to an ordinal scale. 
Ordinal regression approaches can be applied to regression by discretizing the numeric range of the real-valued labels~\citep{dorn,multilabel}. 
% and have been evaluated for depth estimation, age estimation, and historical image dating~\citep{dorn,multilabel, agecnn}. 
In the existing works on ordinal regression by binary classification, $N-1$ binary classifiers are used for target labels $ \in  \{1,2,...,N\}$, where classifier-$k$ predicts if the label is greater than $k$ or not for a given input. 
%%%%The prediction is determined by combining the outputs of $N$ binary classifiers using a reduction rule. 
%This output binary sequence is similar to the unary code of label $y$ as shown in Figure~\ref{fig:1b}. 
~\cite{ordext} provided a reduction framework and generalization bound for the same. 
%%The proposed approach has been applied to a few regression problems such as age estimation and depth estimation~\citep{agecnn, dorn}. 
However, the proposed binary classification formulation is restricted. It requires several binary classifiers if the numeric range of output is extensive, whereas reducing the number of classifiers by using fewer quantization levels increases quantization error. 
%Thus, a more generalized approach for using binary classification for regression is desirable to allow flexible design of output classifiers. 
Thus, a more generalized approach for using binary classification for regression is desirable to allow flexibility in the design of classifiers.
\\
\textbf{Binary classification for multiclass classification: } 
%%Previous works have proposed binary classification and introduced a general framework for multiclass classification. 
~\cite{ecoc} proposed the use of error-correcting output codes (ECOC) to convert a multiclass classification to a set of binary classification problems.
%, where each label is converted to a unique binary string.  
This improves accuracy as it introduces tolerance to binary classifiers' errors depending upon the hamming distance (i.e., number of bits changed between two binary strings) between two codes. 
%%%%They proposed desirable properties of the codes and showed that the specific assignment of codes to labels is not important for multiclass classification. 
~\cite{multiclass} provided a unifying framework and multiclass loss bounds in terms of binary classification loss. 
More recent works have also used Hadamard code, a widely used error-correcting code~\citep{Song2021,Verma2019}. 
%%Other works have focused on the problem of extreme classification and the use of ECOC and compact codes (sublinear increase in the length of codes with the number of classes) for a classification problem with a large number of classes~\citep{extremecode,extremegraph}.
{Other works have focused on the use and design of compact codes that exhibit a sublinear increase in the length of codes with the number of classes for extreme classification problems with a large number of classes~\citep{extremecode,extremegraph}. } 
%%%%\blue{Regression problems can be solved by classification. However, the multiclass classification loss does not use the ordinal information contained in the labels.}
However, the proposed encoding and decoding approaches do not consider the task objective and labels' ordinality for regression. 
Further, the binary classifiers possess distinct error probability distribution properties for regression problems as observed empirically (Section~\ref{sec:err}), which can be exploited to design codes suitable for regression. 

Multiclass classification and ordinal regression by binary classification can be viewed as special cases falling under the BEL framework. As shown in Section~\ref{sec:eval}, other BEL designs yield improvements in accuracy over these approaches.  Task-specific regression techniques are well explored as summarized below (see also Appendix~\ref{sec:supexp}). While effective, task-specific approaches lack generality by design.

%Our goal is uncovering generic approaches to improve regression accuracy,
%Note that most of these approaches focus on specific tasks, whereas we aim to introduce a generic framework for regression problems. We compare with existing specialized regression approaches for several benchmarks in Section~\ref{sec:eval}.  \deval{we can move problem-specific related work to the Supplemental material and just keep this paragraph. }

\textbf{Head pose estimation: } 
%%Other works focus on the regressor architecture or loss formulation. 
SSR-Net~\citep{ssrnet} and FSA-Net~\citep{fsanet} used a soft stagewise regression approach.  
HopeNet~\citep{hopenet} used a combination of classification and regression loss. 
~\cite{quatnet} used a combination of regression and ordinal regression loss. 
\\
\textbf{Facial landmark detection: }
%%Facial landmark detection is an extensively studied problem used for facial analysis and modeling. 
%%Most approaches for facial landmark detection can be divided into direct regression or heatmap-based approaches. 
%~\cite{hrnetface} proposed HRNet architecture for feature extraction and directly minimized the L2 loss between predicted and target 2D heatmap, which is generated by placing a Gaussian distribution with fixed small variance around the ground truth landmark location. 
~\cite{hrnetface} minimize L2 loss between predicted and target 2D heatmaps with the latter formed using small variance Gaussians centered on ground truth landmarks. 
AWing~\citep{awing} modified loss for different pixels in the heatmap. 
%%AnchorFace~\citep{anchorface} demonstrates that anchoring facial landmarks on templates improves regression performance.  
%%LAB~\citep{lab} exploits extra boundary information to improve the accuracy. 
LUVLi~\citep{luvli} proposed a landmark's location, uncertainty, and visibility likelihood-based loss. 
~\cite{binaryheatmap} used binary heatmaps with pixel-wise binary cross-entropy loss. 
%%%%The proposed approaches are specific to facial landmark detection, whereas we aim to propose a generic regression framework that can be applied to a wide range of problems. 
\\
\textbf{Age estimation: }
%%%Existing approaches for age estimation include ordinal regression~\cite{agecnn, coralcnn}, soft regression~\cite{ssrnet}, and expected value ordinal regression ~\cite{mvloss, dldl}. 
OR-CNN~\citep{agecnn} and CORAL-CNN~\citep{coralcnn} used ordinal regression via binary classification. 
%CORAL-CNN~\citep{coralcnn} refined this approach by enforcing the ordinality of the model output. 
%OR-CNN~\citep{agecnn} used ordinal regression via binary classification. 
%CORAL-CNN~\citep{coralcnn} refined this approach by enforcing the ordinality of the model output. 
MV-Loss~\citep{mvloss} proposed to penalize the model output based on the age distribution's variance, while ~\cite{dldl} proposed to use the KL-divergence between the softmax output and a generated label distribution for training. 
\iffalse
\textbf{End-to-end autonomous driving: }
The autonomous driving model's task is to predict the future driving angle based on a forward-facing image from the vehicle's perspective. 
PilotNet~\cite{pilotnet} used a small network trained by minimizing the mean squared error for end-to-end autonomous driving. 
\fi

\section{Binary-encoded Labels for Regression  (\BEL)}
\label{proposed}
\begin{figure}[t]
  \centering
 % \begin{subfigure}[t]{0.85\textwidth}
 %   \centering
      \includegraphics[width=0.85\textwidth]{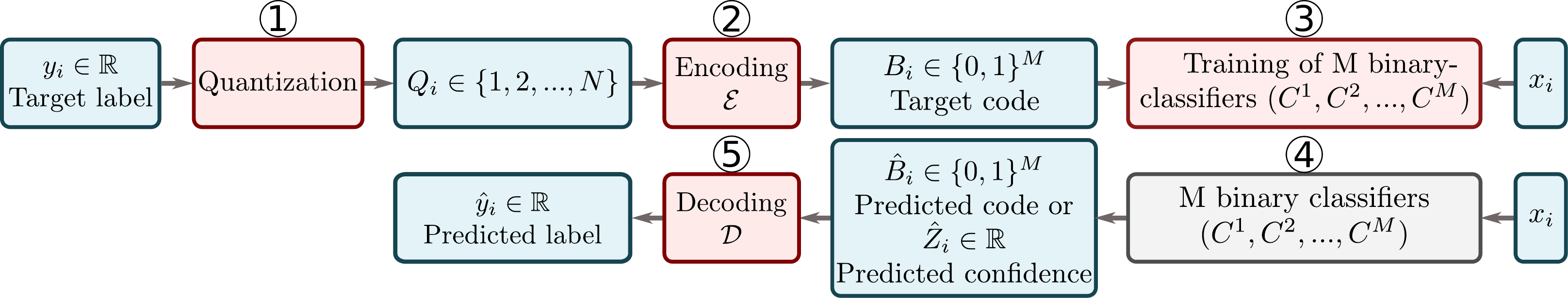}
 %  \caption{The training (top) and inference (bottom) flow}
 %  \label{fig:p1}
 % \end{subfigure} 
 % \hfill
 % \begin{subfigure}[t]{0.7\textwidth}
 %   \centering
 %     \includegraphics[width=\textwidth]{figures/ed_examples.pdf}
 %  \caption{Encoding and decoding examples}
 %  \label{fig:p2}
 % \end{subfigure} 
 \caption{The training (top) and inference (bottom) flow of binary-encoded labels (BEL) for regression networks. Red colored blocks represent design aspects we focus on. }
 % (b) Examples of two lookup table-based encoding and decoding functions. $\hat{Q}_i$ represents the decoded quantized level.  }
 \label{fig:p1}
\end{figure} 
We consider regression problems where the goal is to minimize the error between real-valued target labels $y_i$ and predicted labels $\hat{y}_i$, over a set of training samples $i$.
We transform this problem to a set of binary classification sub-problems by converting a real-valued label to a binary code. 
%In the rest of the paper, unless otherwise stated, a classifier refers to a binary classifier. 

Figure~\ref{fig:p1} shows the training and inference flow for BEL. The red-colored blocks highlight functions that vary under BEL.  
A real-valued label $y_i \in \mathbb{R}$ is quantized to level $Q_i \in \{1,2,...,N\}$~{\textcircled{\small 1}}. 
The quantized label is converted to a binary vector $B_i \in \{0,1\}^M$, that we call a {\em binary-encoded label}, using encoding function $\mathcal{E}$~{\textcircled{\small 2}}. 
There are $\binom{2^M}{N}$ possible encoding functions---a large number.
The binary-encoded labels $B_i$ are used to train $M$ classifiers~{\textcircled{\small 3}}. 
During inference the $M$~classifiers predict a binary code $\hat{B}_i \in \{0,1\}^M$ for input $x_i$~{\textcircled{\small 4}}. 
The predicted code ($\hat{B}_i$) or the predictions' magnitude ($\hat{Z}_i$), which indicates its confidence, is then decoded to a predicted label $\hat{y}_i \in \mathbb{R}$ using a decoding function $\mathcal{D}$~{\textcircled{\small 5}}. 
%The predicted code ($\hat{B}_i$) or the real-valued confidence vector of classifiers' prediction ($\hat{Z}_i$) is then decoded to a predicted label $\hat{y}_i \in \mathbb{R}$ using a decoding function $\mathcal{D}$~{\textcircled{\small 5}}. 
We explore decoding functions that yield either quantized or continuous predicted outputs.
The latter avoids quantization error by employing expected correlation (Section~\ref{sec:dec}).  

BEL contains five major design parameters resulting in a large design space:
quantization, encoding, decoding, regressor network architecture, and training loss formulation.
%%%%Figure~\ref{fig:p2} illustrates how encoding and decoding functions $\{\mathcal{E}, \mathcal{D}\}$ can impact accuracy.
%%%%The figure shows two distinct lookup table based encoding/decoding functions $\{\mathcal{E}_1,\mathcal{D}_1\}$ and $\{\mathcal{E}_2,\mathcal{D}_2\}$. % (gray code~\cite{gray}). 
%%%%Case~1 and 2 illustrate examples of encoding a given target value and decoding an erroneous predicted code (italicized red bits indicate misclassification). 
%%%%For a given classification error, $\{\mathcal{E}_2,\mathcal{D}_2\}$ exhibits lower regression error versus $\{\mathcal{E}_1,\mathcal{D}_1\}$. 
In this work we consider only uniform quantization while leaving nonuniform quantization~\citep{dorn} to future work.
Section~\ref{sec:cm} and~\ref{sec:dec} explore the characteristics of suitable encoding, decoding, and loss functions. 
%%The suitability of encoding and decoding functions significantly depends upon the classification error, and hence the task, as we show in Section~\ref{sec:err}.  
%As we show in Section~\ref{sec:err}, classification error has a significant impact on the suitability of encoding and decoding functions for a given task. 
Section~\ref{sec:proposed} explores the impact of regressor network architecture. 
%%Section~\ref{sec:proposed} explore the impact of regression network architecture and training loss for given encoding/decoding functions.
%%%We find varying any of these aspects can improve accuracy.  While BEL provides a framework and some codes appear generally better than others, the most suitable BEL code to employ appears to vary across task, dataset, and network architecture, as we show both theoretically (Section~\ref{sec:err}) and empirically (Section~\ref{sec:eval}). 
We find varying any of these aspects can improve accuracy.  While BEL provides a framework, and some design choices appear generally better than others, the most suitable BEL parameters to employ vary across task, dataset, and network architecture, as we show both theoretically (Section~\ref{sec:err}) and empirically (Section~\ref{sec:eval}). 
%%%%This work mainly focuses on and explores the encoding and decoding ($\{\mathcal{E},\mathcal{D}\}$) functions. We also perform a limited exploration of the training loss function.    
\subsection{Analysis of Encoding/Decoding Functions} 
\label{sec:err}
This section analyzes the potential impact of encoding/decoding functions on regression error assuming empirically observed error distributions for the underlying classifiers. % for \BEL. 
We compare Unary and Johnson codes (Figure~\ref{fig:cu} and~\ref{fig:cj}) to determine when each is preferable.
With this analytical study, we aim to obtain insight into ordinal label classifier impact on regression error when employing simple encoding and decoding functions $\{\mathcal{E},\mathcal{D}\}$.
Based upon this analysis we identify desirable properties for these functions. % for regression and empirical results. 
The design of the codes and intuition for trying them are discussed in Section~\ref{sec:cm}. 
%The encoding and decoding functions are important design aspects introduced by BEL. 
We divide our analysis into three parts: 
First, the expected error of predicted labels is derived in terms of classifiers' errors for two $\{\mathcal{E},\mathcal{D}\}$ functions. 
Next, we propose an approximate classifier's error probability distribution over the numeric range of target labels for regression based on empirical study. 
Last, we compare the expected error of sample $\{\mathcal{E},\mathcal{D}\}$ functions based on our analysis. 
We use labels $  y_i \in [1,N-1] $, with quantization levels $Q_i \in \{1,2,...,N-1\}$. Quantization error is not included as it is not affected by $\{\mathcal{E},\mathcal{D}\}$ functions. 
%%%We compare the impact of a unary (Figure~\ref{fig:cu}) and a Johnson code (Figure~\ref{fig:cj}) (further described in Section~\ref{sec:cm}). 
\paragraph{Expected absolute error bounds in terms of classification error:} 
%%First, we analyze the expected absolute error given the classification error for a unary code, BEL-U. 
First, we analyze the unary code (BEL-U). 
%The same steps can be used for other encoding/decoding functions.  
The encoding function $\mathcal{E}^{\text{BEL-U}}$ converts $Q_i$ to $B_i =   b_i^1, b_i^2,...,b_i^{N-2}$, where $b_i^k = 1$ for $k< Q_i$, else $0$. 
In this case, a good choice of decoding function turns out to be simply counting the number of 1 outputs across all $N-2$ classifiers since
a error in a single classifier changes the prediction by only one quantization level. 
%%A single-bit error changes the predicted value by one quantization level for a decoding function that counts the number in the binary code.  
%So the decoding function is defined as:
Adding one since $Q_i=1$ is encoded by all zeros gives:
\begin{equation}
  \label{eq:8}
  \mathcal{D}^{\text{BEL-U}}(\hat{b}_i^{1},\hat{b}_i^2,...,\hat{b}_i^{N-2}) =  \sum_{k=1}^{N-2} \hat{b}_i^k + 1
\end{equation}
%\begin{equation}
%\label{eq:6}
%\mathcal{D}^{\text{BEL-J}}(\hat{b}_i^1,\hat{b}_i^2,...,\hat{b}_i^{\frac{N}{2}}) =  -  \max_{k \in \{1...frac{N}{2}\}} k  \hat{b}_i^k + \max_{k \in \{1...\frac{N}{2}\}} \Big( \frac{N}{2} - k + 1\Big) \hat{b}_i^k  + \frac{N}{2}
%\end{equation}

Let $e_{k}(n)$ be the error probability of classifier $k$ for target quantized label $Q_i=n$. 
For a uniform distribution of $y_i$ in the range $[1,N-1]$ the expected error for BEL-U can be shown (see Appendix~\ref{sec:a1}) to be bounded as follows: %encoding/decoding can be shown to be:
\begin{equation}
  \label{eq:10}
  \EX(| \hat{y}^{\text{BEL-U}} -  y |) \leqslant \frac{1}{N-1} \sum_{n=1}^{N-1} \Big( \sum_{k=1}^{N-2} e_{k}(n) \Big)
\end{equation}

%Next, we derive the relation between the expected error of label and the error probability distribution of binary classifiers for BEL-J. 
%Equation~\ref{eq:10} is used to find the bound on the expected absolute error from the error probability distributions of classifiers for $\mathcal{E}^{\text{BEL-U}}$/$\mathcal{D}^{\text{BEL-U}}$. 
A similar analysis of expected error can be applied to binary encoded labels constructed to yield Johnson codes (BEL-J), in which $Q_i$ is encoded using $B_i =  b_i^1, b_i^2,...,b_i^{\scriptsize{N/2}}$, where, $b_i^k = 1$ for $\frac{N}{2}-Q_i < k-1 \leqslant  N-Q_i$, else $0$ (see \Eqref{eq:fin} in Appendix~\ref{sec:a1}). 
%expected error of predicted labels for BEL-J encoding/decoding functions 
%\deval{added this}
%%For BEL-J code, as shown in~\Eqref{eq:6}, the decoding function $\mathcal{D}^{\text{BEL-J}}$ consists of two terms. The first term determines the bit position of the first occurrence of $1$ in the binary code, and the second term determines the bit position of the last occurrence of $1$. 
%%Let $J^f(e_{1}(n),..,e_{\frac{N}{2}}(n))$ and $J^l(e_{1}(n),..,e_{\frac{N}{2}}(n))$ represent the absolute error in these bit positions. 
%%%The expected error of predicted labels for BEL-J encoding/decoding can be shown to be:
%%%\begin{equation}
%%%\label{eq:9}
%%%\EX(|\hat{y} -  y  |) \leqslant \frac{1}{N-1} \sum_{n=1}^{N-1} \Big( \EX\big(J^f(e_{1}(n),..,e_{\frac{N}{2}}(n))\big) + \EX\big(J^l(e_{1}(n),..,e_{\frac{N}{2}}(n))\big) \Big)
%%%\end{equation}
%Expanded equation~\Eqref{eq:9} for BEL-J and its explanation is provided in Appendix~\ref{sec:a1}. 
\paragraph{Error probability of classifiers: }
{
\captionsetup[sub]{font=footnotesize, skip=5pt, belowskip=-5pt}
\newdimen\imageheight
\settoheight{\imageheight}{%
\includegraphics[width=0.021\textwidth]{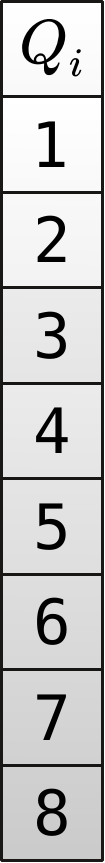}}
\begin{figure}[t]
  \centering
  \begin{subfigure}[t]{0.04\linewidth}
    \centering
  \includegraphics[height=\imageheight]{figures/codes_a.pdf}
 \caption{}
 \label{fig:1a}
\end{subfigure}
  \begin{subfigure}[t]{0.10\linewidth}
      \centering
    \includegraphics[height=\imageheight]{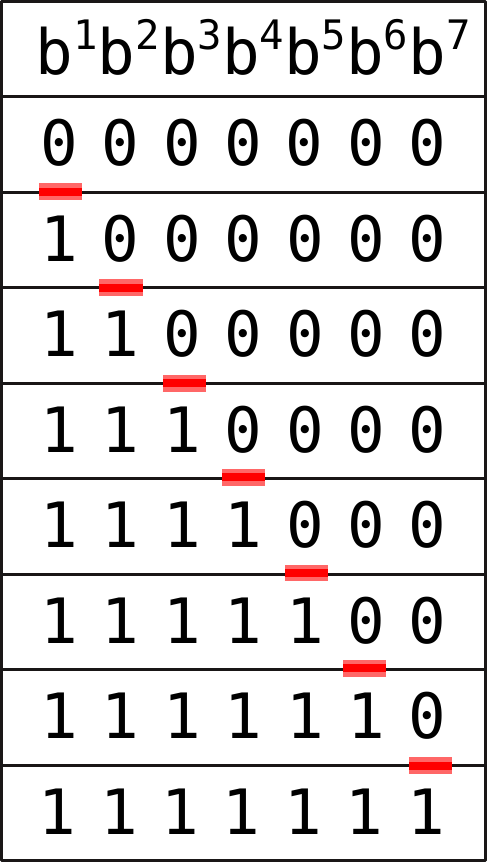}
   \caption{Unary}
   \label{fig:cu}
  \end{subfigure}
  \begin{subfigure}[t]{0.11 \linewidth}
    \centering
  \includegraphics[height=\imageheight]{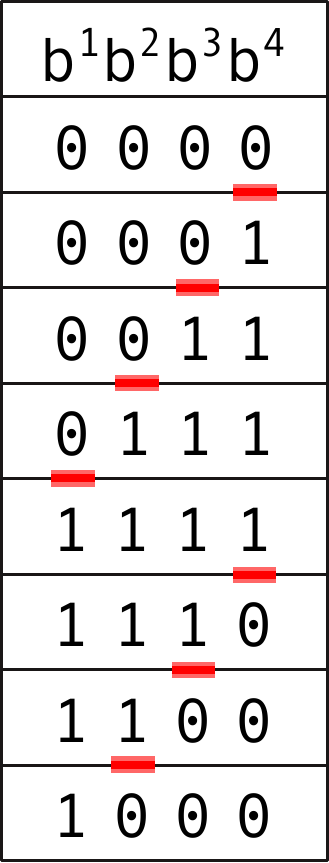}
 \caption{Johnson}
 \label{fig:cj}
\end{subfigure}
\begin{subfigure}[t]{0.12\linewidth}
  \centering
\includegraphics[height=\imageheight]{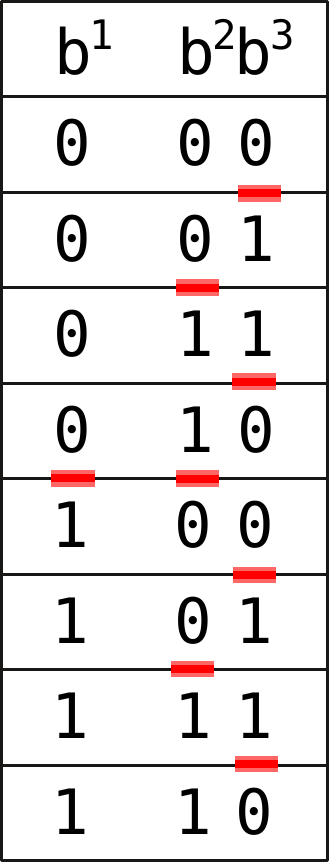}
  \caption{B1JDJn}
\label{fig:ce1jmjr}
\end{subfigure}
\begin{subfigure}[t]{0.18\linewidth}
  \centering
\includegraphics[height=\imageheight]{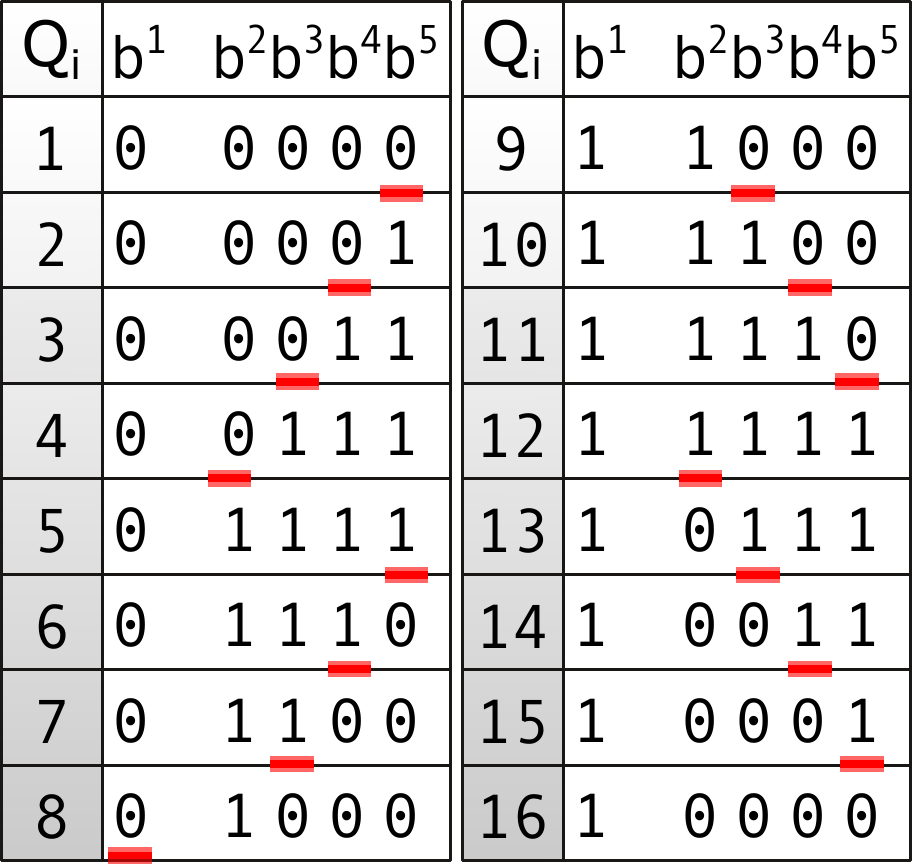}
\caption{B1JDJ}
\label{fig:ce1jmj}
\end{subfigure}
\begin{subfigure}[t]{0.18\linewidth}
  \centering
\includegraphics[height=\imageheight]{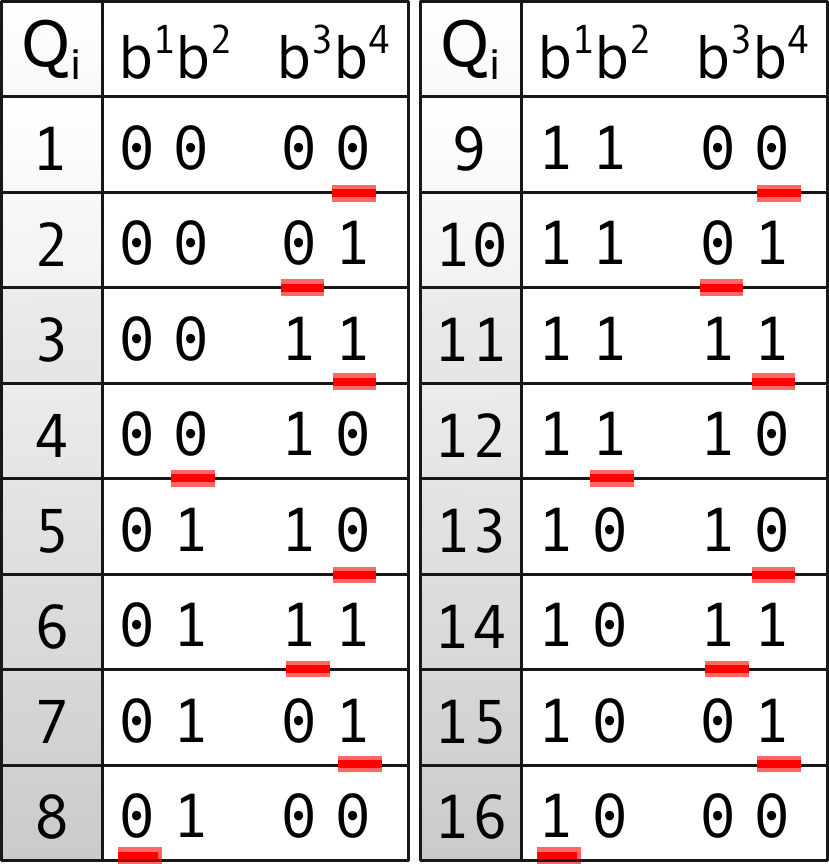}
\caption{B2JDJ}
\label{fig:ce2jmj}
\end{subfigure}
\begin{subfigure}[t]{0.22\linewidth}
  \centering
\includegraphics[height=\imageheight]{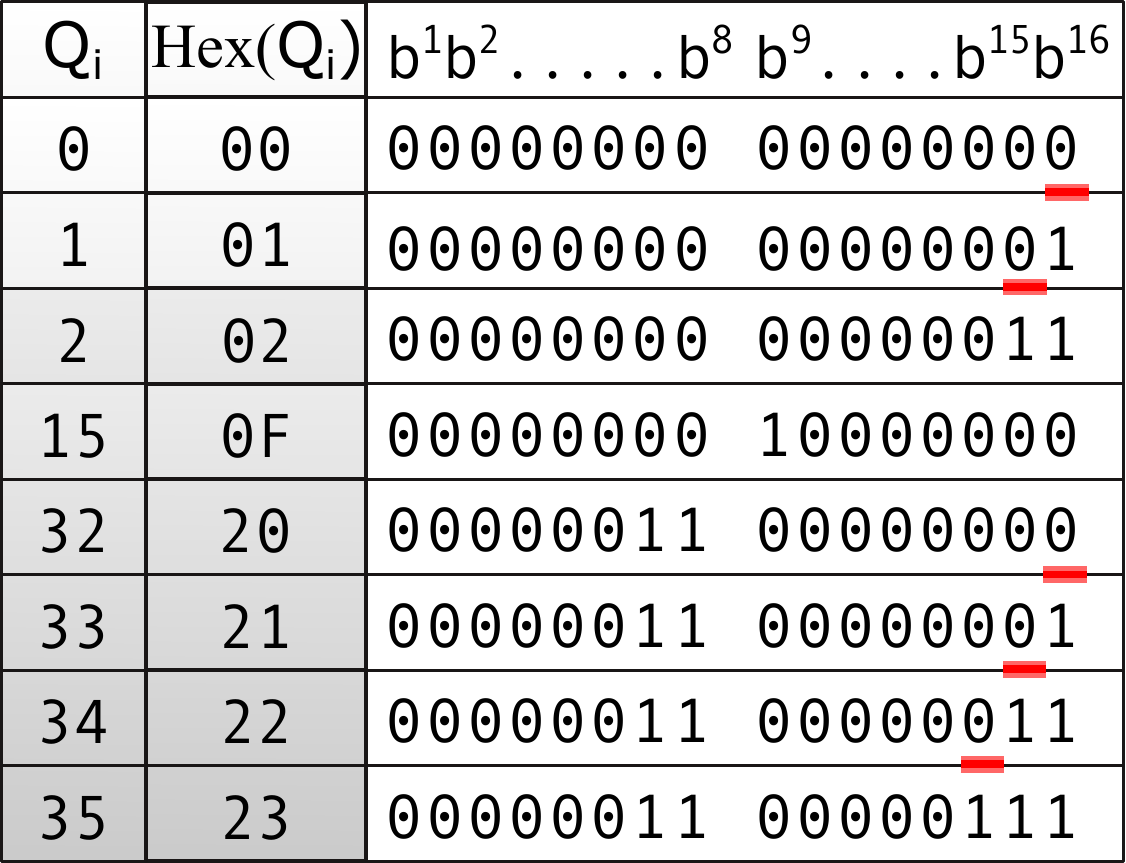}
\caption{HEXJ}
\label{fig:chex}
\end{subfigure}
  \hfill
  \caption{Examples of BEL codes. Part (a) represents the quantized values of the labels for Unary and Johnson codes shown in Parts (b) and (c). Part (d) shows a B1JDJ code without reflected binary; Parts (e) and (f) show B1JDJ and B2JDJ codes for targets in the range $1$ to $16$. Part (g) shows quantized and encoded values for a HEXJ code (space added to differentiate between base and displacement, or digits). Red lines represent \db{s}. These BEL codes described in Section~\ref{sec:cm}.}
 \label{fig:encoding}
\end{figure} 
}
\begin{figure}[t]
  \centering
  \begin{subfigure}[t]{0.31\linewidth} 
  \includegraphics[width=\textwidth]{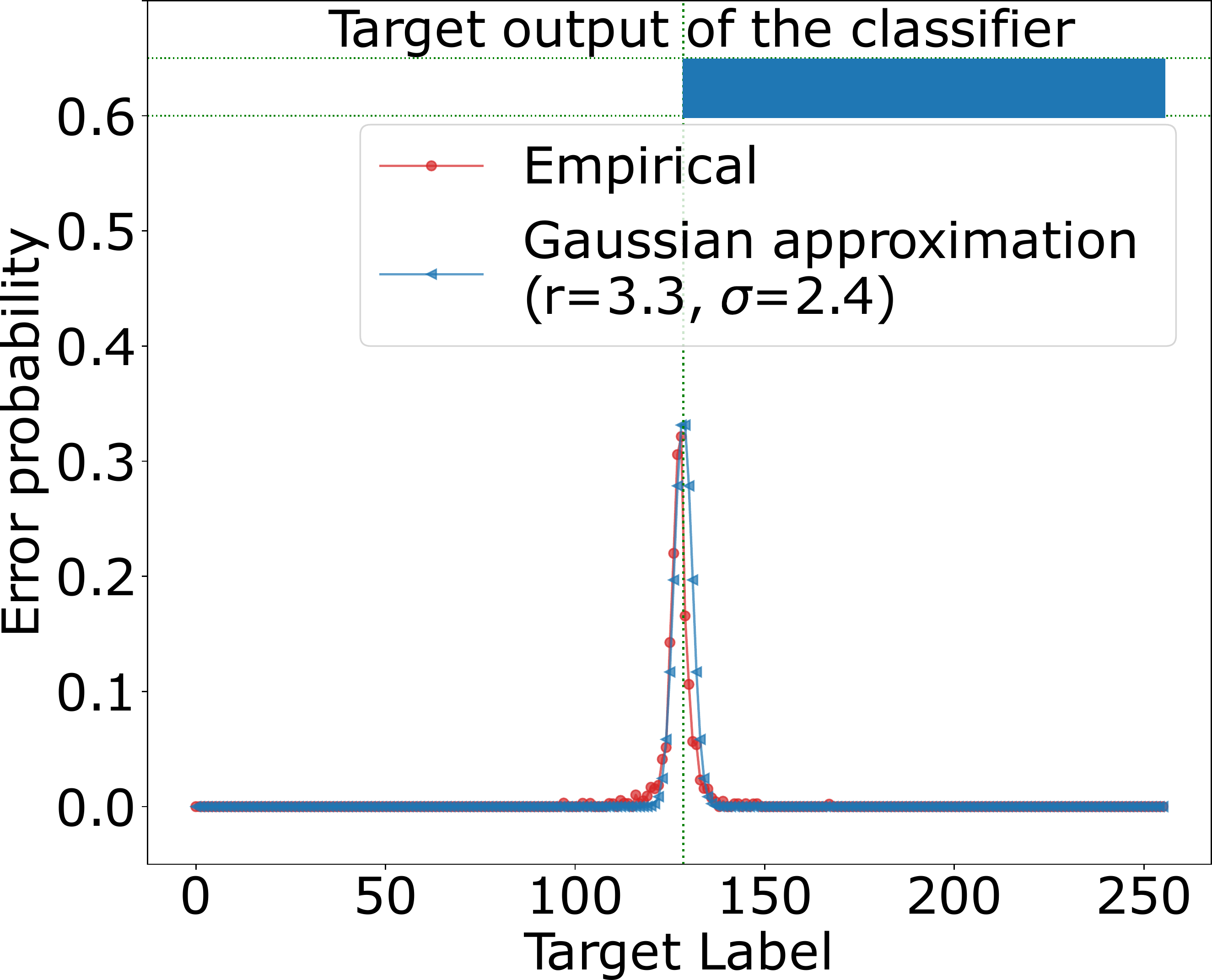}
 \caption{{Classifier A}}
 \label{fig:a1}
\end{subfigure}
  \begin{subfigure}[t]{0.31\linewidth}
      \centering
    \includegraphics[width=\textwidth]{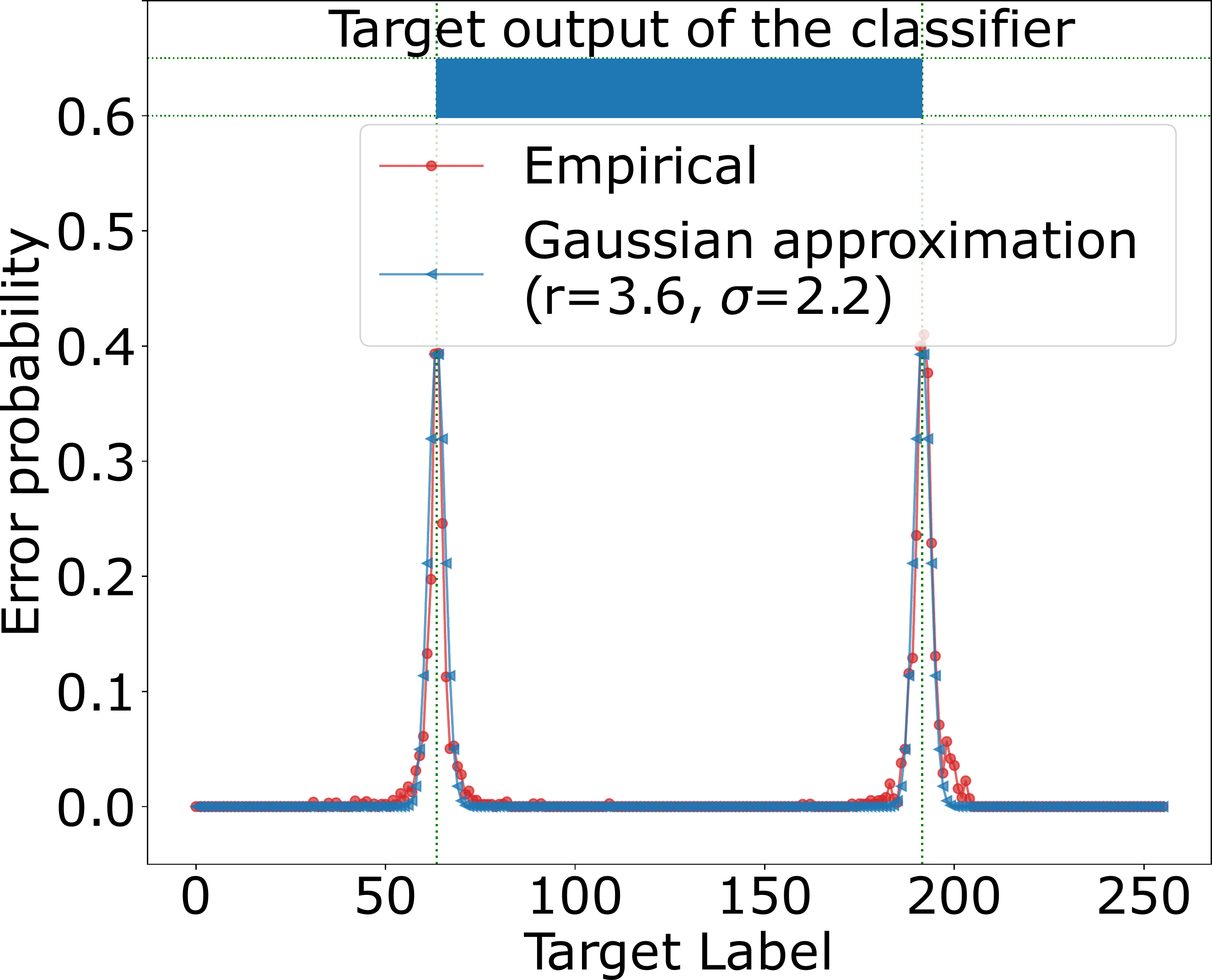}
   \caption{{Classifier B}}
   \label{fig:a2}
  \end{subfigure}
  \begin{subfigure}[t]{0.32\textwidth}
    \centering
   \includegraphics[width=\textwidth]{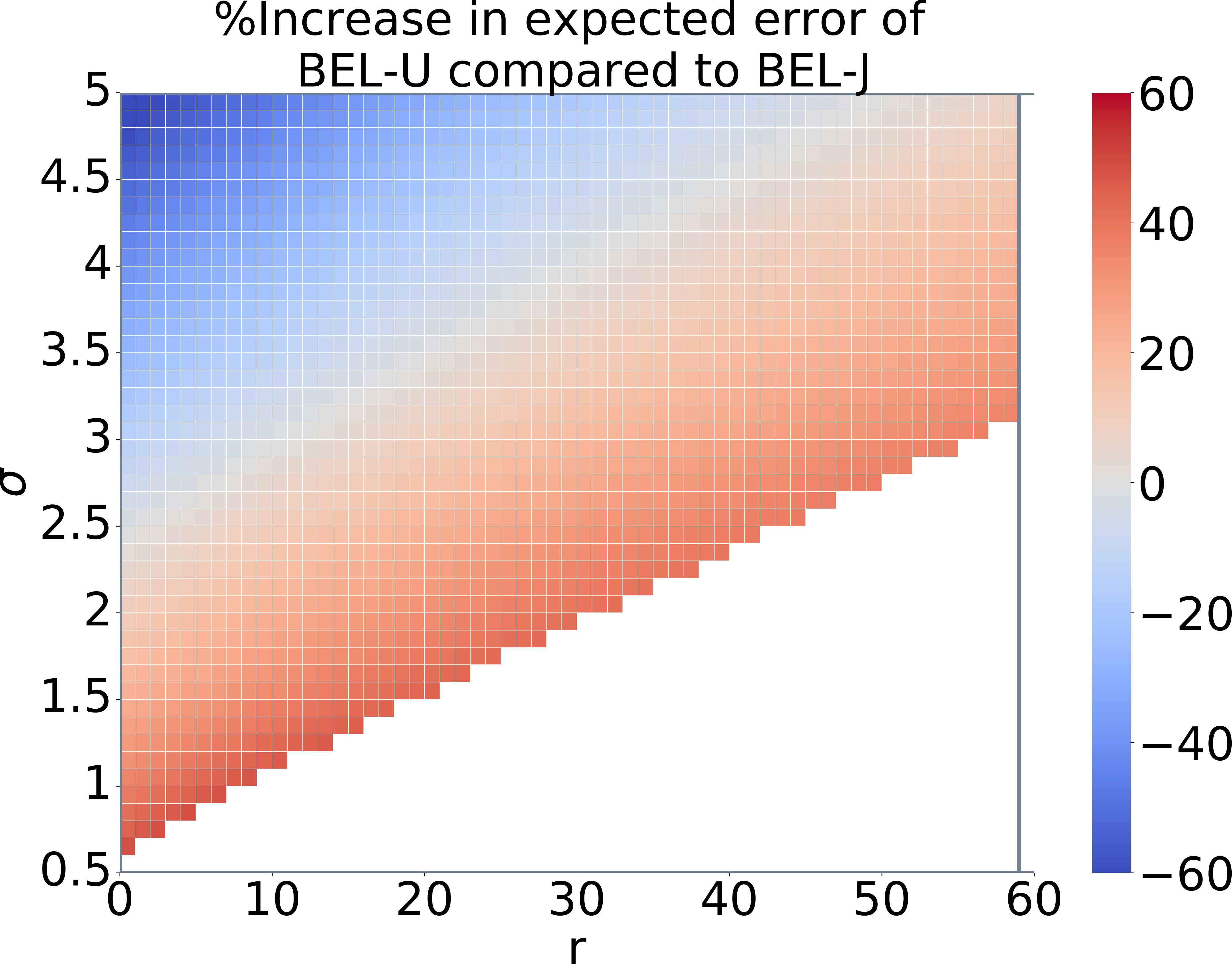}
   \caption{Expected error comparison}
   \label{fig:compare}
  \end{subfigure} 
  \caption{Part (a) and (b): classification error probability vs. target output for two classifiers. Target output $1$ where blue and $0$ elsewhere. Part (c): expected error increase of BEL-U versus BEL-J based on \Eqref{eq:10} to \Eqref{eq:n2error} (blank means that combination of $r$ and $\sigma$ results in an error probability greater than one). }
 \label{fig:accuracy}
\end{figure} 
%%The area under the error probability distribution can exceed $1$ as the labels are independent.
%In regression, the target quantized levels $Q_i \in \{1,2,...,N\}$ are on an ordinal scale. 
To use \Eqref{eq:10} we need to determine $e_{k}(n)$.
A classifier's target output is $0$ or $1$. For BEL, the target labels of a given classifier will have one or more \emph{\db{s}}
from $0 \rightarrow 1 $ or $1 \rightarrow 0$ as the target value of the regression network's output varies. 
For example, for the unary code (Figure~\ref{fig:cu}), the target output of classifier $b^2$ has a bit transition from $0$ to $1$ going from $Q_i=2$ to $Q_i=3$. 
%the target output of classifier $2$ ($b^2$) is $1$ for $Q_i > 2$ and $0$ otherwise. 
The classifier should learn a decision boundary in $(2,3)$.  Each BEL classifier is tasked with learning decision boundaries for all \db{s}. 
As the difficulty of this task varies with the number of \db{s} it varies with different encoding functions. 
%%\red{Intuitively}\deval{Yes, I am searching for some papers. I will modify this}, 
Moreover, the misclassification probability of a classifier tends to increase as the target label is closer to the classifier's decision boundaries~\citep{decision}. 
Thus, we approximate  $e_{k}(y)$ for a classifier $k$ with $t$ \db{s} as a linear combination of $t$ Gaussian distributions. Here, each Gaussian term is centered around a \db. 
Let $f_{\mathcal{N}(\mu,\sigma^2)}(y)$ denote the probability density of a normal distribution with mean $\mu$ and variance $\sigma^2$. % over the range on $n$. 
Each classifier for BEL-U encoding has one \db, whereas, each classifier for BEL-J encoding has two \db{s} (except the first and last classifiers). $e_{k}(y)$ of a classifier $k$ for BEL-U and BEL-J encoding is approximated as:
\begin{equation}
    \label{eq:temperror}
   e^{\text{BEL-U}}_{k}(y) = r  f_{\mathcal{N}(\mu_{k},\sigma^2)}(y) , \text{where, }\mu_{k} =  k + 0.5 
   \end{equation}
   \vspace{-3mm}
   \begin{equation}
 \label{eq:n2error}
e^{\text{BEL-J}}_{k}(y) = r  f_{\mathcal{N}(\mu_{1k},\sigma^2)}(y) +  r f_{\mathcal{N}(\mu_{2k},\sigma^2)}(y) , \text{where, }\mu_{1k} = \footnotesize{\frac{N}{2}} -k + 1.5 , \mu_{2k} = N -k + 1.5 
\end{equation}
Here, $r$ is a scaling factor.
Figure~\ref{fig:a1} and \ref{fig:a2} compares \Eqref{eq:temperror} and \ref{eq:n2error} against empirically observed error distributions for two classifiers using an HRNetV2-W18~\citep{hrnetface} feature extractor (backbone) trained with the COFW facial landmark detection dataset~\citep{cofw}. % for labels $y \in [1,256]$. %The top bar represents the classifier's target output ($0/1$).  
%Figure~\ref{fig:a1} and Figure~\ref{fig:a2} shows the proposed approximated  error probability distributions for two classifiers. 
\paragraph{Comparison of expected absolute error for BEL-U and BEL-J: }
Based on the above analysis, we compare the expected absolute errors of BEL-U and BEL-J. % as shows in Figure~\ref{fig:compare}. 
Figure~\ref{fig:compare} represents the percentage increase in absolute error for BEL-U compared to BEL-J for valid values of standard deviation $\sigma$ ($y-$axis) and scaling factor $r$ ($x-$axis) as used in~\Eqref{eq:temperror} and~\ref{eq:n2error}. 
Here, BEL-J has a lower error in the red-colored region (\%increase$ > 0$), whereas BEL-U has a lower error in the blue-colored region (\%increase $ < 0$). 
The figure shows that whether BEL-J or BEL-U has lower error depends upon the values of $\sigma$ and $r$. % used to approximate the classifiers’ error probability distribution. 
This suggests that the best $\{\mathcal{E},\mathcal{D}\}$ function will depend upon the classifier error probability distribution. 
The classifier error distribution in turn may depend upon the task, dataset, label distribution, network architecture, and optimization approach. 
Derivation of expected error for BEL-U and BEL-J and classifiers' empirical error probability distributions for different architectures, datasets, and encodings are provided in Appendix~\ref{sec:a1} to \ref{sec:a3}. 

\subsection{Design of Encoding Functions}
\label{sec:cm}
{
%The objective of regression is to minimize the absolute/squared error between predicted and target values.
Based on the above analysis and further empirical observation %the encoding function for BEL can be ``designed'' taking this into account as follows. 
we identify three principles for selecting BEL codes 
for regression so as to minimize error.
First, \emph{individual classifiers should require fewer \db{s}} as this makes them easier to train.
Second, a desirable property for a BEL encoding function is that
\emph{the hamming distance between two codes (number of bits that differ) should be proportional to the difference between the target values they encode.}
However, hamming distance weighs all bit changes equally. Thus, hamming distance based code design provides equal error protection capability to all bits~\citep{Wu2018DesigningCS,1180579} and does not account for which classifiers are more likely to mispredict for a given input. %target value.  
This matters because the misclassification probability of BEL classifiers is not uniform, but rather increases the closer the target value of an input is to a \db{ } (e.g., Figure~\ref{fig:a1} and \ref{fig:a2}). 
These observations yield a third important consideration: \emph{
  %The selection of codes should account for the fact that, 
  For a given target value %-target pair and BEL code, %different classifiers may have different error probabilities because
  %for a given target value, 
  classifiers closer to a \db{} are more likely to incur an error.} 
%This design property covers the impact of \db{s} in codes on continuous predictions. 

%We evaluate several encodings that to greater or lesser extent satisfy the above principles.
The principles above 
%The properties above 
%sometimes conflict 
%can conflict
highlight a 
%suggest a
tradeoff between classification error probability and error-correction properties when selecting BEL codes. %thus it is not obvious {\em a priori} which code is best.
To evaluate the trade-offs, we empirically evaluate encodings that, to greater or lesser extent, satisfy one or more of the principles
while focusing on reducing the number of classifiers (bits) so as to avoid increasing model parameters.
%We empirically evaluate the impact of these encodings on several regression tasks. 
Development of algorithms that might systematically optimize encoding functions is left to future work. 
Specifically, we explore the following codes:

\textbf{Unary code (U): }
Unary codes (Figure~\ref{fig:cu}) have only one \db{ }per classifier and thus require $M=N-1$ bits to encode $N$ values. 
Unary codes satisfy the first two principles and
prior work on ordinal regression by binary classification~\citep{ordext,agecnn} uses similar codes. %formulation of classifiers as the unary code. 

\textbf{Johnson code (J): }
%Another example of binary code is the Johnson counter code. It is a
%The Johnson code sequence (Figure~\ref{fig:cj}) is used in the design of counters in digital circuits and is based on Libaw-Craig code~\citep{libaw}. 
The Johnson code sequence (Figure~\ref{fig:cj}) is based on Libaw-Craig code~\citep{libaw}. 
We select this code as it has well-separated \db{s} and requires $M=\frac{N}{2}$ bits compared to $N$ required for unary code. 
This code exemplifies the impact of considering non-uniform classifier error probabilities (third principle).  
For example the hamming distance between $1$ and $8$ is just one.
However, the \db{} for the differing bit, for classifier $b^1$, is far from $1$ or $8$. 
Assuming equal error probability distributions centered on each \db{} for each classifier (as in \Eqref{eq:n2error}), 
$b^1$ is less likely to mispredict than $b^2$, $b^3$ or $b^4$ for inputs with target values near $1$ or $8$.  

\textbf{Base+displacement based code (B1JDJ/B2JDJ): } 
We further reduce the number of bits using a base+displacement-based representation. 
In this representation, a value is represented in base-k using a base-term {\tt b} and displacement {\tt d} via {\tt b} * k + {\tt d}. {\tt b} and {\tt d} are represented using Johnson codes. 
%%Further, to minimize the misclassification probability around the \db{ }for the radix-term, we adapt reflected binary codes~\citep{gray}. 
Further, to improve the distance between two remote codes, we adapt reflected binary codes for term {\tt d}~\citep{gray}. 
We evaluate base-2 (B1JDJ - Figure~\ref{fig:ce1jmj}) and base-4 codes (B2JDJ - Figure~\ref{fig:ce2jmj}). 
%For a naive representation in Figure~\ref{fig:ce1jmjr}, %which uses $1$ bit for the base and $2$ bits for displacement with Johnson code, 
%%only one bit changes between codes for label $1$ and $5$, which has a high misclassification probability as it is very close to the decision boundary, resulting in a high impact on the absolute error. We adapt reflected binary codes or also known as gray codes~\cite{gray} to fix this. 
%%We evaluate B1JDJ (Figure~\ref{fig:ce1jmj}) and B2JDJ (Figure~\ref{fig:ce2jmj}) that use 1-bit and 2-bit base, respectively.

\textbf{Binary coded hex - Johnson code (HEXJ): } 
%%A number can be encoded by converting each digit to a fixed binary code. We propose 
In HEXJ (Figure~\ref{fig:chex}), each digit (0-F) of the hexadecimal representation of a number is converted to an 8-bit binary code using Johnson code. For example, for the decimal number 47 (i.e., 2F in hex), HEXJ(47) =  Concetanate(Johnson(2), Johnson(F)).  A 16-bit HEXJ code can represent numbers in the range of 00 to FF (a total of 256).
%, reducing the number of classifiers significantly for a large number of quantization levels. 
The number of bits increases sublinearly with the number of quantization levels for HEXJ, making it suitable for regression problems with many quantization levels.
%A number k is converted to its hex value in this code, and each hex digit ($0-$F) is encoded separately using an $8-$bit Johnson code. 
%Figure~\ref{fig:chex} represents examples of quantized labels, their hex values, and encoded values. A $16$-bit HEXJ code can represent $256$ values, reducing the number of classifiers significantly for a large number of quantization levels. 

\textbf{Hadamard code (HAD): }
Hadamard codes~\citep{BOSE1959183} are widely used as error-correcting codes 
%due to their fault-tolerant properties and also 
and have been used 
%to encode classes 
for multiclass classification~\citep{ecoc,Verma2019}. They require $M=N$ bits to encode $N$ values. 
However, Hadamard codes violate all three BEL code selection principles: 
First, each classifier has many \db{s}. %, increasing the complexity of the decision boundary. % by each binary classifier.
Second, as each code is equidistant (hamming distance of $\frac{M}{2}$), 
the difference between target values is ignored. %not proportional to hamming distance between codes.
Finally, they protect all bits equally so do not take advantage of non-uniform error probabilities.
%Moreover, the hamming distance is not proportional to the difference in encoded values as each code word is equal distance from every other which
%makes sense if errors are equally likely in each bit, which they are not for regression problems.
%Further, the Hadamard code does not follow the desirable properties proposed in this work for regression. 
We verify empirically Hadamard codes are unsuitable for regression (Section~\ref{sec:abl}). 

\subsection{Design of Decoding Functions} 
\label{sec:dec}
We explore three decoding functions: custom decoding, correlation-based decoding, and expected correlation-based decoding. 
Custom decoding functions are specific to the encoding function, and are only evaluated for unary and Johnson codes. 
In contrast, correlation-based decoding, first explored in prior work studying ECOC for multiclass classification~\citep{multiclass}, can be applied to all codes. 
For quantized labels in $\{1,2,...,N\}$, we define a code matrix $\mC$ of size $N \times M$, where $M$ is the number of bits/classifiers used for the binary-encoded label. Each row $\mC_{k, :}$ in this matrix represents the binary code for label $Q_i=k$.  
For example, Figure~\ref{fig:cu} can be considered a code matrix, where the first row represents code for label $Q_i=1$. 
Let $\hat{Z}_i \in \mathbb{R}^M$ denote the output logit values of the classifiers. 
For decoding, the row with the highest correlation with the output $\hat{Z}_i$ is selected as the decoded label. 
Here, real-valued output $\hat{Z}_i$ is used instead of output binary code $\hat{B}_i$ to find the correlation as it uses the confidence of a classifier to make a more accurate prediction. 
For target quantized labels $Q_i \in \{1,2,...,N\}$, the decoding function is defined as:
\begin{equation}
  \label{eq:gen}
  \mathcal{D}^{\text{GEN}}(\hat{Z}_i) = \argmaxB_{k \in \{1,2,...,N\}} \Big(\hat{Z}_i \cdot\mC_{k, :} \Big)
\end{equation}

However, $\mathcal{D}^{\text{GEN}}$ outputs a quantized prediction, introducing quantization error.
To remedy this concern and demonstrate the potential of more sophisticated decoding rules, we propose and evaluate an expected correlation-based decoding function, which allows prediction of real-valued label $\hat{y}_i$. For target labels $y_i \in [1,N]$, the decoding function is defined as:
\begin{equation}
  \label{eq:genex}
  \mathcal{D}^{\text{GEN-EX}}(\hat{Z}_i) = \sum_{k=1}^N k \sigma_k, \text{where } \sigma_k = \frac{e^{\hat{Z}_i \cdot\mC_{k, :}}}{\sum_{j=1}^N e^{\hat{Z}_i \cdot\mC_{j, :}} } 
\end{equation}

\paragraph{Training loss functions: } 
A deep neural network with multiple output binary classifiers can be trained using the binary cross-entropy (BCE) loss $\mathcal{L}_{\text{BCE}} \big(\hat{Z_i}, \mathcal{E}(Q_i)\big)$. 
However, this loss minimizes the mismatch between predicted and target code but does not directly minimize the error between the target and predicted values. 
Decoding functions $\mathcal{D}^{\text{GEN}}$ and $\mathcal{D}^{\text{GEN-EX}}$ can be used to calculate the loss and minimize the mismatch between decoded predictions and target values directly. 
%%Depending upon the decoding function, different loss functions that minimize the mismatch between decoded predictions and target values can be used. 
%We also explore the use of cross-entropy and L1/L2 loss suitable for different decoding functions. 
Decoding function $\mathcal{D}^{\text{GEN}}$ finds the correlation between each row of the code matrix ($\mC_{k, :}$) and the output $\hat{Z}_i$. $\mC{ }\hat{Z}_i$ gives the correlation vector, and the index with the highest correlation is used as the predicted label. 
In this case,  cross-entropy loss  $\mathcal{L}_{\text{CE}} \big(\mC{ } \hat{Z}_i, Q_i\big)$ can be used to train the network. 
Similarly, for decoding function $\mathcal{D}^{\text{GEN-EX}}$, which predicts a continuous value, L1 or L2 loss $\mathcal{L}_{\text{L1/L2}} \big(\mathcal{D}^{\text{GEN-EX}}(\hat{Z}_i), y_i\big)$ can also be used for training. We evaluate multiple combinations of decoding and loss functions in Section~\ref{sec:eval}. 

\subsection{Regression Network Architecture for BEL} % for Binary-encoded Label Regression}
\label{sec:proposed}
\newdimen\imageheight
\settoheight{\imageheight}{%
\includegraphics[width=0.40\textwidth]{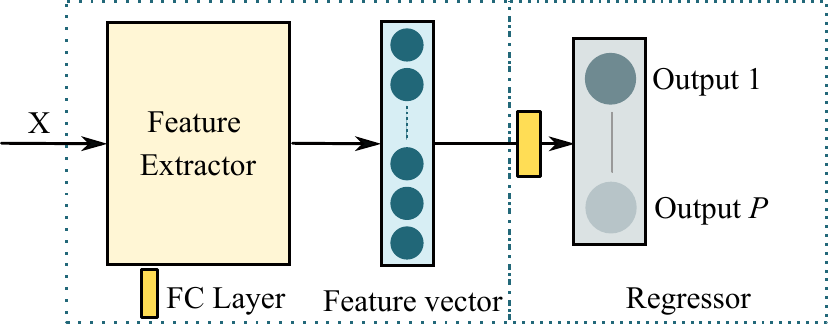}}
\begin{figure*}[t]
  \centering
  \begin{subfigure}[b]{0.40\linewidth}
      \centering
    \includegraphics[height=\imageheight]{figures/baseline.pdf}
   \caption{Direct regression}
   \label{fig:2a}
  \end{subfigure}
  \begin{subfigure}[b]{0.59\linewidth}
      \centering
    \includegraphics[height=\imageheight]{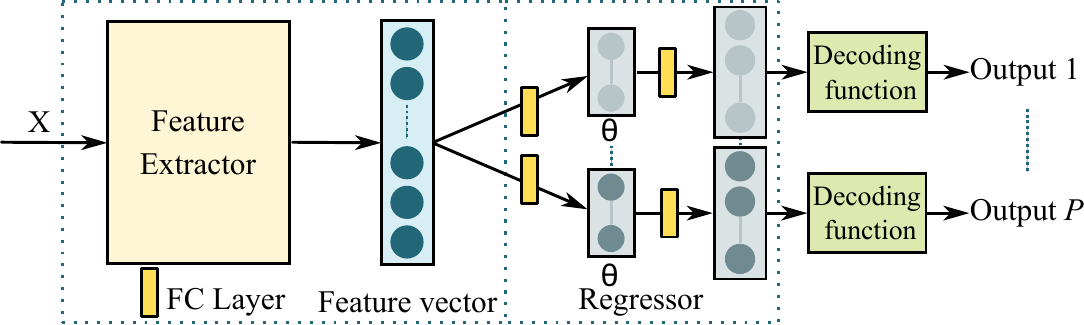}
   \caption{Binary-encoded labels}
   \label{fig:2b}
  \end{subfigure}
 \caption{Network architecture for direct and BEL regression; only the regressor architecture is modified, but the entire network is trained end to end. 
  $P$ is the number of dimensions of the regression network output.}
 \label{fig:method}
\end{figure*} 
%%%%This section explains the integration of \BEL{ }regression approach with deep neural networks and the proposed modifications to reduce the number of network parameters. 
A regression network typically consists of a feature extractor and regressor. The regressor consists of a fully connected layer between the feature extractor's output (i.e., feature vector) and output logits for direct regression as shown in Figure~\ref{fig:2a}. 
In BEL, the number of output logits is increased to the number of classifiers (bits) used.
%All classifiers share parameters for higher-level feature extraction. 
%\deval{added classifiers in the previous statements}
When $y \in \mathbb{R}^P$, with $P>1$, the required number of output logits---$P \times M$ assuming $M$-bit BEL encoding per output dimension---can significantly increase the size of the regression layer.
However, empirically, we find small feature vectors suffice as the output logits are highly correlated for the explored encoding functions.
%in such cases. 
Adding a fully connected bottleneck layer to reduce feature vector size to \emph{$\theta$} reduces the number of parameters and provides a trade-off between the model size and accuracy. 
Figure~\ref{fig:2b} shows the modified network architecture for BEL. % to predicts $P$ labels. %%, where each label is encoded to $M_p$ bits. 
% Thus, in both the case, smaller feature vectors is sufficient even though output neurons are more than $\theta$, as we show in Section~\ref{sec:abl}.  
%Figure~\ref{fig:2a} represents the baseline neural network architecture for direct regression. 
%For example, in head pose estimation, a CNN is used to predict three angles yaw, pitch, and roll to determine head pose.  
%In the modified architecture, each label is encoded to $M_p$ bits.
% uses a feature vector of size $\theta$ and $M_p$ output neurons where $M_p$ is the number of bits required to encode label $p$. 
%%%If \text{FV} is the feature vector size used in direct regression, the weight parameters for $\BEL$ regressor is equal to $\sum_{p=1}^{\text{P}} ( \text{FV} \times \theta + M_p \times \theta)$. 

%\input{sections/methodology.tex}
%\section{Experimental Setup}
\section{Evaluation}
%% The table for benchmarks
\begin{table}[t]
  \centering
  \setlength\tabcolsep{4pt}
  \caption{Benchmarks used for evaluation} 
  \label{tab:benchmarks}
  \scriptsize
  \begin{tabular}{C{2.6cm}C{1.1cm}C{3.0cm}ccC{1.9cm}C{0.4cm}}
    \toprule
   Task  & Feature Extractor  & Specialized Approach &  Dataset & Benchmark & Label range/ Quantization levels & $\theta$ \\ \midrule 
   \multirow{4}{\linewidth}{\centering Landmark-free 2D head pose estimation}  & \multirow{2}{\linewidth}{\centering ResNet50} & \multirow{2}{\linewidth}{\centering Regression+classification\\\citep{hopenet}} & BIWI & HPE1 & -100-100/200 & 10 \\ \cline{4-7}
   &  &   & 300LP/AFLW2000 & HPE2  & -100-100/200 & 10 \\ \cline{2-3}\cline{4-7}
   & \multirow{2}{\linewidth}{\centering RAFANet} & \multirow{2}{\linewidth}{\centering Direct regression\\\citep{rafanet}}  & BIWI & HPE3 & -180-180/360 & 50 \\  \cline{4-7}
   &  &   & 300LP/AFLW2000 & HPE4 & -180-180/360 & 50 \\  \cline{1-7}
   \multirow{4}{\linewidth}{\centering Facial Landmark Detection}  & \multirow{4}{\linewidth}{\centering HRNetV2-W18} & \multirow{4}{\linewidth}{\centering Heatmap regression\\\citep{hrnetface,anchorface}} & COFW & FLD1 & 0-256/256 & 10 \\ \cline{4-7}
   &  &   & 300W & FLD2 & 0-256/256 & 10  \\  \cline{4-7}
   &  &   & WFLW & FLD3  & 0-256/256 & 10\\  \cline{4-7}
   &  &   & AFLW & FLD4  & 0-256/256 & 30\\  \cline{1-7}
  \multirow{2}{\linewidth}{\centering Age estimation}  & \multirow{2}{\linewidth}{\centering ResNet50 /ResNet34} & \multirow{2}{\linewidth}{\centering Ordinal regression\\\citep{coralcnn}} & MORPH-II & AE1 & 0-64/64 & 10 \\ \cline{4-7}
   &  &   & AFAD & AE2 & 0-32/32 & 10 \\  \cline{1-7}
  End-to-end autonomous driving & PilotNet & Direct regression \citep{pilotnet} & PilotNet & PN & 0-670/670 & 10 \\ 
  \bottomrule
  %\cmidrule(r){1-4}
  \end{tabular}
  \vspace{-3mm}
  \end{table}
%% End of the table
Table~\ref{tab:benchmarks} summarizes the tasks, datasets, and network architectures used for the evaluation of BEL. These tasks are commonly used for evaluation of regression approaches by prior works due to the complexity of problem and network architectures~\citep{softcode}. 
Landmark-free 2D head pose estimation (HPE) aims to find a human head's pose in terms of three angles: yaw, pitch, and roll from a 2D image without landmarks. Facial landmark detection (FLD) is a problem of detecting the $(x,y)$ coordinates of keypoints in a given face image. Age estimation aims to predict the age of a person from an image. In end-to-end autonomous driving, the steering wheel's next angle is predicted from an image of the road. Normalized Mean Error (NME) and Mean Absolute Error (MAE) with respect to raw real-valued labels are used as the evaluation metric for FLD and the rest, respectively.  

%Our main goal is to compare the regression approach; hence 
%We use the data augmentation and evaluation protocols used by the existing approach (Table~\ref{tab:benchmarks}) for each benchmark.
We also evaluate direct regression and multiclass classification as baseline regression approaches. 
For direct regression, L1 or L2 loss functions are used. Label values are scaled to reduce the range of labels. The loss function and the scaling factors are set using hyperparameter tuning.  
In the multiclass classification-based regression, the target values are quantized and converted to a class. The network is trained using cross-entropy loss in this case. 
In our evaluation, the entire network (i.e., feature extractor and regressor) are trained end-to-end for direct regression, multiclass classification, and BEL. 
The feature extractor, data augmentation, evaluation protocols, and the number of training iterations are kept uniform across different methods for each benchmark. 
We report average of five training runs and error margin of $95\%$ confidence interval. Details on datasets, training parameters, related work for specific tasks, and other evaluation metrics are provided in Appendix~\ref{sec:a3}. 
\label{sec:eval}
\begin{figure*}[t]
  \centering
  \includegraphics[width=0.95\linewidth]{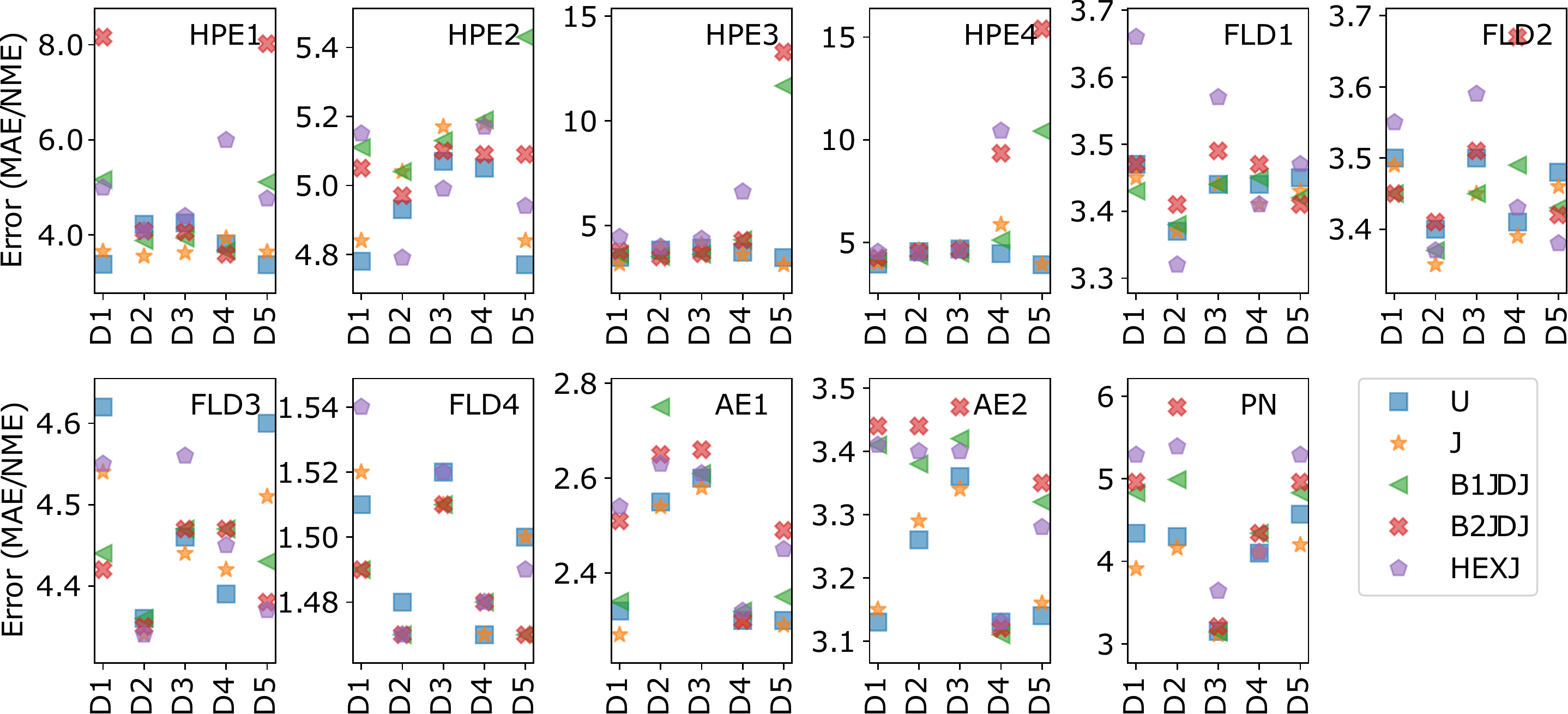}
\caption{Error (MAE or NME) for different encoding, decoding, and loss functions for \BEL. D1-D5 represents different combinations of decoding and loss functions: D1 (BCE loss with BEL-U/BEL-J/GEN decoding for U/J/others), D2  (CE/GEN-EX), D3 (CE/GEN), D4 (L1 or L2/GEN-EX), and D5 (BCE/GEN-EX). }
\label{fig:compare_decoding}
\end{figure*}

BEL introduces several design parameters for regression by binary classification. 
We evaluate different encoding ($\mathcal{E}$), decoding ($\mathcal{D}$), and training loss ($\mathcal{L}$) functions for BEL across all the benchmarks and study the extent and nature of the impact of these design parameters on accuracy. 

\paragraph{Encoding function ($\mathcal{E}$):}
\iffalse
\begin{figure*}[t]
  \centering
  \includegraphics[width=0.75\linewidth]{figures/compare_encode.pdf}
\caption{Mean Absolute Error(MAE)/Normalized Mean Error (NME) for different encoding functions for \BEL.}
\label{fig:compare_encoding}
\end{figure*}
\fi
%%we further analyze the effect of different encoding functions on the error to demonstrate the vast design space of encoding functions. 
Figure~\ref{fig:compare_decoding} plots error (MAE or NME) using different encodings.
We do not show results for Hadamard codes here as it results in significantly higher error than other encodings (Appendix~\ref{sec:abl}). On average, Hadamard codes result in $\sim60\%$ higher error than J encoding, which shows that these codes are unsuitable for regression. 
The results show the encoding function significantly affects the accuracy and the
best-performing encoding function varies across tasks, datasets, and network architectures (e.g., HPE1 and HPE3 are trained on the same dataset and different architecture). 
In Section~\ref{sec:err} we observed that which encoding/decoding functions result in lower error depends upon the classifiers' error distribution. 
%%For HPE3, FLD1, and AE1, J does better than U for D1 (decoding functions used for analytical comparison in Section~\ref{sec:err}); we attribute this to misclassification errors occurring more frequently near \db{s} based on the analytical study.
For decoding functions used for the comparison in Section~\ref{sec:err}, J does better than U for HPE3, FLD1, and AE1; we attribute this to misclassification errors occurring more frequently near \db{s} based on the analytical study.

The encoding function impacts the number of classifiers and the complexity of the function to be learned by a classifier. We observe a trade-off between these two parameters. For some benchmarks, the availability of sufficient training data and network capacity facilitates the learning of complex classifiers such as B2JDJ. In such a case, a reduced number of classifiers compared to U, J, or B1JDJ codes results in a lower error. 
We provide empirical results for the same in Appendix~\ref{sec:abl}. 
%We empirically observe that as the number of training data samples is reduced, codes with simpler classifiers such as U and J perform better (Section A1). 
%Similarly, the number of labels to be learned by a single network affects the suitable encoding. For example, for tasks with more output labels, reducing the number of output classifiers using complex encoding improves accuracy as observed empirically (e.g.,  WFLW). 

\paragraph{Decoding  ($\mathcal{D}$) and training loss  ($\mathcal{L}$) functions:} 
We explore three decoding and three training loss functions (Section~\ref{sec:dec}). 
%The training loss functions can be selected from BCE, CE, and L1/L2 loss. 
However, not all the combinations of decoding and loss functions ($\mathcal{D}$/$\mathcal{L}$) perform well. 
For example, CE, L1, or L2 losses do not use decodings $\mathcal{D}^{\text{BEL-J}}$ or $\mathcal{D}^{\text{BEL-U}}$. Therefore, optimizing the network for these losses does not directly minimize the absolute error between targets and decoded predictions. 
%%In contrast, $\mathcal{D}^{\text{GEN-EX}}$ is used for L1/L2 loss calculation; in such a case, the training loss directly minimizes the error between targets and decoded predictions. 
We present results for five out of nine $\mathcal{D}$/$\mathcal{L}$ combinations.  
%We proposed three decoding functions (Section~\ref{sec:dec}) that can be combined with four training loss functions. We evaluate four out of such combinations to analyze the design space of decoding and loss functions as listed in Table~\ref{tab:loss}. 
Figure~\ref{fig:compare_decoding} compares error (MAE or NME) achieved by different $\mathcal{D}$/$\mathcal{L}$ combinations and highlights the range of error variations. $\mathcal{D}^{\text{GEN-EX}}$ results in the lowest error for the majority of the benchmarks as it reduces quantization error and also utilizes the output logit confidence values. 
$\mathcal{D}^{\text{GEN-EX}}$ consistently perform better than $\mathcal{D}^{\text{GEN}}$ function that has been used for multiclass classification by prior works~\citep{multiclass}. 
The use of CE or L1/L2 loss results in a lower error with $\mathcal{D}^{\text{GEN-EX}}$ for most benchmarks as the training loss function directly minimizes the error between targets and decoded predictions. 
%%In contrast, BCE loss does not directly minimize the error between decoded and target labels.  
%The choice of loss and decoding function has a significant impact on error and can be co-designed to improve the accuracy.   

\paragraph{Comparison of BEL with regression approaches: } 
% Table for the main results
\begin{table}[t]
  \centering
  \setlength\tabcolsep{4pt}
  \caption{Comparison of BEL with different regression approaches. ``Specialized approach'' described in Table~\ref{tab:benchmarks} } 
  \label{tab:compare_regression}
  \scriptsize 
  \begin{tabular}{L{3cm}L{2.35cm}L{2.35cm}L{2.35cm}L{2.35cm}}
  \toprule
  & \multicolumn{4}{c}{ Error (MAE or NME) / Model size }   \\ \hline
  Approach  & \multicolumn{1}{c}{HPE1}        & \multicolumn{1}{c}{HPE2}        & \multicolumn{1}{c}{HPE3}        & \multicolumn{1}{c}{HPE4}         \\ \midrule
  Specialized approach & -           & - &  {3.40}  / 69.8M & 4.14 / 69.8M  \\ \hline
  Direct regression   & 4.76 {$\pm$0.35} / 23.5M & 5.65 {$\pm$0.13} / 23.5M &   3.40 {$\pm$0.26} / 69.8M &   4.14 {$\pm$0.12} / 69.8M  \\ \hline
  Multiclass classification   &  4.49 {$\pm$0.24} / 24.2M & 5.31 {$\pm$0.05} / 24.8M &  4.54 {$\pm$0.04} / 72.0M &  5.14 {$\pm$0.08} / 72.0M \\ \hline
  BEL  &  \textbf{3.56}$\pm$0.01 / 23.6M &  \textbf{4.77} {$\pm$0.05} / 23.6M &  \textbf{3.30}$\pm$0.04 / 69.8M & \textbf{3.90} {$\pm$0.03} / 69.8M   \\ \hline     
  BEL $\mathcal{E}/\mathcal{D}/\mathcal{L}$ functions &U/GEN-EX/L2 &  U/GEN-EX/BCE&  B1JDJ/GEN-EX/BCE&  U/GEN-EX/BCE \\ \bottomrule
  \end{tabular}
  \begin{tabular}{L{3cm}L{2.35cm}L{2.35cm}L{2.35cm}L{2.35cm}}
    \toprule
  Approach   & \multicolumn{1}{c}{FLD1}  & \multicolumn{1}{c}{FLD2} &  \multicolumn{1}{c}{FLD3}    & \multicolumn{1}{c}{FLD4}     \\ \midrule
  %Specialized approach &  3.45 / { }{ }9.6M  &   \textbf{3.32} / { }{ }9.6M &  \textbf{4.32} / { }{ }9.6M &  1.57 / { }{ }9.6M  \\ \hline
Specialized approach  &  3.45 / { }{ }9.6M  &   \textbf{3.32} / { }{ }9.6M &  \textbf{4.32} / { }{ }9.6M &  1.57 / { }{ }9.6M  \\ \hline
Direct regression          & 3.60 {$\pm$0.02} / 10.2M  & 3.54 {$\pm$0.03} / 10.2M &    4.64 {$\pm$0.03} / 10.2M &  1.51 {$\pm$0.01} / 10.2M  \\ \hline
Multiclass classification   & 3.58 {$\pm$0.03} / 25.4M &  3.51 {$\pm$0.02} / 45.2M &  4.50 {$\pm$0.01} / 61.3M & 1.56 {$\pm$0.01} / 20.1M  \\ \hline
BEL                       &  \textbf{3.34} {$\pm$0.02} / 10.6M &  {3.40} {$\pm$0.02} / 11.2M &   {4.36} {$\pm$0.02} / 11.7M &  \textbf{1.47} {$\pm$0.00} / 10.8M\\ \hline  
BEL $\mathcal{E}/\mathcal{D}/\mathcal{L}$ functions  & HEXJ/GEN-EX/CE & U/GEN-EX/CE & B1JDJ/GEN-EX/CE & B1JDJ/GEN-EX/CE \\ \bottomrule   
\end{tabular}
\begin{tabular}{L{3cm}L{2.35cm}L{2.35cm}L{2.35cm}L{2.35cm}}
\toprule
Approach  &  \multicolumn{1}{c}{AE1}   & \multicolumn{1}{c}{AE2}   & \multicolumn{1}{c}{PN}   &      \\ \midrule
%Specialized approach &  2.49 / 21.3M &  3.47 / 21.3M &  4.24 / 1.8M& \\ \hline
Specialized approach   &  2.49 / 21.3M &  3.47 / 21.3M &  4.24 / 1.8M& \\ \hline
Direct regression         & 2.44 {$\pm$0.01} / 23.1M & 3.21 {$\pm$0.02} / 23.1M & 4.24 {$\pm$0.45} / 1.8M& \\ \hline
Multiclass classification   & 2.75 {$\pm$0.03} / 23.1M & 3.38 {$\pm$0.05} / 23.1M & 5.54 {$\pm$0.00} / 1.9M & \\ \hline
BEL                       & \textbf{2.27} {$\pm$0.01} / 23.1M & \textbf{3.11} {$\pm$0.00} / 23.1M &  \textbf{3.11} {$\pm$0.01} / 1.8M& \\ \hline
BEL $\mathcal{E}/\mathcal{D}/\mathcal{L}$ functions   &   J/BEL-J/BCE   &  B1JDJ/GEN-EX/L1 & J/GEN/CE    \\ \bottomrule
\end{tabular}
\end{table}
%% End of the results table
Table~\ref{tab:compare_regression} compares BEL with other approaches for different benchmarks (Table~\ref{tab:benchmarks}).  
We explore and evaluate multiple combinations of encoding ($\mathcal{E}$), decoding ($\mathcal{D}$), and loss ($\mathcal{L}$) functions for BEL in this work. 
In these experiments $20\%$ of the training set is used as validation set and the validation error is used to choose the best BEL approach. 
An ablation study for using more fully connected layers for direct regression and multiclass classification is in Appendix~\ref{sec:abl}. 
%%%For direct regression, multiclass classification, and BEL, the model is trained on $80\%$ training data, and the validation error on remaining $20\%$ data is used to choose the best model. Test error for the best models is then reported.   
%%%However, Earlier works on specialized approaches report results with training using complete training set. We have rerun with training-validation set for some of the specialized approaches.  
%%%The best combination of these functions and its error is reported in Table~\ref{tab:compare_regression}. 
\BEL{ }results in lower error than direct regression and {multiclass classification} and even outperforms task-specific regression approaches for several benchmarks. 
%BEL outperforms direct regression, \red{multiclass classification}, and application-specific regression approaches for the majority of the benchmarks. 

\emph{The results show no single combination of encoding/decoding/loss functions evaluated was best for all benchmarks but also
demonstrate BEL improves accuracy across a range of regression problems.}
%The best performing encoding/decoding function pair may be task, dataset, and network specific.
%An intriguing possibility this suggests is developing automated approaches to optimizing these functions, which we leave to future work.  
%As shown in Figure~\ref{fig:compare_regression}, the best performing encoding function varies across tasks, datasets, and network architectures, which suggests that these parameters affect the suitability of an encoding/decoding functions
%The feature extractor architecture, data augmentation techniques, and the number of training iterations are kept the same for different regression approaches. 
%%%%\red{add results about the effect of model capacity on the optimal coding and the value of theta in the ablation study}
%\begin{table}[]
  
  %\end{table}
%We present our observations from evaluating different encodings, which can be further used to guide the design space explanation. 
%Overall, BEL-J results in comparable or better accuracy than the existing approaches by slight modifications in widely used classification networks. 
%, while also achieving greater performance at the same level of sparsity. 
%We can take advantage of the correlation between different bits using lower values of $\theta$ to reduce the number of parameters in the regression layer. Table~\ref{tab:bit} shows the effect of $\theta$ on the MAE for encodings with a different correlation between their output bits. As shown in the figure, MAE for encodings with higher correlation is affected by lower values of $\theta$, whereas encodings with less correlation are affected. 

\section{Conclusion}
\label{sec:con}
This work proposes binary-encoded labels (BEL) to pose regression as binary classification.  
We propose a taxonomy identifying the key design aspects for regression by binary classification and demonstrate the impact of classification error and encoding/decoding functions on the expected label error. 
Different encoding, decoding, and loss functions are explored to evaluate our approach using four complex regression tasks. 
%%We also perform limited exploration of the regressor architecture. 
BEL results in an average $9.9\%$, $15.5\%$, and $7.2\%$ lower error than direct regression, multiclass classification, and task-specific regression approaches, respectively. 
BEL improves accuracy over state-of-the-art approaches for head pose estimation (BIWI, AFLW2000), facial landmark detection (COFW), age estimation (AFAD), and end-to-end autonomous driving (PilotNet). 
%This work mainly explores the encoding and decoding design characteristics and introduces a large design space. 
Our analysis and empirical evaluation in this work demonstrate the potential of the vast design space of \BEL{ }for regression problems and the importance of finding suitable design parameters for a given task. 
The best performing encoding/decoding function pair may be task, dataset, and network specific.
A possibility this suggests, which we leave to future work, is that it may be beneficial to develop automated approaches for optimizing these functions. 

\section{Acknowledgements}
%%We thank Dave Evans, Jonathan Lew, Saurabh Kumar, and the anonymous reviewers for their valuable comments on this work. 
%This research has been funded in part by the National Sciences and Engineering Research Council of Canada (NSERC) Strategic Project Grant number STP 506681-17. 
This research has been funded in part by the National Sciences and Engineering Research Council of Canada (NSERC) Strategic Project Grant. 
Tor M. Aamodt serves as a consultant for Huawei Technologies Canada Co. Ltd. and Intel Corp. 
Deval Shah is partly funded by the Four Year Doctoral Fellowship (4YF) provided by the University of British Columbia. 
\paragraph{Reproducibility: }
We have provided a detailed discussion about training hyperparameters, experimental setup, and modifications made in publicly available network architectures in Appendix~\ref{sec:hpe}-\ref{sec:pn} for all benchmarks. 
Code is available at \url{https://github.com/ubc-aamodt-group/BEL_regression}. We have provided the training and inference code with trained models. 
\paragraph{Code of Ethics}
Some of the major applications of regression problems are artificial intelligence and autonomous machines, and regression improvement can accelerate the development of autonomous systems. However, depending upon the use, autonomous systems can have some negative societal impacts, such as job loss in some sectors and ethical concerns.
\bibliography{ref}
\bibliographystyle{iclr2022_conference}
\newpage
\appendix
%%%\section{Appendix}
%\appendix
%\section{Supplemental Material}
%%%Section~\ref{sec:a1} provides the derivation of equations uses for expected error analysis. Section~\ref{sec:a2} provides empirical and approximate error probability plots for binary classifiers trained on different networks and datasets. Section~\ref{sec:a3}-\ref{sec:a6} provides the information of datasets, training parameters, related work, and results with error bars for head pose estimation, facial landmark detection, age estimation, and end-to-end autonomous driving. 
%We also validate BEL for sparse models on facial landmark detection in Section~\ref{sec:a4}.
%%%%We have attached the training and inference code implementation ``BEL\_code.zip'' with this submission. 
%%%%Please refer to ``BEL\_code/README.md'' for more information about the code. 

%A1: Ablation study
%A2: proof of all equations
%A3: Empirical error probabilities, statistical validation of all the results
%%A4: dataset, evaluation, related work
%5: Training framework, GPU usage, hyperparameters
%A6: New results
%A7: code and how to run 
\section{Ablation study}
\label{sec:abl}
\paragraph{Impact of combination of encoding, decoding, and loss functions: }
We propose multiple combinations of encoding, decoding, and loss functions that can be used with BEL. In Tables~\ref{tab:edl2resnet}-~\ref{tab:edlpilot}, we show the effect of each combination of encoding, decoding, and loss function on the error of the model. Although general trends exist and some combinations perform consistently well across datasets, the optimal combination varies based on the dataset.
\begin{table}[h]
  \centering
  \scriptsize 
  \caption{Comparison of BEL design parameters on MAE for head pose estimation with BIWI dataset and ResNet50 feature extractor (HPE1).}
  \label{tab:edl2resnet}
  \begin{tabular}{C{2cm}C{1.5cm}rrrrrr}
  \toprule
  & \multicolumn{1}{c}{} & \multicolumn{6}{c}{Encoding function}     \\ \cline{3-8} 
  Decoding function   & Loss function                & \multicolumn{1}{l}{U} & \multicolumn{1}{l}{J} & \multicolumn{1}{l}{B1JDJ} & \multicolumn{1}{l}{B2JDJ} & \multicolumn{1}{l}{HEXJ} & \multicolumn{1}{l}{HAD} \\ \midrule
  BEL-J/BEL-U & BCE & 3.38 & 3.65 & - & - & - & - \\\hline
  GEN-EX & BCE & \textbf{3.37} & 3.64 & 5.11 & 8.02 & 4.76 & 7.53 \\\hline
  GEN & BCE & 3.38 & 3.65 & 5.16 & 8.16 & 4.99 & 7.73 \\\hline
  GEN-EX & CE & 4.22 & 3.55 & 3.88 & 4.08 & 4.09 & 5.50 \\\hline
  GEN & CE & 4.25 & 3.62 & 3.93 & 4.06 & 4.39 & 5.48 \\\hline
  GEN-EX & L2 & 3.56 & 3.93 & 3.66 & 3.59 & 5.99 & 4.21 \\ \bottomrule        
  \end{tabular}
  \end{table}
\begin{table}[h]
\centering
\scriptsize 
\caption{Comparison of BEL design parameters on MAE for head pose estimation with 300LP/AFLW2000 datasets and ResNet50 feature extractor (HPE2).}
\label{tab:edl1resnet}
\begin{tabular}{C{2cm}C{1.5cm}rrrrrr}
\toprule
& \multicolumn{1}{c}{} & \multicolumn{6}{c}{Encoding function}     \\ \cline{3-8} 
Decoding function   & Loss function                & \multicolumn{1}{l}{U} & \multicolumn{1}{l}{J} & \multicolumn{1}{l}{B1JDJ} & \multicolumn{1}{l}{B2JDJ} & \multicolumn{1}{l}{HEXJ} & \multicolumn{1}{l}{HAD} \\ \midrule
BEL-J/BEL-U & BCE & 4.78 & 4.84 & - & - & - & - \\ \hline
GEN-EX & BCE & \textbf{4.77} & 4.84 & 5.43 & 5.09 & 4.94 & 7.84 \\\hline
GEN & BCE & 4.78 & 4.87 & 5.11 & 5.05 & 5.15 & 8.54 \\\hline
GEN-EX & CE & 4.93 & 5.04 & 5.04 & 4.97 & 4.79 & 5.64 \\\hline
GEN & CE & 5.07 & 5.17 & 5.13 & 5.10 & 4.99 & 5.62 \\\hline
GEN-EX & L2 & 5.05 & 5.18 & 5.19 & 5.09 & 5.17 & 5.07 \\ \bottomrule        
\end{tabular}
\end{table}
%%%%
\begin{table}[h]
  \centering
  \scriptsize 
  \caption{Comparison of BEL design parameters on MAE for head pose estimation with BIWI dataset and RAFA-Net feature extractor (HPE3).}
  \label{tab:edl2rafa}
  \begin{tabular}{C{2cm}C{1.5cm}rrrrrr}
  \toprule
  & \multicolumn{1}{c}{} & \multicolumn{6}{c}{Encoding function}     \\ \cline{3-8} 
  Decoding function   & Loss function                & \multicolumn{1}{l}{U} & \multicolumn{1}{l}{J} & \multicolumn{1}{l}{B1JDJ} & \multicolumn{1}{l}{B2JDJ} & \multicolumn{1}{l}{HEXJ} & \multicolumn{1}{l}{HAD} \\ \midrule
  BEL-J/BEL-U & BCE & 3.47 & 3.16 & - & - & - & - \\\hline
  GEN-EX & BCE & 3.46 & \textbf{3.12} & 3.30 & 3.35 & 3.80 & 5.75 \\\hline
  GEN & BCE & 3.49 & 3.14 & 3.62 & 3.78 & 4.44 & 5.83 \\\hline
  GEN-EX & CE & 3.82 & 3.91 & 3.52 & 3.49 & 3.98 & 3.98 \\\hline
  GEN & CE & 3.92 & 4.09 & 3.62 & 3.65 & 4.35 & 4.28 \\\hline
  GEN-EX & L2 & 3.72 & 3.60 & 4.31 & 4.29 & 6.61 & 18.69 \\ \bottomrule        
  \end{tabular}
  \end{table}

\begin{table}[h]
\centering
\scriptsize 
\caption{Comparison of BEL design parameters on MAE for head pose estimation with 300LP/AFLW2000 datasets and RAFA-Net feature extractor (HPE4).}
\label{tab:edl1rafanet}
\begin{tabular}{C{2cm}C{1.5cm}rrrrrr}
\toprule
& \multicolumn{1}{c}{} & \multicolumn{6}{c}{Encoding function}     \\ \cline{3-8} 
Decoding function   & Loss function                & \multicolumn{1}{l}{U} & \multicolumn{1}{l}{J} & \multicolumn{1}{l}{B1JDJ} & \multicolumn{1}{l}{B2JDJ} & \multicolumn{1}{l}{HEXJ} & \multicolumn{1}{l}{HAD} \\ \midrule
BEL-J/BEL-U & BCE & 3.94 & 4.00 & - & - & - & - \\\hline
GEN-EX & BCE & \textbf{3.90} & 3.93 & 4.19 & 4.12 & 4.39 & 9.17 \\\hline
GEN & BCE & 3.93 & 3.94 & 4.21 & 4.25 & 4.53 & 9.21 \\\hline
GEN-EX & CE & 4.55 & 4.62 & 4.34 & 4.53 & 4.45 & 5.12 \\\hline
GEN & CE & 4.68 & 4.75 & 4.46 & 4.61 & 4.63 & 5.29 \\\hline
GEN-EX & L2 & 4.45 & 5.87 & 5.11 & 9.34 & 10.43 & 17.89 \\
\bottomrule        
\end{tabular}
\end{table}

%%%%
%%% EDL for COFW

\begin{table}[h]
\centering
\scriptsize 
\caption{Comparison of BEL design parameters on NME for facial landmark detection with COFW dataset and HRNetV2-W18 feature extractor (FLD1).}
\label{tab:edlcofw}
\begin{tabular}{C{2cm}C{1.5cm}rrrrrr}
\toprule
& \multicolumn{1}{c}{} & \multicolumn{6}{c}{Encoding function}     \\ \cline{3-8} 
Decoding function   & Loss function                & \multicolumn{1}{l}{U} & \multicolumn{1}{l}{J} & \multicolumn{1}{l}{B1JDJ} & \multicolumn{1}{l}{B2JDJ} & \multicolumn{1}{l}{HEXJ} & \multicolumn{1}{l}{HAD} \\ \midrule
BEL-J/BEL-U & BCE & 3.47 & 3.45 & \multicolumn{1}{l}{-} & \multicolumn{1}{l}{-} & \multicolumn{1}{l}{-} & \multicolumn{1}{l}{-} \\ \hline
GEN-EX & BCE & 3.45 & 3.43 & 3.42 & 3.41 & 3.47 & 4.28 \\ \hline
GEN & BCE & 3.46 & 3.45 & 3.43 & 3.47 & 3.66 & 4.43 \\ \hline
GEN-EX & CE & 3.37 & 3.37 & 3.38 & 3.41 & \textbf{3.34} & 3.69 \\ \hline
GEN & CE & 3.44 & 3.44 & 3.44 & 3.49 & 3.57 & 3.69 \\ \hline
GEN-EX & L1 & 3.44 & 3.41 & 3.45 & 3.47 & 3.41 & 4.52   \\ \bottomrule        
\end{tabular}
\end{table}

\begin{table}[h]
\centering
\scriptsize 
\caption{Comparison of BEL design parameters on NME for facial landmark detection with 300W dataset and HRNetV2-W18 feature extractor (FLD2).}
\label{tab:edl300w}
\begin{tabular}{C{2cm}C{1.5cm}rrrrrr}
\toprule
& \multicolumn{1}{c}{} & \multicolumn{6}{c}{Encoding function}     \\ \cline{3-8} 
Decoding function   & Loss function                & \multicolumn{1}{l}{U} & \multicolumn{1}{l}{J} & \multicolumn{1}{l}{B1JDJ} & \multicolumn{1}{l}{B2JDJ} & \multicolumn{1}{l}{HEXJ} & \multicolumn{1}{l}{HAD} \\ \midrule
BEL-J/BEL-U & BCE & 3.5 & 3.49 & - & - & - & - \\ \hline
GEN-EX & BCE & 3.48 & 3.46 & 3.43 & 3.42 & 3.38 & 4.71 \\ \hline
GEN & BCE & 3.50 & 3.49 & 3.45 & 3.45 & 3.55 & 4.78 \\ \hline
GEN-EX & CE & 3.40 & \textbf{3.36} & 3.37 & 3.41 & 3.37 & 3.62 \\ \hline
GEN & CE & 3.50 & 3.45 & 3.45 & 3.51 & 3.59 & 3.65 \\ \hline
GEN-EX & L1 & 3.41 & 3.39 & 3.49 & 3.67 & 3.43 & 4.04 \\ \bottomrule        
\end{tabular}
\end{table}
\begin{table}[h]
\centering
\scriptsize 
\caption{Comparison of BEL design parameters on NME for facial landmark detection with WFLW dataset and HRNetV2-W18 feature extractor (FLD3).}
\label{tab:edlwflw}
\begin{tabular}{C{2cm}C{1.5cm}rrrrrr}
\toprule
& \multicolumn{1}{c}{} & \multicolumn{6}{c}{Encoding function}     \\ \cline{3-8} 
Decoding function   & Loss function                & \multicolumn{1}{l}{U} & \multicolumn{1}{l}{J} & \multicolumn{1}{l}{B1JDJ} & \multicolumn{1}{l}{B2JDJ} & \multicolumn{1}{l}{HEXJ} & \multicolumn{1}{l}{HAD} \\ \midrule
BEL-J/BEL-U & BCE & 4.62 & 4.54 & - & - & - & - \\ \hline
GEN-EX & BCE & 4.6 & 4.51 & 4.43 & 4.38 & 4.37 & 7.18 \\ \hline
GEN & BCE & 4.62 & 4.53 & 4.44 & 4.42 & 4.55 & 7.14 \\ \hline
GEN-EX & CE & 4.36 & 4.34 & 4.36 & \textbf{4.33} & 4.34 & 5.15 \\ \hline
GEN & CE & 4.46 & 4.44 & 4.47 & 4.47 & 4.56 & 4.83 \\ \hline
GEN-EX & L1 & 4.39 & 4.42 & 4.47 & 4.47 & 4.45 & 4.74\\ \bottomrule        
\end{tabular}
\end{table}

\begin{table}[h]
\centering
\scriptsize 
\caption{Comparison of BEL design parameters on NME for facial landmark detection with AFLW dataset and HRNetV2-W18 feature extractor (FLD4).}
\label{tab:edlaflw}
\begin{tabular}{C{2cm}C{1.5cm}rrrrrr}
\toprule
& \multicolumn{1}{c}{} & \multicolumn{6}{c}{Encoding function}     \\ \cline{3-8} 
Decoding function   & Loss function                & \multicolumn{1}{l}{U} & \multicolumn{1}{l}{J} & \multicolumn{1}{l}{B1JDJ} & \multicolumn{1}{l}{B2JDJ} & \multicolumn{1}{l}{HEXJ} & \multicolumn{1}{l}{HAD} \\ \midrule
BEL-J/BEL-U & BCE & 1.51 & 1.52 & - & - & - & - \\ \hline
GEN-EX & BCE & 1.50 & 1.50 & \textbf{1.47} & \textbf{1.47} & 1.49 & 1.52 \\ \hline
GEN & BCE & 1.51 & 1.52 & 1.50 & 1.49 & 1.54 & 1.55 \\ \hline
GEN-EX & CE & 1.48 & \textbf{1.47} & \textbf{1.47} & \textbf{1.47} & \textbf{1.47} & \textbf{1.47} \\ \hline
GEN & CE & 1.52 & 1.51 & 1.51 & 1.51 & 1.52 & 1.52 \\ \hline
GEN-EX & L1 & \textbf{1.47} & \textbf{1.47} & 1.48 & 1.48 & 1.48 & 1.59 \\ \bottomrule        
\end{tabular}
\end{table}

\begin{table}[h]
  \centering
  \scriptsize 
  \caption{Comparison of BEL design parameters on MAE for age estimation with MORPH-II dataset and ResNet50 feature extractor (AE1).}
  \label{tab:edlmorph}
  \begin{tabular}{C{2cm}C{1.5cm}rrrrrr}
  \toprule
& \multicolumn{1}{c}{} & \multicolumn{6}{c}{Encoding function}     \\ \cline{3-8} 
  Decoding function   & Loss function                & \multicolumn{1}{l}{U} & \multicolumn{1}{l}{J} & \multicolumn{1}{l}{B1JDJ} & \multicolumn{1}{l}{B2JDJ} & \multicolumn{1}{l}{HEXJ} & \multicolumn{1}{l}{HAD} \\ \midrule
  BEL-J/BEL-U & BCE & 2.32 & \textbf{2.27} & - & - & - & - \\\hline
  GEN-EX & BCE & 2.30 & 2.29 & 2.35 & 2.49 & 2.45 & 2.99 \\\hline
  GEN & BCE & 2.28 & 2.28 & 2.34 & 2.51 & 2.54 & 3.07 \\\hline
  GEN-EX & CE & 2.55 & 2.54 & 2.75 & 2.65 & 2.63 & 12.33 \\\hline
  GEN & CE & 2.60 & 2.58 & 2.61 & 2.66 & 2.61 & 3.10 \\\hline
  GEN-EX & L1 & 2.30 & 2.30 & 2.32 & 2.30 & 2.32 & 2.29 \\
  \bottomrule        
  \end{tabular}
  \end{table}

\begin{table}[h]
\centering
\scriptsize 
\caption{Comparison of BEL design parameters on MAE for age estimation with AFAD dataset and ResNet50 feature extractor (AE2).}
\label{tab:edlafad}
\begin{tabular}{C{2cm}C{1.5cm}rrrrrr}
\toprule
& \multicolumn{1}{c}{} & \multicolumn{6}{c}{Encoding function}     \\ \cline{3-8} 
Decoding function   & Loss function                & \multicolumn{1}{l}{U} & \multicolumn{1}{l}{J} & \multicolumn{1}{l}{B1JDJ} & \multicolumn{1}{l}{B2JDJ} & \multicolumn{1}{l}{HEXJ} & \multicolumn{1}{l}{HAD} \\ \midrule
BEL-J/BEL-U & BCE & 3.13 & 3.15 & - & - & - & - \\\hline
GEN-EX & BCE & 3.14 & 3.16 & 3.32 & 3.35 & 3.28 & 3.34 \\\hline
GEN & BCE & 3.13 & 3.19 & 3.41 & 3.44 & 3.41 & 3.52 \\\hline
GEN-EX & CE & 3.26 & 3.29 & 3.38 & 3.44 & 3.40 & 3.30 \\\hline
GEN & CE & 3.36 & 3.34 & 3.42 & 3.47 & 3.40 & 3.45 \\\hline
GEN-EX & L1 & 3.13 & 3.12 & \textbf{3.11} & 3.12 & 3.13 & 3.13 \\
\bottomrule        
\end{tabular}
\end{table}
    
\begin{table}[h]
\centering
\scriptsize 
\caption{Comparison of BEL design parameters on MAE for end-to-end autonomous driving with PilotNet dataset and feature extractor (PN).}
\label{tab:edlpilot}
\begin{tabular}{C{2cm}C{1.5cm}rrrrrr}
\toprule
& \multicolumn{1}{c}{} & \multicolumn{6}{c}{Encoding function}     \\ \cline{3-8} 
Decoding function   & Loss function                & \multicolumn{1}{l}{U} & \multicolumn{1}{l}{J} & \multicolumn{1}{l}{B1JDJ} & \multicolumn{1}{l}{B2JDJ} & \multicolumn{1}{l}{HEXJ} & \multicolumn{1}{l}{HAD} \\ \midrule
BEL-J/BEL-U & BCE & 4.34 & 3.91 & - & - & - & - \\ \hline
GEN-EX & BCE & 4.57 & 4.20 & 4.83 & 4.96 & 5.29 & 10.12 \\ \hline
GEN & BCE & 4.37 & 3.95 & 3.51 & 3.61 & 4.01 & 10.00 \\ \hline
GEN-EX & CE & 4.30 & 4.16 & 4.99 & 5.87 & 5.39 & 87.17 \\ \hline
GEN & CE & 3.15 & \textbf{3.11} & 3.14 & 3.21 & 3.64 & 6.20 \\ \hline
GEN-EX & L1 & 4.10 & 4.11 & 4.34 & 4.34 & 4.11 & 5.09 \\
\bottomrule        
\end{tabular}
\end{table}

\FloatBarrier
\paragraph{Impact of quantization and  decoding functions: } 

As discussed in Section~3, a real-valued label is quantized to a discrete value in $\{1,2,...,N\}$ before applying the encoding function. In Table~\ref{tab:quanr}, we show the effect of increasing the number of quantization levels ($N$) on the error for correlation-based decoding ($\mathcal{D}^{\text{GEN}}$, which returns a quantized prediction) and expected correlation-based decoding ($\mathcal{D}^{\text{GEN-EX}}$, which returns a continuous prediction). 
%We vary the number of quantization levels from 64 to 256 and observe that 
As shown in the table, there exists a tradeoff between reducing quantization error and using fewer classifiers. The error is lower for 128 quantization levels than it is for 256 as the improvement resulting from fewer binary classifiers is higher than the increase in quantization error. Moreover, the use of proposed decoding function $\mathcal{D}^{\text{GEN-EX}}$ for regression consistently results in lower error compared to $\mathcal{D}^{\text{GEN}}$.  

  \begin{table}[h]
    \centering
    \scriptsize 
    \caption{Impact of the quantization and decoding functions on NME for facial landmark detection.}
    \label{tab:quanr}
    \begin{tabular}{lccccccc}
    \toprule
     & \multicolumn{3}{c}{COFW} & & \multicolumn{3}{c}{300W} \\ \hline 
     Quantization levels &  64 & 128 & 256&  &  64 & 128 & 256\\ \midrule
     $\mathcal{E}^{\text{BEL-U}}$ + $\mathcal{D}^{\text{GEN}}$ &3.66 & 3.51& 3.46 && 3.79& 3.59& 3.46\\ \hline
     $\mathcal{E}^{\text{BEL-U}}$ + $\mathcal{D}^{\text{GEN-EX}}$ & 3.46 &3.41 & 3.44 && 3.54& 3.47& 3.44\\ \hline
     $\mathcal{E}^{\text{BEL-J}}$ + $\mathcal{D}^{\text{GEN}}$ &  3.65 & 3.49 & 3.43&& 3.76 & 3.58 & 3.46 \\ \hline
     $\mathcal{E}^{\text{BEL-J}}$ + $\mathcal{D}^{\text{GEN-EX}}$ & 3.45 &\textbf{3.40} & 3.42 && 3.52& 3.45 & \textbf{3.43} \\ 
  \bottomrule
    \end{tabular}%
    \end{table}
  \paragraph{Impact of the number of training samples on BEL:}
As discussed in Section~4, the performance of different encoding functions varies depending on the availability of sufficient training data. In Table~\ref{tab:datasetdrop}, we analyze the effect of the number of available training samples for both simple and complex encodings. We use the number of bit transitions as a measure of the complexity of a classifier.  As the number of training samples decreases, simpler encodings (U and J) perform better than more complex encodings (B1JDJ, B2JDJ, and HEXJ). Using a more complex encoding reduces the number of classifiers; however, it increases each classifier's complexity (i.e. the number of \db{s}) and thus performs poorly with less training data.

  \begin{table}[h]
    \centering
    \scriptsize 
    \caption{Effect of training dataset size on optimal encoding function for facial landmark detection. BCE loss function and GEN-EX decoding function are used for the training and evaluation. }
    \label{tab:datasetdrop}
    \begin{tabular}{lL{2cm}L{2cm}rrrrrrr}
      \toprule
     &   &  & \multicolumn{7}{c}{Reduction in the number training samples} \\ \cline{4-10} 
    Encoding & \#Classifiers/label & \#\db{s}/classifier & 0\% & 20\% & 40\% & 60\% & 80\% & 90\% & 95\% \\\midrule
    \multicolumn{10}{c}{COFW (FLD1)} \\ \hline
    U & 256 & 1 & 3.45 & 3.48 & 3.55 & 3.72 & 3.94 & 4.52 & 6.29 \\ \hline
    J & 128 & 2 & 3.43 & 3.48 & 3.51 & 3.61 & \textbf{3.88} & \textbf{4.32} & \textbf{5.39} \\ \hline
    B1JDJ & 65 & 4 & {3.42} & \textbf{3.44} & 3.52 & \textbf{3.60} & 4.11 & 4.50 & 5.68 \\ \hline
    B2JDJ & 34 & 8 & \textbf{3.41} & 3.45 & \textbf{3.48} & 3.80 & 3.94 & 4.80 & 6.56 \\ \hline
    HEXJ & 17 & 32 & 3.47 & 3.69 & 3.78 & 4.03 & 4.61 & 5.48 & 6.69 \\ \midrule
    \multicolumn{10}{c}{300W (FLD2)} \\ \hline
    U & 256 & 1 & 3.48 & 3.55 & 3.58 & 3.64 & 3.89 & 4.26 & 5.66 \\ \hline
    J & 128 & 2 & 3.46 & 3.56 & 3.52 & 3.58 & \textbf{3.79} & \textbf{4.04} & \textbf{4.58} \\ \hline
    B1JDJ & 65 & 4 & 3.43 & 3.48 & 3.53 & 3.61 & 3.89 & 4.31 & 6.10 \\ \hline
    B2JDJ & 34 & 8 & {3.42} & \textbf{3.47} & \textbf{3.51} & \textbf{3.54} & 3.88 & 4.50 & 5.80 \\ \hline
    HEXJ & 17 & 32 & \textbf{3.38} & 3.64 & 3.73 & 3.97 & 4.41 & 5.38 & 6.60 \\ \midrule
    \multicolumn{10}{c}{WFLW (FLD3)} \\ \hline
    U & 256 & 1 & 4.60 & 4.67 & 4.83 & 5.00 & 5.37 & 6.04 & 7.46 \\ \hline
    J & 128 & 2 & 4.51 & 4.60 & 4.65 & 4.84 & 5.23 & \textbf{5.64} & \textbf{6.39} \\ \hline
    B1JDJ & 65 & 4 & 4.43 & \textbf{4.44} & 4.52 & 4.66 & 5.08 & 5.90 & 8.39 \\ \hline
    B2JDJ & 34 & 8 & {4.38} & 4.46 & \textbf{4.49} & \textbf{4.61} & \textbf{5.02} & 5.95 & 8.78 \\ \hline
    HEXJ & 17 & 32 & \textbf{4.37} & 4.60 & 4.72 & 4.96 & 5.72 & 6.86 & 8.09 \\ \midrule
    \multicolumn{10}{c}{AFLW (FLD4)} \\ \hline
    U & 256 & 1 & 1.50 & 1.53 & 1.53 & 1.56 & 1.61 & 1.68 & 1.83 \\ \hline
    J & 128 & 2 & 1.50 & 1.51 & 1.52 & 1.54 & 1.60 & 1.68 & 1.79 \\ \hline
    B1JDJ & 65 & 4 & \textbf{1.47} & \textbf{1.50} & 1.52 & 1.54 & 1.60 & 1.67 & 1.78 \\ \hline
    B2JDJ & 34 & 8 & \textbf{1.47} & \textbf{1.50} & \textbf{1.50} & \textbf{1.52} & \textbf{1.57} & \textbf{1.64} & \textbf{1.73} \\ \hline
    HEXJ & 17 & 32 & 1.49 & 1.54 & 1.54 & 1.55 & 1.59 & 1.71 & 1.89 \\\bottomrule
    \end{tabular}
    \end{table}
    \FloatBarrier
\paragraph{Impact of reflected binary conversion: } 
As mentioned in Section~3, we use reflected binary to increase the distance between distant labels based on the design properties of suitable regression encodings we proposed. 
  Table~\ref{tab:gray} shows the impact of using reflected binary conversion on error for facial landmark detection benchmarks. 
  As shown in the table, the use of reflected binary significantly reduces the error.
  \begin{table}[h]
    \caption{Effect of reflected binary conversion for B1JDJ encoding on facial landmark detection. Here, BCE loss and GEN-EX decoding functions are used.}
    \label{tab:gray}
    \centering
    \scriptsize 
    \begin{tabular}{lrrrr}
      \toprule
     & \multicolumn{1}{l}{COFW} & \multicolumn{1}{l}{300W} & \multicolumn{1}{l}{WFLW} & \multicolumn{1}{l}{AFLW} \\ \midrule
    B1JDJ & 3.43 & 3.46 & 4.43 & 1.47 \\ \hline
    B1JDJ- w/o reflected binary & 4.13 & 4.43 & 5.70 & 1.97 \\ \bottomrule
  \end{tabular}
\end{table}
\paragraph{Use of binary heamaps: } 
Facial landmark detection approaches typically use heatmap regression. 
We also evaluate \BEL-H-x, in which the real-valued heatmaps are converted to binary heatmaps with $8$ quantization levels.
Table~\ref{tab:heatmap} shows the impact of using binary heatmaps on error for facial landmark detection benchmarks. For unary code, a $64\times64$ real-valued heatmap of one facial landmark is converted to eight $64 \times 64$ binary heatmaps, resulting in $32,768$ ($8 \times 64 \times 64$) binary classifiers compared to $512$ for BEL-U. We believe that training a high number of binary classifiers results in high error for BEL-H-x. 
  \begin{table}[h]
    \caption{Comparison of BEL with heatmaps for facial landmark detection. Here, BCE loss and GEN-EX decoding functions are used.}
    \label{tab:heatmap}
    \centering
    \scriptsize 
    \begin{tabular}{lrrrr}
      \toprule
     & \multicolumn{1}{l}{FLD1} & \multicolumn{1}{l}{FLD2} & \multicolumn{1}{l}{FLD3} & \multicolumn{1}{l}{FLD4} \\ \midrule
    BEL-U & 3.45 & 3.46 & 4.60 & 1.50\\ \hline
    BEL-J & 3.48 & 3.46 & 4.51 & 1.50 \\ \hline
    BEL-H-U &  4.13  & 4.43   & 5.70  & 1.97 \\ \hline
    BEL-H-J &  10.17 & 33.02 & 22.50 &  2.99  \\ \bottomrule
  \end{tabular}
\end{table}
\paragraph{Hyperparameter $\theta$:}
As shown in Figure~4b, we introduce a feature vector of size $\theta$ before the output layer. Figure~\ref{fig:bitdep} compares the decrease in the error for different encodings and $\theta$ values.  
 We observe that more complex encodings benefit more from an increase in the value of $\theta$, while a lower value of $\theta$ can be used for simpler encodings. 
\begin{figure}[h]
\centering
  \includegraphics[width=0.5\textwidth]{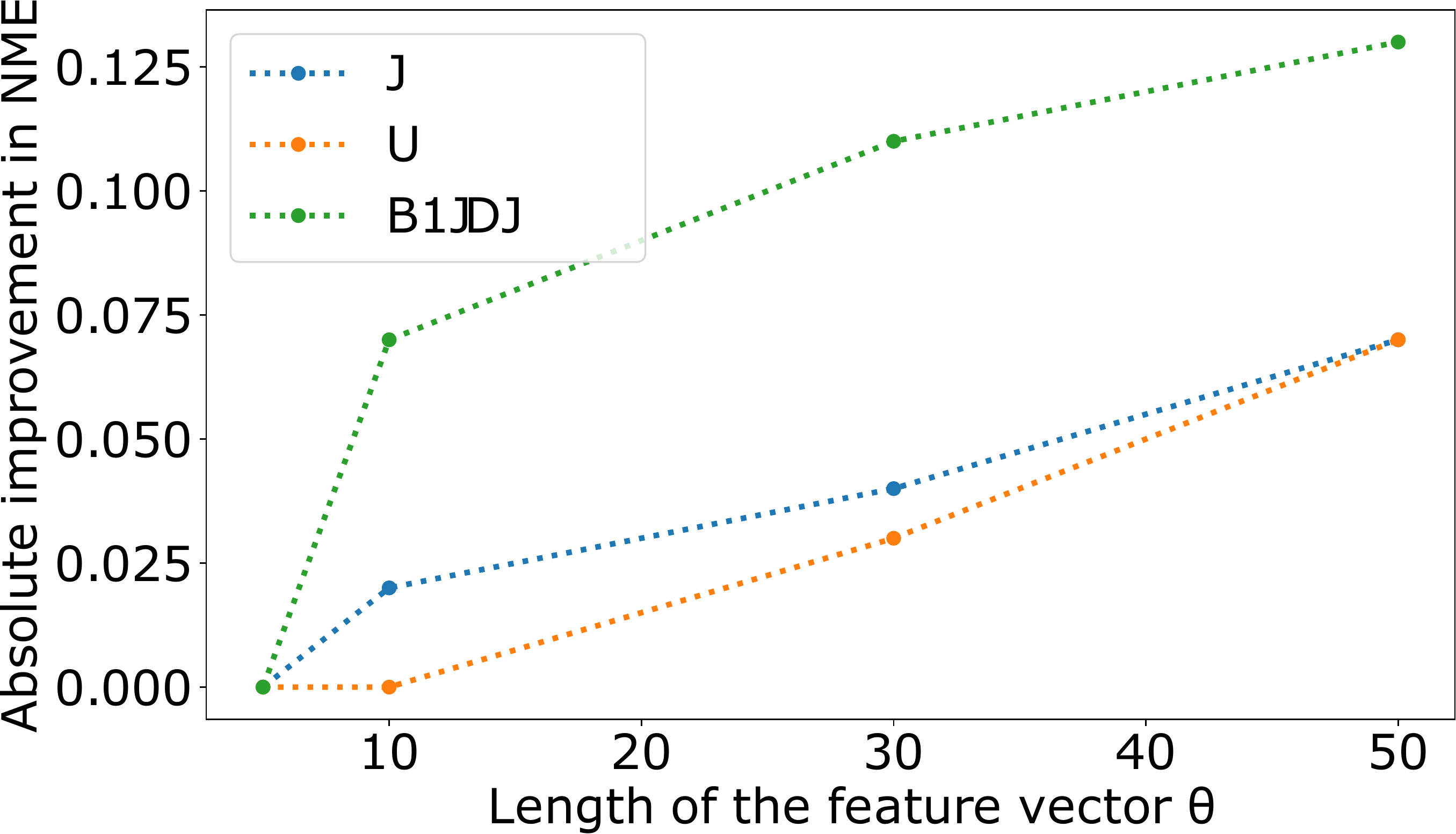}
  \caption{Effect of $\theta$ on error for different encodings on FLD1.}
  \label{fig:bitdep}
\end{figure} 
% \usepackage{multirow}
\paragraph{Impact of increasing the number of fully connected layers:}
\begin{table}[]
    \caption{Impact increasing number of fully connected layers in direct regression and multiclass classification on the error (MAE or NME).}
    \label{tab:FCL1}
    \centering
    \scriptsize
    \begin{tabular}{lrrrrr}
        \toprule
    \multirow{2}{*}{Benchmark} & \multicolumn{2}{c}{Direct regression} & \multicolumn{2}{c}{Multiclass classification} & \multicolumn{1}{c}{BEL} \\ \cline{2-6}
     & \multicolumn{1}{c}{1 FC layer} & \multicolumn{1}{c}{2 FC layers} & \multicolumn{1}{c}{1 FC layer} & \multicolumn{1}{c}{2 FC layers} & \multicolumn{1}{c}{2 FC layers} \\ \midrule
    HPE1 & 4.76 & 5.19 & 4.49 & 4.82 & \textbf{3.37} \\ \hline
    HPE2 & 5.65 & 5.59 & 5.31 & 5.42 & \textbf{4.77} \\ \hline
    HPE3 & 3.40 & 3.54 & 4.45 & 4.54 & \textbf{3.12} \\ \hline
    HPE4 & 4.14 & 4.22 & 5.14 & 5.45 & \textbf{3.90} \\ \hline
    FLD1 & 3.60 & 3.63 & {3.58} & 3.56 & \textbf{3.34} \\ \hline
    FLD2 & 3.54 & 3.58 & 3.51 & 3.62 & \textbf{3.36} \\ \hline
    FLD3 & 4.64 & 4.63 & 4.50 & 4.64 & \textbf{4.33} \\ \hline
    FLD4 & 1.51 & 1.51 & 1.56 & 1.53 & \textbf{1.47} \\ \hline
    AE1 & 2.44 &  2.35 & 2.75 &  2.81 & \textbf{2.27} \\ \hline
    AE2 & 3.21 & 3.14  & 3.38 & 3.40  & \textbf{3.11} \\ \hline
    PN & 4.24 &  4.33 & 4.56 &  5.74 & \textbf{3.11} \\ \bottomrule
    \end{tabular}
    \end{table}
For BEL, we propose to add a fully connected bottleneck layer in the regressor to reduce the feature vector size to $\theta$ and thus decrease the number of parameters in the regressor. 
We perform an ablation study to study the impact of this added fully connected layer on relative performance of direct regression, multiclass classification, and binary encoded labels. 
Table~\ref{tab:FCL1} provides the error (MAE or NME) for direct regression and multiclass classification with one or two fully connected layers after the feature extractor. 
Further, we evaluate BEL, direct regression, and multiclass classification for higher number of fully connected layers as shown in Table~\ref{tab:FCL2}. 
We observe that increasing the number of fully connected layers in direct regression and multiclass classification does not improve the accuracy for most benchmarks (possibly due to overparameterization). 
BEL with two fully connected layers outperforms direct regression and multiclass classification in both cases. 
Furthermore, even for a higher number of fully connected layers in BEL, the suitability of an encoding function varies with the dataset, demonstrating the importance of BEL design space. 

% \usepackage{multirow}
\begin{table}[h]
    \caption{Impact increasing number of fully connected layers in direct regression, multiclass classification, and BEL. GEN-EX decoding function and BCE loss function are used for BEL. }
    \label{tab:FCL2}
    \scriptsize
    \centering
    \begin{tabular}{lL{2.5cm}R{1cm}R{1.5cm}R{0.8cm}R{0.8cm}R{0.8cm}R{0.8cm}R{0.8cm}}
        \toprule
    Benchmark & {\# FC layers (size of FC layers)} & Direct regression & Multiclass classification & U&  {J} & {B1JDJ} & {B2JDJ} & {HEXJ} \\ \midrule
    \multirow{3}{*}{FLD1} & 1 (1024-x) & 3.6 & 3.58 & - & - & - & -& - \\ \cline{2-9}
     & 2 (1024-30-x) & 3.63 & 3.56 & 3.45 & 3.43 & 3.42 & \textbf{3.41} & 3.47 \\ \cline{2-9}
     & 3 (1024-30-10-x) & 3.63 & 3.94 & 3.55 & \textbf{3.47} & 3.82 & 4.02 & 3.62 \\ \hline
    \multirow{3}{*}{FLD2} & 1 (1024-x) & 3.54 & 3.51 & -& -& - & - & - \\ \cline{2-9}
     & 2 (1024-10-x) & 3.58 & 3.62 & 3.48 & 3.46 & 3.43 & 3.42 & \textbf{3.38} \\ \cline{2-9}
     & 3 (1024-30-10-x) & 3.55 & {3.78} & \textbf{3.42} & 3.46 & 3.5 & 3.61 & 3.52 \\ \bottomrule
    \end{tabular}
    \end{table}

\paragraph{Training-validation set based evaluation: }
Ideally, a validation set should be used for model selection. Hence we have reevaluated the benchmarks with a validation set to select the best design parameters and the best model (i.e., which model is the best over multiple epochs). 
Since datasets used in benchmarks do not provide separate validation datasets, we use $20\%$ of the training data as a validation set. Since earlier works use $100\%$ training data for the reported results and use test error for model selection, we have re-run specialized approaches (if possible), direct regression, and multiclass classification. It was not possible for us to re-run experiments for all specialized approaches due to resource constraints, and the comparison is conservative for many benchmarks. 

Table~\ref{tab:compare_regression_80} compares different regression approaches for this evaluation setup. Note that the additional results do not diminish the effectiveness of BEL and BEL outperforms direct regression and multiclass classification for all benchmarks and specialized approaches for several benchmarks. 
\begin{table}[h]
    %\centering
    \setlength\tabcolsep{4pt}
    \caption{Comparison of BEL with different regression approaches. Specialized approaches for each benchmark are described in Table~1 } 
    \label{tab:compare_regression_80}
    \scriptsize 
    \begin{tabular}{L{3.3cm}L{2.35cm}L{2.35cm}L{2.35cm}L{2.35cm}}
    \toprule
    & \multicolumn{4}{c}{ Error (MAE or NME) / Model size }   \\ \hline
    Approach (\% training set)& \multicolumn{1}{c}{HPE1}        & \multicolumn{1}{c}{HPE2}        & \multicolumn{1}{c}{HPE3}        & \multicolumn{1}{c}{HPE4}         \\ \midrule
    Specialized approach (100\%)& -           & - &  \textbf{3.40}  / 69.8M & 4.14 / 69.8M  \\ \hline
    Specialized approach (80\%) & -           & - &  4.08$\pm$0.11  / 69.8M & 4.69$\pm$0.02 / 69.8M  \\ \hline
    Direct regression (80\%)           & 6.12$\pm$0.02 / 23.5M & 5.97$\pm$0.09 / 23.5M &   4.08$\pm$0.11 / 69.8M &   4.67+4.70 / 69.8M  \\ \hline
    Multiclass classification (80\%)   &  5.38$\pm$0.03 / 24.2M & 5.60$\pm$0.13 / 24.8M & 5.58$\pm$0.04 / 72.0M & 5.86$\pm$0.10 / 72.0M \\ \hline
    BEL (80\%)        &  \textbf{3.91}$\pm$0.08 / 23.6M &  \textbf{4.91}$\pm$0.10 / 23.6M &  {3.50}$\pm$0.08 / 69.8M & \textbf{3.99}$\pm$0.04 / 69.8M  \\ \hline     
    BEL $\mathcal{E}/\mathcal{D}/\mathcal{L}$ functions &U/GEN-EX/L2 &  U/GEN-EX/BCE&  B1JDJ/GEN-EX/BCE&  U/GEN-EX/BCE \\ \bottomrule
    \end{tabular}
    \begin{tabular}{L{3.3cm}L{2.35cm}L{2.35cm}L{2.35cm}L{2.35cm}}
      \toprule
    Approach (\%training set) & \multicolumn{1}{c}{FLD1}  & \multicolumn{1}{c}{FLD2} &  \multicolumn{1}{c}{FLD3}    & \multicolumn{1}{c}{FLD4}     \\ \midrule
    %Specialized approach &  3.45 / { }{ }9.6M  &   \textbf{3.32} / { }{ }9.6M &  \textbf{4.32} / { }{ }9.6M &  1.57 / { }{ }9.6M  \\ \hline
  Specialized approach  (100\%) &  3.45 / { }{ }9.6M  &   \textbf{3.32} / { }{ }9.6M &  \textbf{4.32} / { }{ }9.6M &  1.57 / { }{ }9.6M  \\ \hline
  Direct regression  (80\%)           & 3.70$\pm$0.04 / 10.2M  & 3.69$\pm$0.06 / 10.2M &    4.71$\pm$0.02 / 10.2M &  1.51$\pm$0.01 / 10.2M  \\ \hline
  Multiclass classification  (80\%)    & 3.64$\pm$0.02 / 25.4M &  3.68$\pm$0.02 / 45.2M &  4.77$\pm$0.02 / 61.3M & 1.56 {$\pm$0.01} / 20.1M  \\ \hline
  BEL  (80\%)                         &  \textbf{3.35}$\pm$0.02 / 10.6M &  3.40$\pm$0.03 / 11.2M &   4.37$\pm$0.01 / 11.7M &  \textbf{1.48}$\pm$0.01 / 10.8M \\ \hline  
  BEL $\mathcal{E}/\mathcal{D}/\mathcal{L}$ functions  & HEXJ/GEN-EX/CE & U/GEN-EX/CE & B1JDJ/GEN-EX/CE & B1JDJ/GEN-EX/CE \\ \bottomrule   
  \end{tabular}
  \begin{tabular}{L{3.3cm}L{2.35cm}L{2.35cm}L{2.35cm}L{2.35cm}}
  \toprule
  Approach (\% training set) &  \multicolumn{1}{c}{AE1}   & \multicolumn{1}{c}{AE2}   & \multicolumn{1}{c}{PN}   &      \\ \midrule
  %Specialized approach &  2.49 / 21.3M &  3.47 / 21.3M &  4.24 / 1.8M& \\ \hline
  Specialized approach (100\%) &  2.49 / 21.3M &  3.47 / 21.3M &  4.24 / 1.8M& \\ \hline
  Direct regression  (80\%)          & 2.45 {$\pm$0.01} / 23.1M & 3.34 {$\pm$0.02} / 23.1M & 4.56 {$\pm$0.45} / 1.8M& \\ \hline
  Multiclass classification  (80\%)    & 2.85 {$\pm$0.03} / 23.1M & 3.47 {$\pm$0.05} / 23.1M & 6.37 {$\pm$0.00} / 1.9M & \\ \hline
  BEL  (80\%)                        & \textbf{2.36} {$\pm$0.01} / 23.1M & \textbf{3.20} {$\pm$0.00} / 23.1M &  \textbf{3.49} {$\pm$0.01} / 1.8M& \\ \hline
  BEL $\mathcal{E}/\mathcal{D}/\mathcal{L}$ functions   &   J/BEL-J/BCE   &  B1JDJ/GEN-EX/L1 & J/GEN/CE    \\ \bottomrule
  \end{tabular}
  \end{table}
\section{Expected Error Derivation}
\label{sec:a1}
This section explains the expected error equations used to compare BEL-U and BEL-J in Section~3. We first explain the encoding and decoding function used for BEL-U and derive the relation between the expected regression error and classification error for BEL-U. Then, we explain the encoding/decoding functions and expected error relation for BEL-J. 
\subsection{Preliminaries}

%We compose binary-encoded labels as a quantization function $q$, an encoding function $\mathcal{E}$, and a decoding function $\mathcal{D}$. 
Given a sample $i$ drawn from a dataset with minimum label $a$ and maximum label $b$, let $y_i \in [a, b]$ represent the target label for that sample. Assuming uniform quantization, the range of target labels can be quantized using $q: [a, b] \to \{1, 2, ..., N\}$ through \Eqref{eq:quantization}.
\begin{equation}
  \label{eq:quantization}
  q(y_i) = (y_i - a) * \frac{N-1}{b-a}  + 1
\end{equation}
We define the encoding function $\mathcal{E}: \{1, 2, ..., N-1\} \to \{0, 1\}^M$ to convert a target quantized level $Q_i \in \{1,2,...,N-1\}$ to a binary code $B_i \in \{0, 1\}^M$. We further define the decoding function $\mathcal{D}:\{0, 1\}^M \to [a, b]$ to convert the predicted binary code $\hat{B_i}$ to the predicted label $\hat{y_i}$.

Although the decoding functions used in this analysis predict the quantized label and introduce quantization error, we do not include quantization error in the expected absolute error for our analysis as it is constant for both BEL-U and BEL-J. The expected value of absolute error between the target $y$ and predicted labels $\hat{y}$ is used for the analysis as typically mean absolute error is used as the evaluation metric in regression problems. 

Let us denote the error probability of a binary classifier $C^k$ used to predict bit $k$ in a binary code $B_i = \mathcal{E}(n)$ as $e_k (n)$, where $n$ is the target quantized label $Q_i$. Then,
\begin{equation}
  \label{eq:err}
  \begin{split}
  e_k(n) &= \EX(|\hat{b}_i^k-b_i^k|) \\
  &= Pr(\hat{b}_i^k=F) \text{ for target label $Q_i = n$ and target binary code $B_i =\mathcal{E}(n) $}
  \end{split}
\end{equation}
where $\hat{b}_i^k=T$ indicates a correct binary classification by classifier $C^k$ ($\hat{b}_i^k==b_i^k$) for sample $i$ and $\hat{b}_i^k=F$ indicates an incorrect binary classification by classifier $C^k$ ($\hat{b}_i^k \neq b_i^k$) for sample $i$. 

\subsection{Expected Error for BEL-U}
%ORBC is based on the existing approach for ordinal regression by binary classification. %We use the reduction rule proposed and used by the earlier approaches~\cite{ordext}. 
\paragraph{Encoding and decoding functions:}
The encoding and decoding functions for BEL-U are defined as:
\begin{equation}
  \label{eq:orb1}
  \mathcal{E}^{\text{BEL-U}}(Q_i) = b_i^1, b_i^2,..,b_i^{N-2} \text{, where } b_i^k  = \begin{cases}
    1, & k  < Q_i\\
    0, & \text{Otherwise}\\
  \end{cases}    
  \end{equation}
  %\begin{equation}
  %  \label{eq:6}
    \begin{equation}
      \label{eq:orb2}
    \mathcal{D}^{\text{BEL-U}}(\hat{b}_i^1,\hat{b}_i^2,..., \hat{b}_i^{N-2}) =  \sum_{k=1}^{N-2} \hat{b}_i^k + 1
    \end{equation}

\paragraph{Expected error:}
For target quantized label $Q_i = n$ ($n \in \{1,2,...,N-1\}$), ignoring the quantization error, the expected error  between target $y_i$ and predicted label $\hat{y}_i$ can be derived as:
\begin{equation}
  \label{eq:orb3_p}
  \begin{split}
  \EX(|\hat{y}_i^{\text{BEL-U}}-y_i|) & = \EX\Big(\Big|\sum_{k=1}^{N-2} \big( \hat{b}_i^k + 1 \big)-\sum_{k=1}^{N-2} \big( b_i^k + 1\big)\Big|\Big)\\
  & = \EX\Big(\Big|\sum_{k=1}^{N-2} (\hat{b}_i^k-b_i^k)\Big|\Big)\\
  & \leqslant \EX\Big(\sum_{k=1}^{N-2} |\hat{b}_i^k-b_i^k|\Big)\\
  &= \sum_{k=1}^{N-2} \EX|\hat{b}_i^k-b_i^k|\\
  &= \sum_{k=1}^{N-2} e_k(n) \text{ (using ~\Eqref{eq:err}) }
  \end{split}
\end{equation} 
For a uniform distribution of target labels in the range $[1,N-1]$, the expected error can be derived as:
\begin{equation}
    \label{eq:orb3}
    \EX(|\hat{y}^{\text{BEL-U}}-y|)  \leqslant \frac{1}{N-1} \sum_{n=1}^{N-1} \sum_{k=1}^{N-2} e_k(n) 
  \end{equation} 
\subsection{Expected Error for BEL-J}
\paragraph{Encoding and decoding functions:}

For target quantized label $Q_i \in \{1,2,...,N-1\}$, BEL-J encoding requires $\frac{N}{2}$ bits/binary classifiers. The encoding for BEL-J can be defined as:\\
\begin{equation}
\label{eq:5}
\begin{aligned}
\mathcal{E}^{\text{BEL-J}}(Q_i) = b_i^1, b_i^2,..,b_i^{\frac{N}{2}} \text{,where } b_i^k  = \begin{cases}
  1, &  \frac{N}{2}-Q_i < k \leqslant N-Q_i\\
  0, & \text{Otherwise}
\end{cases} 
\end{aligned}
\end{equation}
\begin{figure}[b!]
    \centering
      \includegraphics[width=0.45\textwidth]{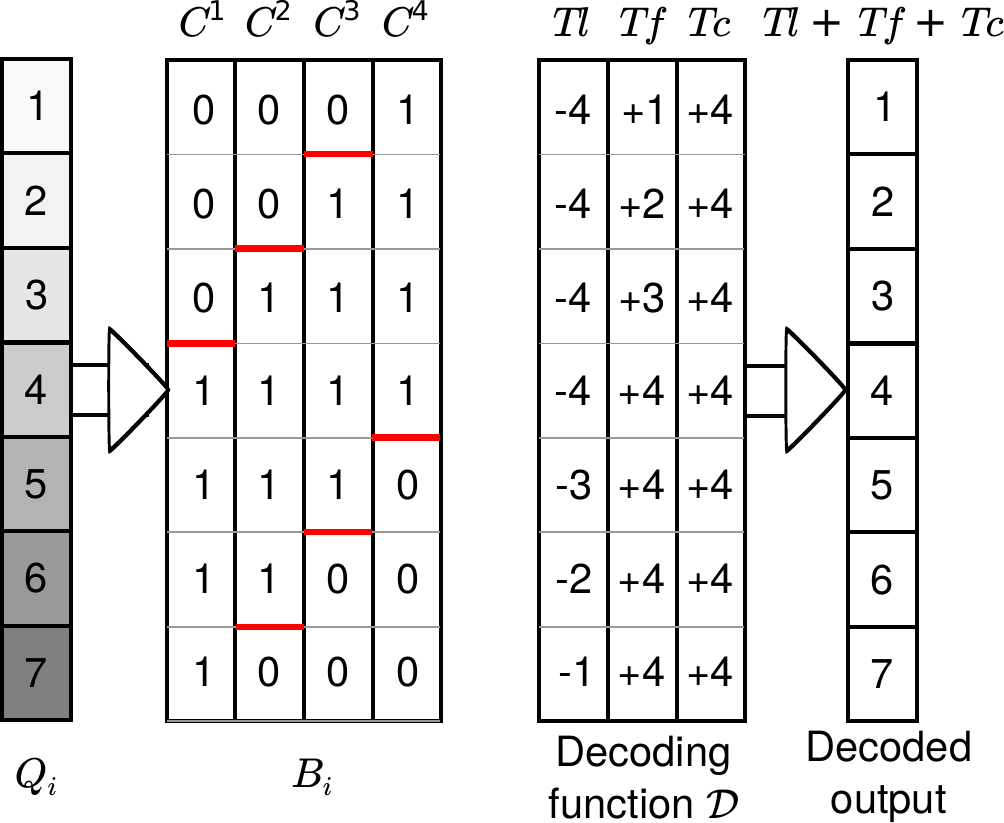}
   \caption{Encoding and Decoding functions' output for BEL-J approach and label $y \in [1,N-1], \text{where } N=8$. Decoding function's output is calculated using $y' =  Tl + Tf + Tc$, where $Tl= -  \max_{k\in\{1...\frac{N}{2}\}} k  \hat{b}_i^k$ , $Tf = \max_{k\in\{1...\frac{N}{2}\}} \Big( \frac{N}{2} - k + 1\Big) \hat{b}_i^k$, and $Tc =  \frac{N}{2} $}.
   \label{fig:belj7}
  \end{figure}

Similarly, the decoding functions for BEL-J can be defined as:
%\begin{equation}
%  \label{eq:6}
  \begin{equation}
    \label{eq:6}
    \begin{aligned}
  \mathcal{D}^{\text{BEL-J}}(\hat{B}_i) =    Tl(\hat{B}_i) + Tf(\hat{B}_i) + Tc
  \\
  \text{where, } Tl(\hat{B}_i) =  - \max_{k\in\{1...\frac{N}{2}\}} k  \hat{b}_i^k \\
  Tf(\hat{B}_i) = \max_{k\in\{1...\frac{N}{2}\}} \Big( \frac{N}{2} - k + 1\Big) \hat{b}_i^k,  Tc =  \frac{N}{2}
    \end{aligned}
  \end{equation}
  In~\Eqref{eq:6}, $Tl()$ finds the location of the last occurrence of ``$1$'' in the predicted binary code $\hat{B}_i$. Similarly, $Tf()$ finds the location of the first occurrence of ``$1$'' in the binary code $\hat{B}_i$. Figure~\ref{fig:belj7} gives examples of binary codes for label  $Q_i \in \{1,2,...,7\}$ and the corresponding values of the different terms in ~\Eqref{eq:6}. For example, for label $Q_i=3$, the binary code is ``$0 1 1 1$''. Here, the last occurrence of  ``$1$''  is at position $4$, and $Tl =-4$. Similarly, the first occurrence of ``$1$'' is at position $2$, and $Tf = (4+1) - 2 = 3$.  
\paragraph{Expected error:}
For BEL-J code, binary classifiers $(C^1,C^2,...,C^{\frac{N}{2}})$ are used.  
For a given input sample i, an error in any of the binary classifiers' outputs $(\hat{b}_i^1,\hat{b}_i^2,..., \hat{b}^{\frac{N}{2}})$ will result in an error between $Tf(\hat{B}_i)/Tl(\hat{B}_i)$ and $Tf(B_i)/Tl(B_i)$ in~\Eqref{eq:6}. 
We refer to $Tf(\hat{B}_i)$ and $Tl(\hat{B}_i)$ as $\hat{Tf}_i$ and $\hat{Tl}_i$ (predicted binary code), and  $Tf(B_i)$ and $Tl(B_i)$ as $Tf_i$ and $Tl_i$ (target binary code) for brevity. 
Expected value of the absolute error can be further expanded as:
\begin{equation}
    \label{eq:9}
    \begin{split}
    \EX(|\hat{y}_i^{\text{BEL-J}}-y_i|) & = \EX(|\hat{Tf}_i + \hat{Tl}_i +Tc - (Tf_i + Tl_i +Tc) |)\\
    & = \EX(|(\hat{Tf}_i- Tf_i) + (\hat{Tl}_i  - Tl_i ) |)\\
    & \leqslant \EX(|\hat{Tf}_i- Tf_i|+ |\hat{Tl}_i- Tl_i|)\\
    &= \EX(|\hat{Tf}_i- Tf_i|) + \EX(|\hat{Tl}_i- Tl_i|) 
    \end{split}
\end{equation} 
%Here, subscript $n$ represents the target label, i.e., $y_n=n$. 
Thus, the sum of the expected error of $Tf()$ and $Tl()$ is the upper bound of the label's expected error. Further, we derive the relation between binary classifiers' error probabilities and $\EX(|\hat{Tf}_i- Tf_i|)$ and $\EX(|\hat{Tl}_i- Tl_i|)$. 
%Further, we derive the expected errors of $Tf$ and $Tl$. 

We consider $Q_i = n, \text{where } 1\leqslant n \leqslant \frac{N}{2}$ for our derivation. In such a case, $Tf_i =n$ and $Tl_i = -\frac{N}{2}$. However, as the code is symmetric around $Q_i =\frac{N}{2}$, it can be shown that the derived equation for $\EX|\hat{y}_i-y_i|$ can be used for $1 \leqslant Q_i \leqslant N-1$.

\textbf{1. Derivation of $\EX|\hat{Tf}_i-Tf_i|$:}
\begin{figure}[t]
  \centering
    \includegraphics[width=0.6\textwidth]{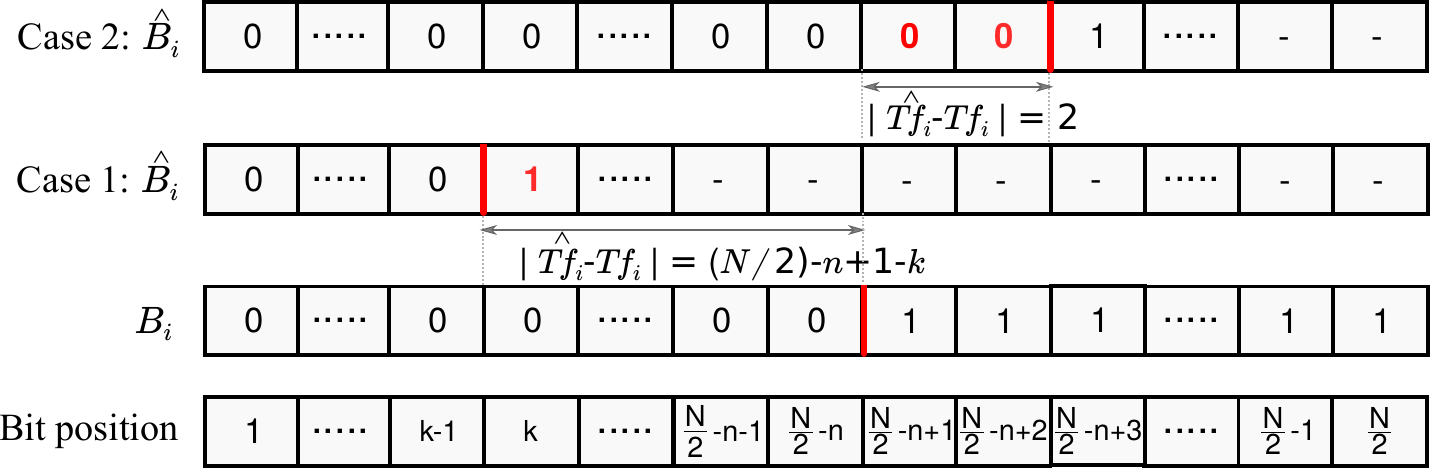}
 \caption{Effect of classifier error on $\hat{Tf}_i-Tf_i$ for label $Q_i=n$. Case 1 and case 2 represent erroneous outputs. $0/1$ highlighted in red color represents an error in the classifier's output. ``-'' represents error/no error in both cases.  }
 \label{fig:beltf}
\end{figure} 
As shown in ~\Eqref{eq:6}, $Tf()$ finds the location $k$ of the first occurrence of ``$1$'' in the binary sequence.
% and $Tf = \frac{N}{2}-k+1$. Thus for a binary code for label n, the first occurrence of $1$ would be at position $k = \frac{N}{2}-n +1$. 
In the case of an erroneous binary sequence, the position of the first occurrence of ``$1$'' might shift, which results in an error between $\hat{Tf}_i$ and $Tf_i$.
Figure~\ref{fig:beltf} shows examples of the correct and erroneous outputs of classifiers for label $Q_i=n$. For label $Q_i=n$, $b_i^k=0$ for $k \in \{1,2,...,\frac{N}{2}-n\}$ and $b_i^k=1$ for $k \in \{\frac{N}{2}-n+1,\frac{N}{2}-n+2,...,\frac{N}{2}\}$.  

For case~1, error in a classifier $ C^k, k\in \{1,2,...,\frac{N}{2}-n\}$ is considered, where $b_i^k=0$ and $\hat{b}_i^k=1$. 
For $k \in \{1,2,...,\frac{N}{2}-n\}$, an error at classifier $C^k$ will result in erroneous $\hat{Tf}_i$ only if all proceeding classifiers are correct, since if any of the proceeding classifier $z$ is incorrect, i.e. $\hat{b}_i^z=1$, then the location of the first occurrence of ``$1$'' will be shifted to $z$, and any error in the following classifiers will not affect the value of $\hat{Tf}$. Such a case ($\hat{b}_i^1=T,\hat{b}_i^2=T,...,\hat{b}_i^{k-1}=T,\hat{b}_i^k=F,\hat{b}_i^{k+1}=T/F,...,\hat{b}_i^{\frac{N}{2}}=T/F$) considers a total of $2^{\frac{N}{2}-k}$ combinations out of $2^{\frac{N}{2}}$ for $k\in \{1,2,...,\frac{N}{2}-n\}$. Assuming that the binary classifiers are mutually independent, the error value and the probability of this combination can be shown to be: 
\begin{equation}
    \label{eq:tf1}
    |\hat{Tf}_i-Tf_i| =   \Big(  \frac{N}{2} -n + 1 -k\Big) 
\end{equation}
\begin{equation}
  \label{eq:tf2}
  \begin{split}
Pr(\hat{b}_i^1=T,\hat{b}_i^2=T,...,\hat{b}_i^{k-1}=T,\hat{b}_i^k=F)  & = Pr(\hat{b}^1=T) Pr( \hat{b}_i^2=T)...Pr(\hat{b}_i^{k-1}=T ) Pr(\hat{b}^k=F) \\
    & =\Big( \prod_{j=1}^{k-1} (1-e_j(n)) \Big) \cdot e_k(n) 
    \end{split}
\end{equation}

The above term considers combinations ($b'^1=T,b'^2=T,...,b'^{k-1}=T,b'^k=F,b'^{k+1}=T/F,...,b'^{\frac{N}{2}=T/F}$) for $k\in \{1,2,...,\frac{N}{2}-n\}$, which constitutes to a total of $\sum_{k=1}^{\frac{N}{2}-n} 2^{\frac{N}{2}-k}$ combinations out of $2^{\frac{N}{2}}$. 
\\
\\

For case~2, error in a classifier $ C^k, k\in \{\frac{N}{2}-n+1,\frac{N}{2}-n+2,...,\frac{N}{2}\}$ is considered, where $b_i^k=1$ and $\hat{b}_i^k=0$. 
We consider a combination ($\hat{b}_i^1=T,\hat{b}_i^2=T,...,\hat{b}_i^{\frac{N}{2}-n}=T,\hat{b}_i^{\frac{N}{2}-n+1}=F,...,\hat{b}_i^{k-1}=F,\hat{b}_i^k=T,\hat{b}_i^{k+1}=T/F,...,\hat{b}_i^{\frac{N}{2}=T/F}$). For this case, the position of the first occurrence of ``$1$'' will be moved to $k$, which will result in erroneous $\hat{Tf}_i$. Such a case would cover $2^{\frac{N}{2}-k}$ combinations out of $2^{\frac{N}{2}}$ for $k\in \{\frac{N}{2}-n+1,\frac{N}{2}-n+2,...,\frac{N}{2}\}$. The error value and the probability of this combination can be shown to be: 
\begin{equation}
  \label{eq:tf3}
  |\hat{Tf}_i-Tf_i| =   \Big( k - (\frac{N}{2} - n + 1 ) \Big) 
\end{equation}
\begin{equation}
\label{eq:tf4}
\begin{split}
  Pr(\hat{b}_i^1=T,\hat{b}_i^2=T,...,\hat{b}_i^{\frac{N}{2}-n}=T,\hat{b}_i^{\frac{N}{2}-n+1}=F,...,\hat{b}_i^{k-1}=F,\hat{b}_i^k=T) = \\
   \Big( \prod_{j=1}^{\frac{N}{2}-n} (1-e_j(n)) \Big) \cdot \Big( \prod_{j=\frac{N}{2}-n+1}^{k-1} e_j(n) \Big) \cdot \Big(1-e_k(n)\Big) 
\end{split}
\end{equation}
The above term considers combinations $(\hat{b}_i^1=T,\hat{b}_i^2=T,...,\hat{b}_i^{\frac{N}{2}-n}=T,\hat{b}_i^{\frac{N}{2}-n+1}=F,...,\hat{b}_i^{k-1}=F,\hat{b}_i^k=T,\hat{b}_i^{k+1}=T/F,...,\hat{b}_i^{\frac{N}{2}=T/F})$, which constitutes to a total of $\sum_{k=\frac{N}{2}-n+1}^{\frac{N}{2}} 2^{\frac{N}{2}-k}$ combinations out of $2^{\frac{N}{2}}$ for $k\in \{\frac{N}{2}-n+1,\frac{N}{2}-n+2,...,\frac{N}{2}\}$. 
%Error at classifier $k$ will result in erroneous $Tf$ only if all the classifiers in ${1,2,...,\frac{N}{2}-n}$ are correct and all the classifiers in ${\frac{N}{2}-n+1,\frac{N}{2}-n+2,...,k}$ are incorrect. We consider 
%As we move from classifier~$\frac{N}{2}-n+1$ to classifier~$k$, each erroneous classifier contributes to error$=1$ to $Tf$. 

Combining~\Eqref{eq:tf1} to \Eqref{eq:tf4}, the expected value of $|\hat{Tf}_i-Tf_i|$ can be derived as:
\begin{equation}
    \label{eq:tf5}
    \begin{aligned}
    \EX(|\hat{Tf}_i-Tf_i|) = & \sum_{k=1}^{\frac{N}{2}-n} \Big(\frac{N}{2}-n+1-k\Big) \cdot \Big(\prod_{j=1}^{k-1}(1-e_j(n))\Big) \cdot e_k(n)  \\ & + \sum_{k=\frac{N}{2}-n+1}^{\frac{N}{2}} \Big( k - (\frac{N}{2} - n + 1 ) \Big) \cdot \Big( \prod_{j=1}^{\frac{N}{2}-n} (1-e_j(n)) \Big) \cdot  \Big( \prod_{j=\frac{N}{2}-n+1}^{k-1} (e_j(n)) \Big) \cdot \Big(1-e_k(n)\Big) \\
     = & \sum_{k=1}^{\frac{N}{2}-n} \Big(\frac{N}{2}-n+1-k\Big) \cdot e_k(n) \cdot \Big(\prod_{j=1}^{k-1}(1-e_j(n))\Big) + \sum_{k=\frac{N}{2}-n+1}^{\frac{N}{2}} \Big( \prod_{j=\frac{N}{2}-n+1}^{k} e_j(n) \Big) 
    \end{aligned}
\end{equation}
The first term in~\Eqref{eq:tf5} covers $\sum_{k=1}^{\frac{N}{2}-n} 2^{\frac{N}{2}-k}$ combinations and the second term considers $\sum_{k=\frac{N}{2}-n+1}^{\frac{N}{2}} 2^{\frac{N}{2}-k}$ combinations. Adding one combination where all the classifiers are correct, ~\Eqref{eq:tf5} considers all of the possible combinations $2^{\frac{N}{2}}$ to find expected value of $|\hat{Tf}_i-Tf_i|$.  

\textbf{2. Derivation of $\EX|\hat{Tl}_i-Tl_i|$:}
\begin{figure}[t]
  \centering
    \includegraphics[width=0.6\textwidth]{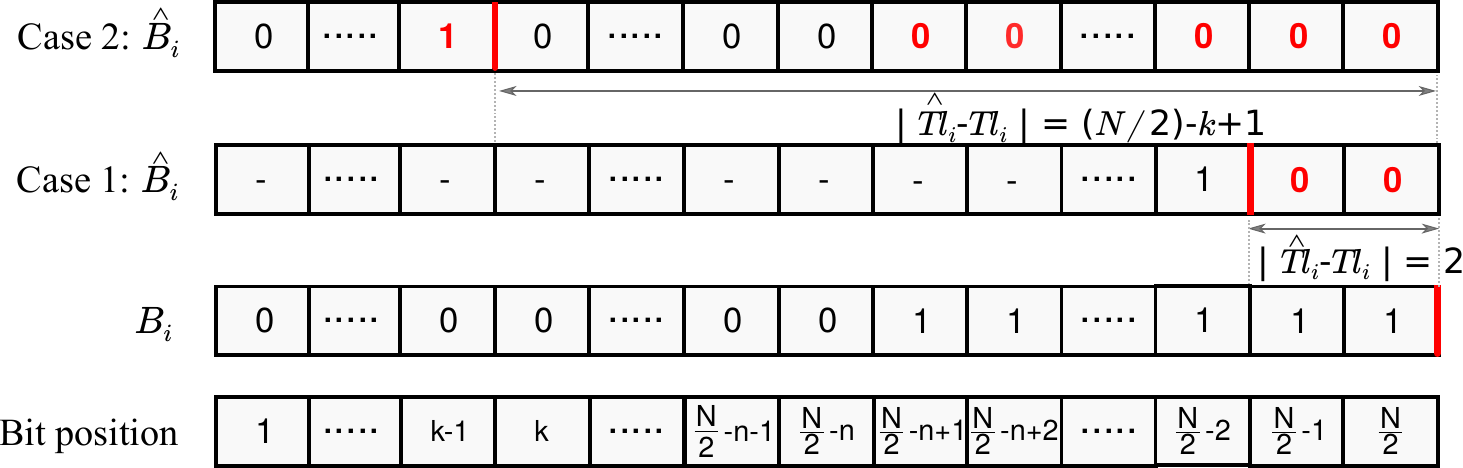}
 \caption{Effect of classifier error on $\hat{Tl}_i-Tl_i$ for label $Q_i=n$. Case 1 and case 2 represent erroneous outputs. $0/1$ highlighted in red color represents an error in the classifier's output. ``-'' represents error/no error in both cases.  }
 \label{fig:beltl}
\end{figure} 
As shown in~\Eqref{eq:6}, $Tl()$ finds the location $k$ of the last occurrence of ``$1$'' in the binary sequence.
% and $Tf = \frac{N}{2}-k+1$. Thus for a binary code for label n, the first occurrence of $1$ would be at position $k = \frac{N}{2}-n +1$. 
In the case of an erroneous binary sequence, the position of the last occurrence of ``$1$'' might shift, which results in an erroneous value of $\hat{Tl}_i$.
Figure~\ref{fig:beltl} shows examples of correct and erroneous outputs of classifiers for label $Q_i=n$. 

For case~1, an error in a classifier $ C^k, k \in \{\frac{N}{2}-n+1,\frac{N}{2}-n+2,...,\frac{N}{2}\}$ is considered, where $b_i^k=1$ and $\hat{b}_i^k=0$. 
We consider a combination $(\hat{b}_i^{\frac{N}{2}}=F,\hat{b}_i^{\frac{N}{2}-1}=F,...,\hat{b}_i^{k+1}=F,\hat{b}_i^k=T,\hat{b}_i^{k-1}=T/F,...,\hat{b}_i^1=T/F)$. For this case, position of the last occurrence of ``$1$'' will be moved to $k$, which will result in erroneous $\hat{Tl}_i$. Such a case would cover $2^{k-1}$ combinations out of $2^{\frac{N}{2}}$. The error value and the probability of this combination can be shown to be: 
\begin{equation}
  \label{eq:tl1}
  |\hat{Tl}_i-Tl_i| =   \Big( \frac{N}{2} - k \Big) 
\end{equation}
\begin{equation}
\label{eq:tl2}
\begin{aligned}
  Pr(\hat{b}_i^{\frac{N}{2}}=F,\hat{b}_i^{\frac{N}{2}-1}=F,...,\hat{b}_i^{k+1}=F,\hat{b}_i^k=T) =  \Big( \prod_{j=k+1}^{\frac{N}{2}} e_j(n) \Big) \cdot (1-e_k(n))   
\end{aligned}
\end{equation}
The above term considers combinations $(\hat{b}_i^{\frac{N}{2}}=F,\hat{b}_i^{\frac{N}{2}-1}=F,...,\hat{b}_i^{k+1}=F,\hat{b}_i^k=T,\hat{b}_i^{k-1}=T/F,...,\hat{b}_i^1=T/F)$ for $k \in \{\frac{N}{2}-n+1,\frac{N}{2}-n+2,...,\frac{N}{2}\}$, which constitutes to a total of $\sum_{k=\frac{N}{2}-n+1}^{\frac{N}{2}} 2^{k-1}$ combinations out of $2^{\frac{N}{2}}$. 
\\

For case~2, an error in a classifier $ C^k,k \in \{1,2,...,\frac{N}{2}-n\}$ is considered, where $b_i^k=0$ and $\hat{b}_i^k=1$. 
We consider a combination ($\hat{b}_i^{\frac{N}{2}}=F,...,\hat{b}_i^{\frac{N}{2}-n+1}=F,\hat{b}_i^{\frac{N}{2}-n}=T,...,\hat{b}_i^{k+1}=T,\hat{b}_i^k=F,\hat{b}_i^{k-1}=T/F,...,\hat{b}_i^{1}=T/F$). 
For this case, position of the last occurrence of ``$1$'' will be moved to $k$, which will result in erroneous $\hat{Tl}_i$. Such a case would cover $2^{k-1}$ combinations out of $2^{\frac{N}{2}}$. The error value and the probability of this combination can be shown to be: 
\begin{equation}
  \label{eq:tl3}
  |\hat{Tl}_i-Tl_i| =   \Big( \frac{N}{2} - k \Big) 
\end{equation}
\begin{equation}
\label{eq:tl4}
\begin{aligned}
  Pr(\hat{b}_i^{\frac{N}{2}}=F,...,\hat{b}_i^{\frac{N}{2}-n+1}=F,\hat{b}_i^{\frac{N}{2}-n}=T,...,\hat{b}_i^{k+1}=T,\hat{b}_i^k=F) =  \\
  \Big( \prod_{j=\frac{N}{2}-n+1}^{\frac{N}{2}} e_j(n) \Big)  \cdot \Big( \prod_{j=k+1}^{\frac{N}{2}-n} (1-e_j(n)) \Big) \cdot (e_k(n)) 
\end{aligned}
\end{equation}
The above term considers combinations $(\hat{b}_i^{\frac{N}{2}}=F,...,\hat{b}_i^{\frac{N}{2}-n+1}=F,\hat{b}_i^{\frac{N}{2}-n}=T,...,\hat{b}_i^{k+1}=T,\hat{b}_i^k=F,\hat{b}_i^{k-1}=T/F,...,\hat{b}_i^{1}=T/F)$ for $k \in \{1,2,...,\frac{N}{2}-n\}$, which constitutes to a total of $\sum_{k=1}^{\frac{N}{2}-n} 2^{k-1}$ combinations out of $2^{\frac{N}{2}}$. 
\\
\\
Combining~\Eqref{eq:tl1} to \Eqref{eq:tl4}, the expected value of $|\hat{Tl}_i-Tl_i|$ can be derived as:
\begin{equation}
    \label{eq:tl5}
    \begin{aligned}
    \EX(|\hat{Tl}_i-Tl_i|)  = & \sum_{k=\frac{N}{2}-n+1}^{\frac{N}{2}}  \Big( \frac{N}{2} - k \Big) \cdot \Big( \prod_{j=k+1}^{\frac{N}{2}} e_j(n) \Big) \cdot (1-e_k(n)) \\ & + \sum_{k=1}^{\frac{N}{2}-n} \Big( \frac{N}{2} - k  \Big) \cdot \Big( \prod_{j=\frac{N}{2}-n+1}^{\frac{N}{2}} e_j(n) \Big)  \cdot \Big( \prod_{j=k+1}^{\frac{N}{2}-n} (1-e_j(n)) \Big) \cdot (e_k(n))
    \\
     = & \sum_{k=\frac{N}{2}-n+1}^{\frac{N}{2}} \Big( \prod_{j=k}^{\frac{N}{2}} e_j(n) \Big)  
    + \Big( \prod_{j=\frac{N}{2}-n+1}^{\frac{N}{2}} e_j(n) \Big) \cdot \sum_{k=1}^{\frac{N}{2}-n}  \Big( \prod_{j=k}^{\frac{N}{2}-n} (1-e_j(n)) \Big)
    \end{aligned}
\end{equation}
The first term in~\Eqref{eq:tl5} covers $\sum_{k=\frac{N}{2}-n+1}^{\frac{N}{2}} 2^{k-1}$ combinations and the second term considers $\sum_{k=1}^{\frac{N}{2}-n} 2^{k-1}$ combinations. Adding one combination where all the classifiers are correct, ~\Eqref{eq:tl5} considers all of the possible combinations $2^{\frac{N}{2}}$ to find expected value of $|\hat{Tl}_i-Tl_i|$.  
\\
\\

Combining~\Eqref{eq:9},~\Eqref{eq:tf5}, and~\Eqref{eq:tl5}, the expected value of error for $Q_i = n$ in terms of classifiers' error probabilities can be derived as:
\begin{multline}
\label{eq:finprev}
\EX(\hat{y}_i^{\text{BEL-J}}-y_i) \leqslant \sum_{k=1}^{\frac{N}{2}-n} \Big(\frac{N}{2}-n+1-k\Big) \cdot e_k(n) \cdot \Big(\prod_{j=1}^{k-1}(1-e_j(n))\Big) + \sum_{k=\frac{N}{2}-n+1}^{\frac{N}{2}} \Big( \prod_{j=\frac{N}{2}-n+1}^{k} e_j(n) \Big)  \\
+ \sum_{k=\frac{N}{2}-n+1}^{\frac{N}{2}} \Big( \prod_{j=k}^{\frac{N}{2}} e_j(n) \Big)  
+ \Big( \prod_{j=\frac{N}{2}-n+1}^{\frac{N}{2}} e_j(n) \Big) \cdot \sum_{k=1}^{\frac{N}{2}-n}  \Big( \prod_{j=k}^{\frac{N}{2}-n} (1-e_j(n)) \Big)
\end{multline}

As the binary code is symmetric around $\frac{N}{2}$ as shown in Figure~\ref{fig:belj7}, the expected errors for label $y_i \in [1,\frac{N}{2}]$ can be mirrored to find expected errors for label $y_i \in [\frac{N}{2},N-1]$. 
For a uniform distribution of target labels in the range $[1,N-1]$, the expected error can be derived as:
\begin{multline}
  \label{eq:fin}
  \EX(\hat{y}^{\text{BEL-J}}-y) \leqslant \frac{1}{N-1} \sum_{n=1}^{N-1} \Big[ \sum_{k=1}^{\frac{N}{2}-n} \Big(\frac{N}{2}-n+1-k\Big) \cdot e_k(n) \cdot \Big(\prod_{j=1}^{k-1}(1-e_j(n))\Big) + \sum_{k=\frac{N}{2}-n+1}^{\frac{N}{2}} \Big( \prod_{j=\frac{N}{2}-n+1}^{k} e_j(n) \Big)  \\
  + \sum_{k=\frac{N}{2}-n+1}^{\frac{N}{2}} \Big( \prod_{j=k}^{\frac{N}{2}} e_j(n) \Big)  
  + \Big( \prod_{j=\frac{N}{2}-n+1}^{\frac{N}{2}} e_j(n) \Big) \cdot \sum_{k=1}^{\frac{N}{2}-n}  \Big( \prod_{j=k}^{\frac{N}{2}-n} (1-e_j(n)) \Big) \Big]
  \end{multline}

We also verify the equation by comparing the expected value of error based on ~\Eqref{eq:finprev} for $Q_i \in \{1,2,...,N-1\}$ with the expected error calculated by $100,000$ random samples of binary sequences for the same error probabilities $e_k(n)$. Figure~\ref{fig:verify} compares the expected error from~\Eqref{eq:finprev} and measured from statistical samples, and validates error upper bounds calculated using~\Eqref{eq:finprev} and~\Eqref{eq:fin}.   
\begin{figure}[h]
  \centering
    \includegraphics[width=0.6\textwidth]{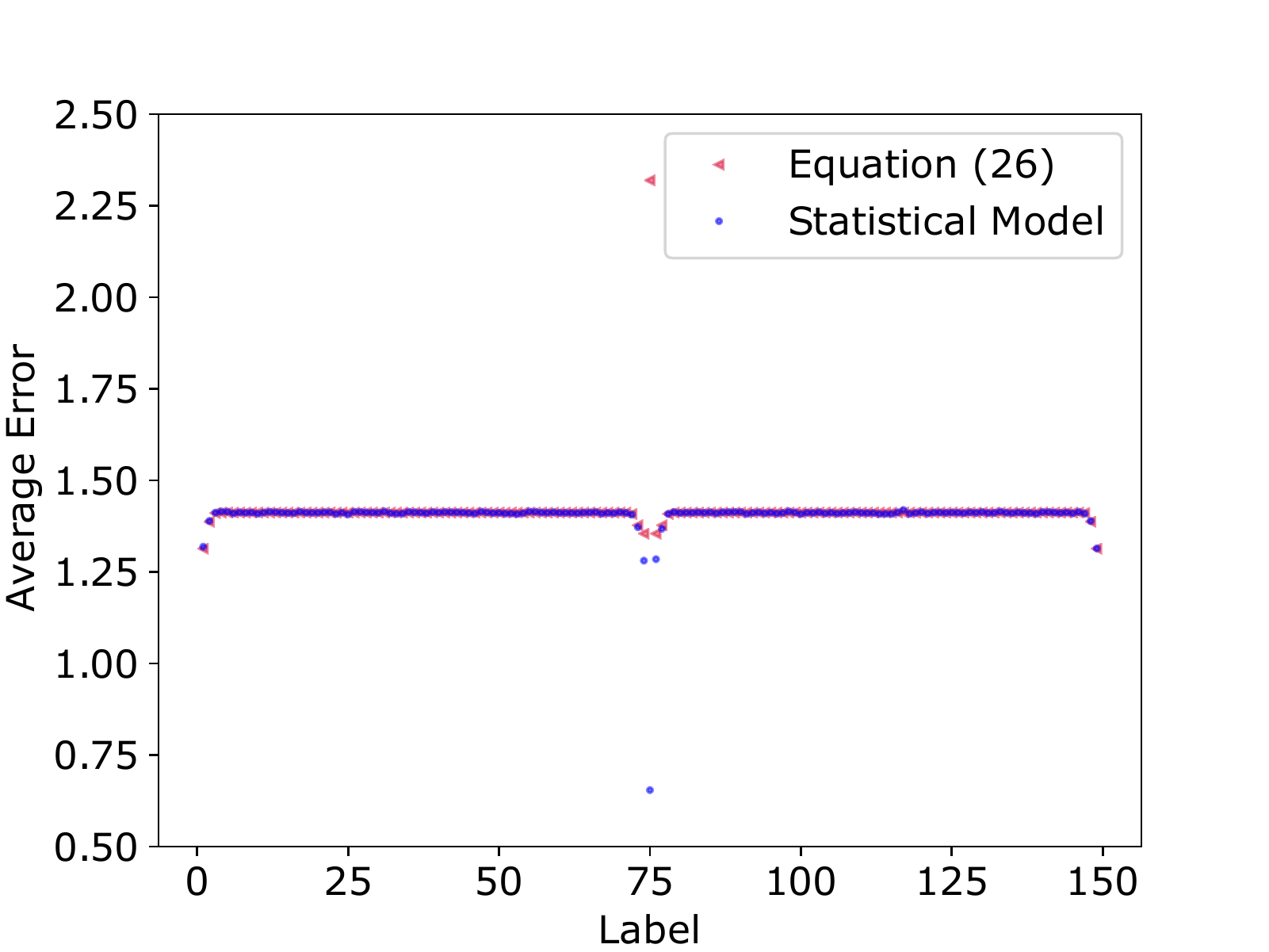}
 \caption{Comparison of expected value of error from ~\Eqref{eq:finprev} and random samples for given error probabilities of the classifiers. }
 \label{fig:verify}
\end{figure}
\pagebreak

\section{Error Probability of Classifiers}
\label{sec:a3}
It is known that the error/misclassification probability $e_{k}(n)$ of a classifier tends to increase as the target label value $n$ is closer to the classifier's decision boundaries~\citep{decision}. 
We approximate  $e_{k}(y)$ for a classifier $C^k$ with $t$ \db{s} as a linear combination of $t$ Gaussian distributions. Here, each Gaussian term is centered around a \db. 
Figure~\ref{fig:accuracy} shows the empirically observed error probability distributions for different classifiers trained for different combinations of network and dataset. We also show the approximate error probability distribution using a linear combination of Gaussian distributions. 
Here r is a scalar multiplied with probability density of gaussian distribution and $\sigma$ is the standard deviation (Equation 3 and 4). 
{
  \captionsetup[figure]{font=footnotesize , belowskip=0pt}
\begin{figure}[t]
\centering
  \begin{subfigure}[t]{0.45\linewidth} 
  \includegraphics[width=\textwidth]{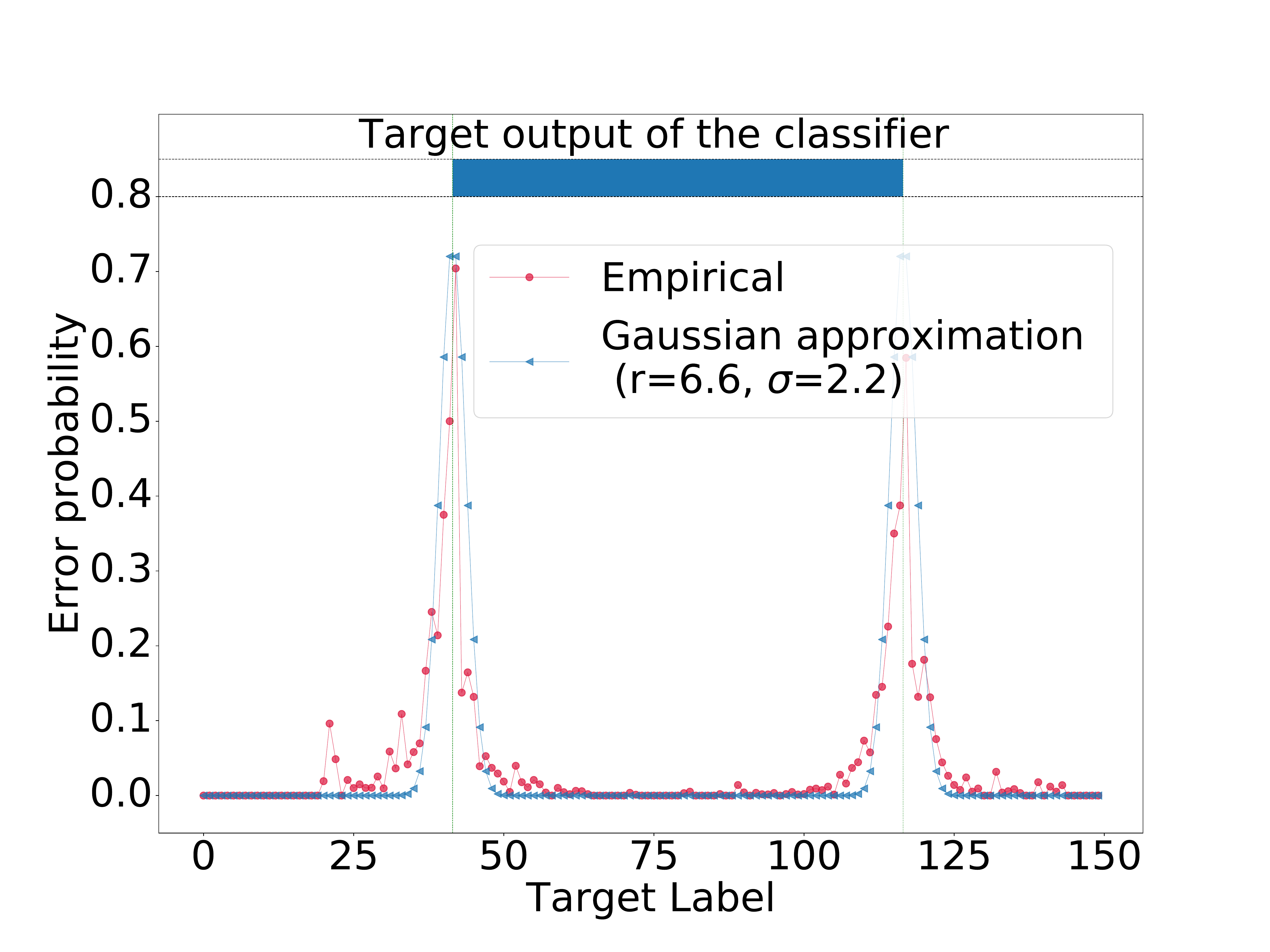}
  \caption{{ResNet50 (BIWI head pose estimation)}}
  \label{fig:a4}
  \end{subfigure}
  \begin{subfigure}[t]{0.45\linewidth}
  \centering
  \includegraphics[width=\textwidth]{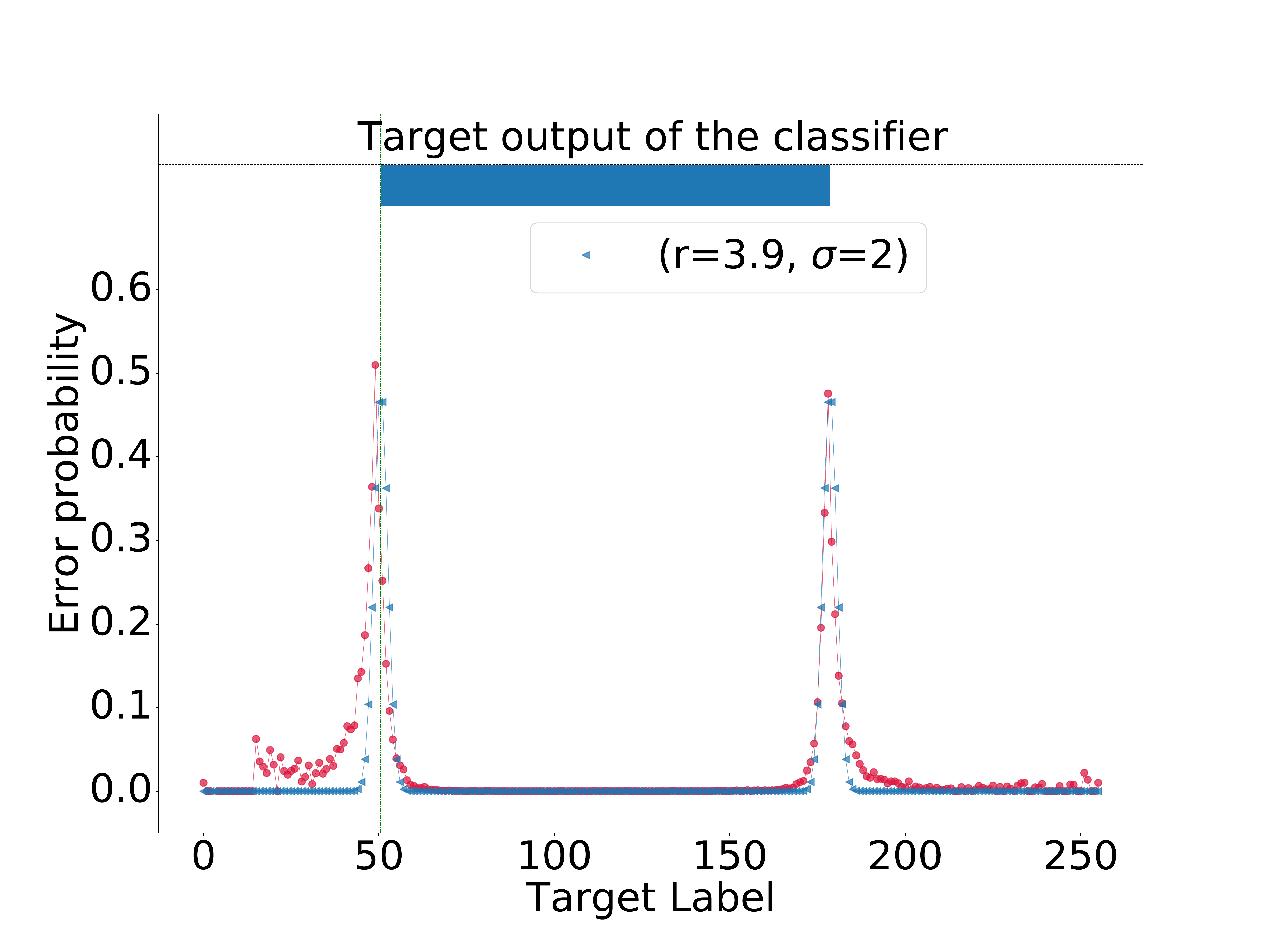}
  \caption{{HRNetV2-W18 (AFLW facial landmark detection)}}
  \label{fig:a5}
  \end{subfigure}
\\
  \begin{subfigure}[t]{0.45\linewidth} 
  \includegraphics[width=\textwidth]{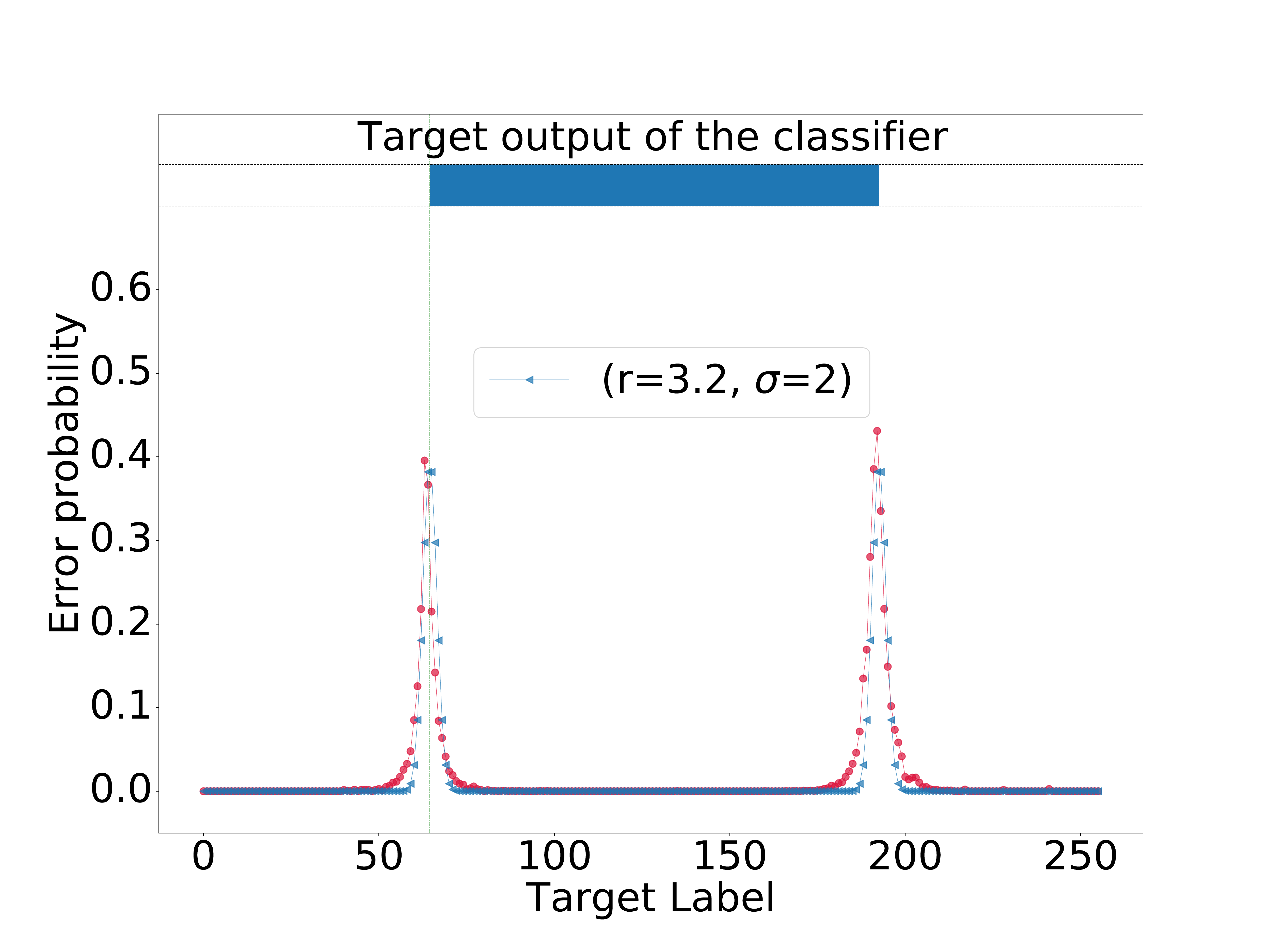}
  \caption{{HRNetV2-W18 (300W facial landmark detection)}}
  \label{fig:a6}
  \end{subfigure}
  \begin{subfigure}[t]{0.45\linewidth}
  \centering
  \includegraphics[width=\textwidth]{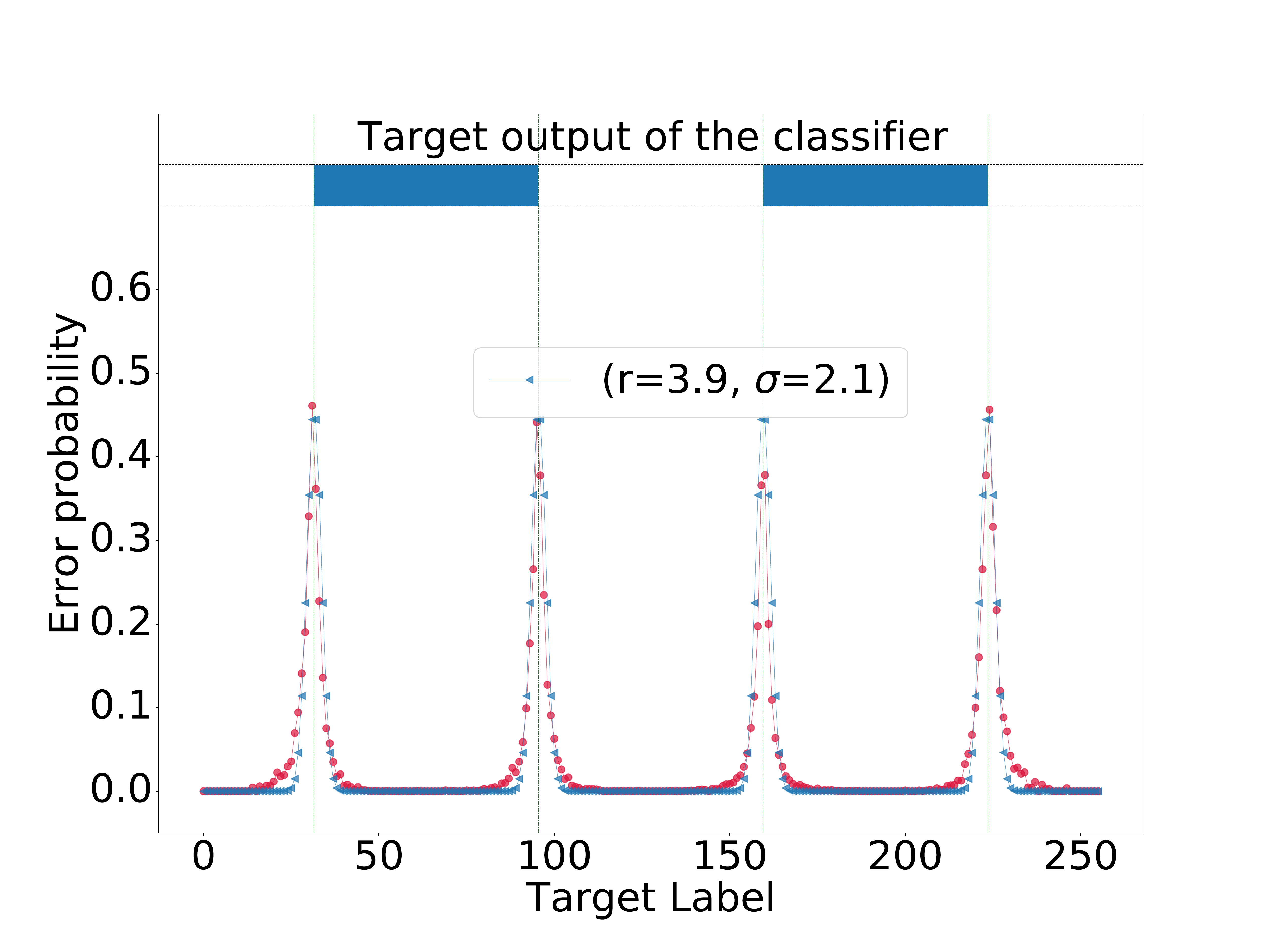}
  \caption{{HRNetV2-W18 (300W facial landmark detection)}}
  \label{fig:a7}
  \end{subfigure}
  \\
  \begin{subfigure}[t]{0.45\linewidth} 
  \includegraphics[width=\textwidth]{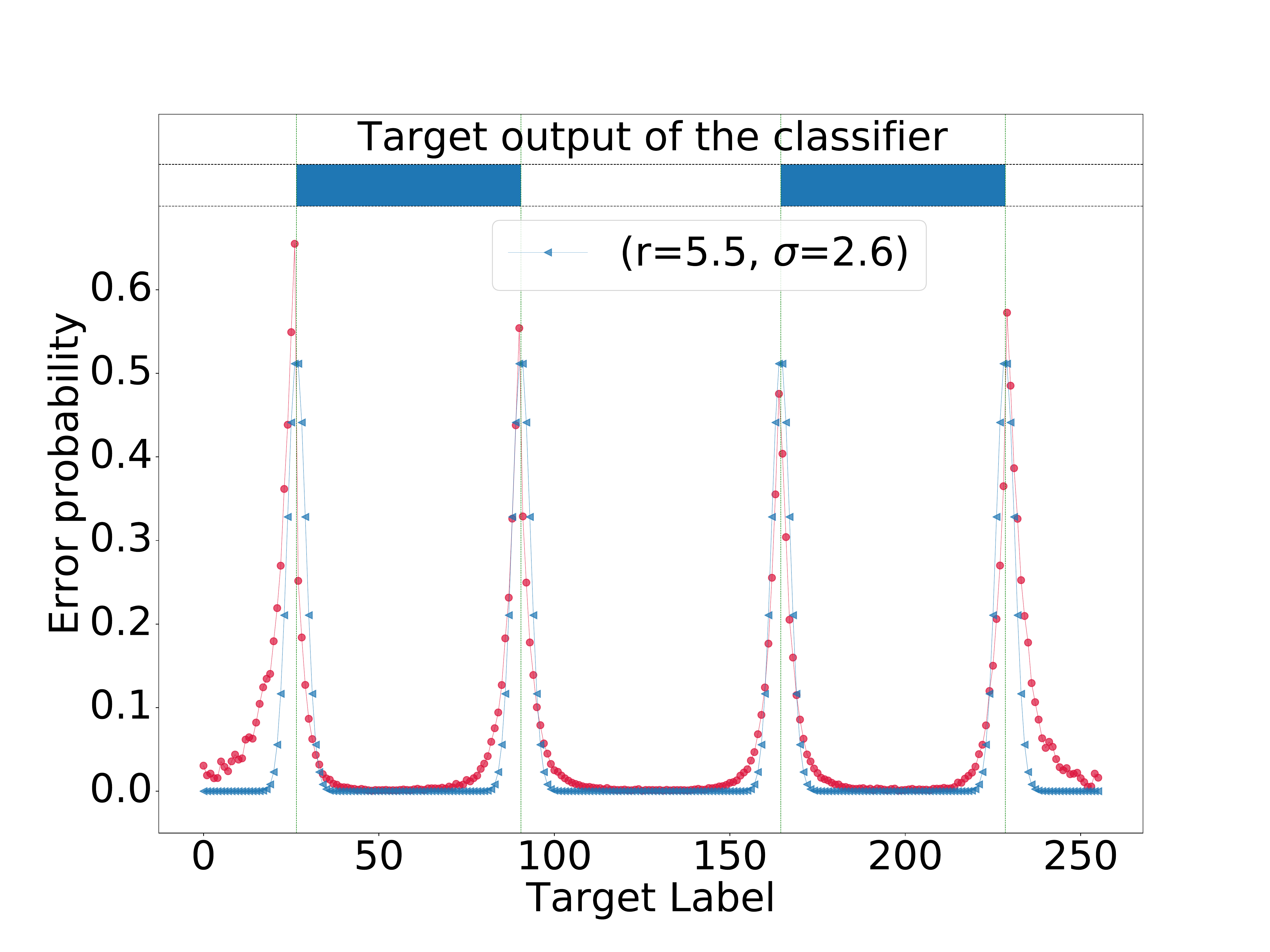}
  \caption{{HRNetV2-W18 (WFLW facial landmark detection)}}
  \label{fig:a8}
  \end{subfigure}
  \begin{subfigure}[t]{0.45\linewidth}
  \centering
  \includegraphics[width=\textwidth]{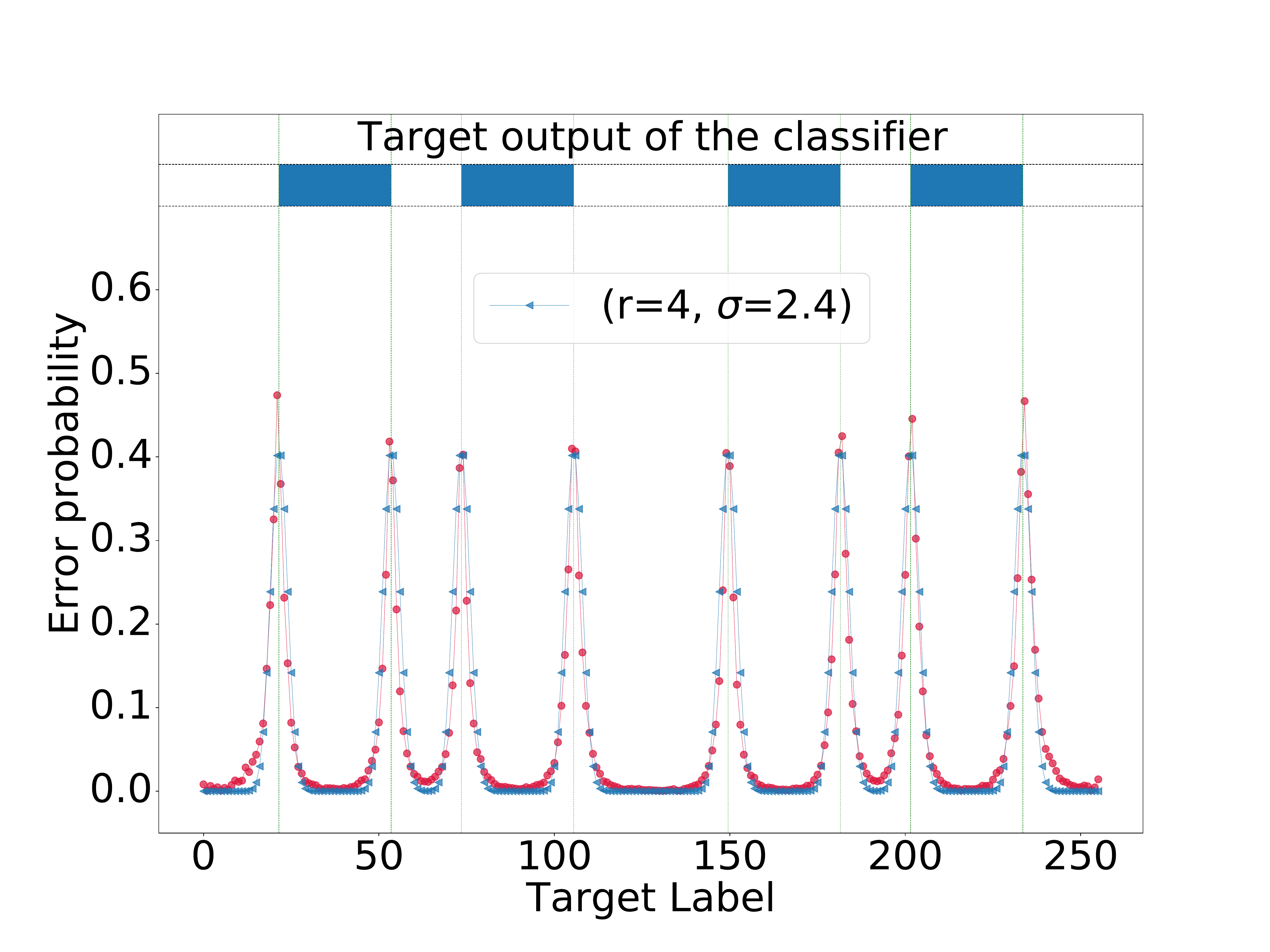}
  \caption{{HRNetV2-W18 (WFLW facial landmark detection)}}
  \label{fig:a9}
  \end{subfigure}
 \caption{Classification error probability versus target label $y$ for different classifiers. The top horizontal bar represent target output of the classifier. Blue color represents output $1$. }
 \label{fig:accuracy}
\end{figure} 
}
\FloatBarrier
%In regression, the target quantized levels $Q_i \in \{1,2,...,N\}$ are on an ordinal scale. 

\section{Experimental Methodology}
\label{sec:supexp}
All experiments are conducted on a Linux machine with an Intel i9-9900X processor and an Nvidia RTX 2080 Ti GPU with 11GB of memory. Our code is implemented using Python 3.8.3 with Pytorch 1.5.1 using CUDA 10.2. Our evaluation is averaged over 5 training runs with separate seeds. 

\subsection{Head Pose Estimation}
\label{sec:hpe}
Head pose estimation aims to find a human head's pose in terms of three angles: yaw, pitch, and roll. 
In this work, we consider landmark-free 2D head pose estimation. 

\paragraph{Datasets:} 
We follow the evaluation setting of Hopenet~\citep{hopenet} and FSA-Net~\citep{fsanet} and use two evaluation protocols with three widely used datasets: 300W-LP~\citep{300wlp}, BIWI~\citep{biwi}, and AFLW2000~\citep{300wlp}. 

\underline{Protocol 1:} BIWI dataset is used for training and evaluation in this protocol. BIWI dataset consists of $24$ videos of $20$ subjects with total $15,128$ frames. Three random splits of 70\%-30\% images are used for training and evaluation. For the BIWI dataset, the yaw angle is in the range $[-75^{\circ},75^{\circ}]$, the pitch is in the range $[-65^{\circ},85^{\circ}]$, and the roll angle is in the range $[-55^{\circ},45^{\circ}]$. 

\underline{Protocol 2:} In this setting, the synthetic 300W-LP dataset is used for training, consisting of $122,450$ samples. The trained network is tested on a real-world AFLW2000 dataset. Yaw, pitch, and roll angles are in the range $[-99^{\circ},99^{\circ}]$ for both datasets. 
%Angles in range [$-180^{\circ}$,$180^{\circ}$] with 360 quantization levels are considered for BEL. 

\paragraph{Evaluation metrics:} 
Mean Absolute Error (MAE) between the target and predicted values is used as the evaluation metric for this benchmark. MAE for a regression task is defined as:
\begin{equation}
    \label{eq:mae}
    \text{MAE} = \frac{1}{N} \sum_{i=1}^{N} \frac{1}{P} \sum_{j=1}^{P} |y_{i,j} - \hat{y}_{i,j}|
  \end{equation} 
  Here, $N$ is the number of test samples, and $P$ is the dimension of the regression task output. For head pose estimation, the dimension of regression output is three (i.e., yaw, pitch, and roll). $y$ is the target, and $\hat{y}$ is the predicted label. 
\paragraph{Network architecture and training parameters: }
We evaluate our approach on two models: ResNet-50 and RAFA-Net. 
With ResNet-50, two runs with different random seeds for each combination of learning rate $\{0.001,0.0001,0.00001\}$ and batch size $\{8,16\}$ are used for hyperparameter tuning. For data augmentation, images are loosely cropped around the center in the training and testing datasets with random flipping. 
With RAFA-Net, we use the training parameters and data augmentation used in~\cite{rafanet}. 
%For data augmentation, we use the approach described in RAFA-Net~\cite{rafanet}.

We refer to Protocol 1 evaluated with ResNet-50 as \textbf{HPE1}, Protocol 1 evaluated with RAFA-Net as \textbf{HPE3}, Protocol 2 evaluated with ResNet-50 as \textbf{HPE2}, and Protocol 2 evaluated with RAFA-Net as \textbf{HPE4}. Table~\ref{tab:trp1} provides a summary of the training parameters used with protocol 1. Table~\ref{tab:trp2} provides a summary of the training parameters used with protocol 2. 

\begin{table}[h]
  \begin{center}
      \setlength\tabcolsep{3pt}
      \caption{Training parameters for head pose estimation with protocol 1. }
      \label{tab:trp1}
      \scriptsize
      \begin{tabular}{L{1cm}L{2.7cm}C{1.7cm}cC{1cm}C{1.5cm}C{2cm}C{1.5cm}}
      \toprule
      Approach &  Label range/Quantization levels & Optimizer & Epochs & Batch size &  Learning rate & Learning rate schedule  & Training time (GPU hours) \\  \midrule
      HPE1 & Yaw: $[-75^{\circ},75^{\circ}]/150$, Pitch:$[-65^{\circ},85^{\circ}]/150$ , Roll: $[-55^{\circ},45^{\circ}]/100$ & Adam, weight decay=0, momentum = 0 & 50 & 8 & 0.0001 & 1/10 after 30 Epochs & 2 \\ \hline
      HPE3 & $[-179^{\circ},180^{\circ}]/360$ & RMSProp, momentum=0, rho = 0.9 & 100 & 16 & 0.001 & - & 6 \\ 
      \bottomrule
      \end{tabular}
  \end{center}
  \end{table}

\begin{table}[h]
\begin{center}
    \setlength\tabcolsep{3pt}
    \caption{Training parameters for head pose estimation with protocol 2. }
    \label{tab:trp2}
    \scriptsize
    \begin{tabular}{L{1cm}L{2.7cm}C{1.7cm}cC{1cm}C{1.5cm}C{2cm}C{1.5cm}}
    \toprule
    Approach & Label range/Quantization levels & Optimizer & Epochs & Batch size &  Learning rate & Learning rate schedule  & Training time (GPU hours) \\  \midrule
    HPE2 & $[-99^{\circ},99^{\circ}]/200$ &Adam, weight decay=0, momentum = 0 & 20 & 16 & 0.00001 & 1/10 after 10 Epochs & 4 \\ \hline
    HPE4 & $[-179^{\circ},180^{\circ}]/360$  &RMSProp, momentum=0, rho = 0.9 & 100 & 16 & 0.001 & - & 48 \\ 
    \bottomrule
    \end{tabular}
\end{center}
\end{table}

\paragraph{Related work}
Existing approaches for head pose estimation include stage-wise soft regression~\citep{ssrnet,fsanet}, a combination of classification and regression~\citep{deepheadpose,hopenet}, and ordinal regression~\citep{quatnet}. 
SSR-Net~\citep{ssrnet} proposes the use of stage-wise soft regression to use the softmax values of classification output to refine the label. 
FSA-Net ~\citep{fsanet} proposes extending stage-wise estimation to head pose estimation using feature aggregation. 
HopeNet~\citep{hopenet} uses a combination of classification and regression loss to train a model for head pose estimation. Whereas, QuatNet~\citep{quatnet} proposes a combination of L2 loss and a custom ordinal regression loss. 
RAFA-Net~\citep{rafanet} uses an attention based approach for feature extraction with direct regression. 

We compare BEL with the performance of related work in Table~\ref{tab:hpe2} and Table~\ref{tab:hpe1}. 95\% confidence intervals are given.

\begin{table}[h]
    \begin{center}
    \setlength\tabcolsep{6pt}
    \caption{Landmark-free 2D Head poses estimation evaluation for protocol 1 (HPE1 and HPE3). } 
    \label{tab:hpe2}
    \scriptsize
    \begin{tabular}{L{3.2cm}C{1.4cm}C{1.1cm}llll@{}}
      \toprule
      Approach & Feature Extractor  & \#Params (M) &  Yaw & Pitch & Roll & MAE  \\ \midrule
    SSR-Net-MD~\citep{ssrnet} { }{ }  (Soft regression) & SSR-Net & 1.1  & 4.24  & 4.35 & 4.19 & 4.26 \\ \hline
    FSA-Caps-Fusion~\citep{fsanet} (Soft regression)  & FSA-Net & 5.1 & \underline{2.89} & 4.29 & 3.60 & 3.60    \\  \hline \hline
    Direct regression (L2 loss) & ResNet50 (HPE1) & 23.5  &  4.62  & 5.24  & 4.43 & 4.76 $\pm$ 0.35  \\ \hline
    BEL-U/GEN-EX/L2  & ResNet50 (HPE1) & 23.6  & \textbf{3.32} & \textbf{{3.80}}  & \textbf{3.53}  & \textbf{3.56} $\pm$ 0.01\\  \hline 
    %BEL-U/GEN-EX/BCE  & ResNet50 (HPE1) & 23.6  & {3.28} & {{3.36}}  & {3.44}  & {3.37} $\pm$ 0.09\\  \hline  \hline
    RAFA-Net~\citep{rafanet} (Direct Regression)  & RAFA-Net (HPE3) & 69.8 & 3.07 & 4.30 & \textbf{\underline{2.82}} & 3.40 
   \\ \hline 
   BEL-B1JDJ/GEN-EX/BCE  & RAFA-Net (HPE3) & 69.8 & \textbf{{3.21}} & \textbf{\underline{3.34}} & {{3.43}}& \textbf{\underline{3.30}} $\pm$ 0.04\\ %%\hline
    %%BEL-J/GEN-EX/BCE*  & RAFA-Net (HPE3) & 69.8 & {{2.83}} & {{3.82}} & {{2.70}}& {{3.12}} $\pm$ 0.12\\
    \bottomrule
    %\cmidrule(r){1-4}
    \end{tabular}
  %%\\  *Best BEL approach based on test error. 
  \end{center}
  \end{table}

\begin{table}[h]
    \scriptsize
    \begin{center}
    \setlength\tabcolsep{6pt}
    \caption{Landmark-free 2D Head poses estimation evaluation for protocol 2 (HPE2 and HPE4). } 
    \label{tab:hpe1}
    \scriptsize
    \begin{tabular}{L{3.2cm}C{1.4cm}C{1.1cm}llll@{}}
      \toprule
      Approach & Feature Extractor  & \#Params (M) &  Yaw & Pitch & Roll & MAE  \\ \midrule
    SSR-Net-MD~\citep{ssrnet} { }{ }  (Soft regression) & SSR-Net & 1.1 & 5.14 & 7.09 & 5.89 & 6.01 \\ \hline
    FSA-Caps-Fusion~\citep{fsanet} (Soft regression)  & FSA-Net & 5.1 & 4.50 & 6.08 & 4.64 & 5.07  \\   \hline
    HopeNet* ($\alpha$ = 2)~\citep{hopenet} (classification + regression loss) &  ResNet50 &  23.9 & 6.47 & 6.56 & 5.44 & 6.16 \\ \hline \hline
    Direct regression (L2 loss) & ResNet50 (HPE2) & 23.5  &  5.85 & 6.34 & 4.80  & 5.65 $\pm$ 0.13   \\ \hline
    BEL-U/GEN-EX/BCE  & ResNet50 (HPE2) & 23.6 & \textbf{4.54} & \textbf{5.76} & \textbf{3.96}  & \textbf{4.77} $\pm$ 0.05  \\  \hline \hline
    RAFA-Net~\citep{rafanet} (Direct Regression)  & RAFA-Net (HPE4) & 69.8 & 3.60 & {{4.92}} & 3.88 & 4.13 \\ \hline 
    BEL-U/GEN-EX/BCE  & RAFA-Net (HPE4) & 69.8 &\textbf{\underline{3.28}} & \textbf{\underline{4.78}} & \textbf{\underline{3.55}} & \textbf{\underline{3.90}} $\pm$ 0.03 \\
   \bottomrule
    %\cmidrule(r){1-4}
    \end{tabular}\\
    %\scriptsize{$*$uses different data preprocessing}
  \end{center}
    \end{table}

\subsection{Facial Landmark Detection}
\label{sec:fld}
Facial landmark detection is a problem of detecting the $(x,y)$ coordinates of keypoints in a given face image. 

\paragraph{Datasets} We use the COFW~\citep{cofw}, 300W~\citep{300w}, WFLW~\citep{lab}, and AFLW~\citep{aflw} datasets with  data augmentation and evaluation protocols described in~\citep{hrnetface}. 
Data augmentation is performed by random flipping, $0.75-1.25$ scaling, and $\pm 30$ degrees in-plane rotation for all the datasets. 
We use $256$ quantization levels for binary-encoded labels. 

\underline{COFW: }
The COFW dataset~\citep{cofw} consists of $1,345$ training and $507$ testing images. Each image is annotated with 29 facial landmarks.  

\underline{300W: } 
This dataset is a combination of HELEN, LFPW, AFW, XM2VTS, and IBUG datasets. Each image is annotated with 68 facial landmarks. The training dataset consists of $3,148$ images. We evaluate the trained model on four test sets: full test set with $689$ images, common subset with $554$ images from HELEN and LFPW, challenging subset with $135$ images from IBUG, and the official test set with $300$ indoor and $300$ outdoor images. 

\underline{WFLW: }
WFLW dataset consists of $7,500$ training images where each image is annotated with $98$ facial landmarks. Full test dataset consists of $2,500$ images. We use test subsets: large pose ($326$ images), expression ($314$ images), illumination ($698$ images), make-up ($206$ images), occlusion ($736$ images), and blur ($773$ images). 

\underline{AFLW: } 
Each image has $19$ annotated facial key points in this dataset. 
AFLW dataset consists of $20,000$ training images where each image is annotated with 19 facial landmarks. The full test dataset consists of $4,836$ images, and the frontal test set consists of $1,314$ images.

\paragraph{Evaluation metrics:} 
Mean Normalized Error (NME) between the target and predicted values is used as the evaluation metric for this benchmark. NME for a regression task is defined as:
\begin{equation}
\label{eq:nme}
\text{NME} = \frac{1}{N} \sum_{i=1}^{N} \frac{1}{P} \cdot \frac{1}{L} \sum_{j=1}^{P} |y_{i,j} - \hat{y}_{i,j}|_2
\end{equation} 
Here, $N$ is the number of test samples, and $P$ is the dimension of the regression task output, i.e., the number of landmarks for facial landmark detection. $y$ is the target, and $\hat{y}$ is the predicted label. $L$ is the normalization factor. 
. Inter-ocular distance normalization is used for COFW, 300W, and WFLW datasets, and bounding box-based normalization is used for AFLW dataset. 

We also report failure rate (f@10\%) for some datasets. 
The failure rate (f@10\%) is defined as the fraction of test samples with normalized errors higher than 0.1. 

\paragraph{Network architecture and training parameters: }
We evaluate BEL by applying it on HRNetV2-W18. HRNetV2-W18 feature extractor's output is $240$ channels of size $64 \times 64$. For heatmap regression, a $1\times 1$ convolution is used to get $P$ heatmaps of size $64 \times 64$, where $P$ is the number of landmarks. 
Since \BEL-x predicts $(x,y)$ coordinates directly we modify the architecture of HRNetV2-W18 to support direct prediction of landmarks. {Figure}~\ref{fig:hrnet} shows the modified architecture of HRNetV2-W18 for \BEL-x.  

The state-of-the-art approaches for facial landmark detection uses heatmap regression, which minimizes the pixel-level loss between the predicted and target heatmaps. We evaluate the applicability of \BEL{ }on heatmap regression in Appendix~\ref{sec:abl}.
In contrast, \BEL-x predicts $(x,y)$ coordinates directly with $256$ quantization levels. 
\begin{figure}[h]
\centering
    \includegraphics[width=0.9\textwidth]{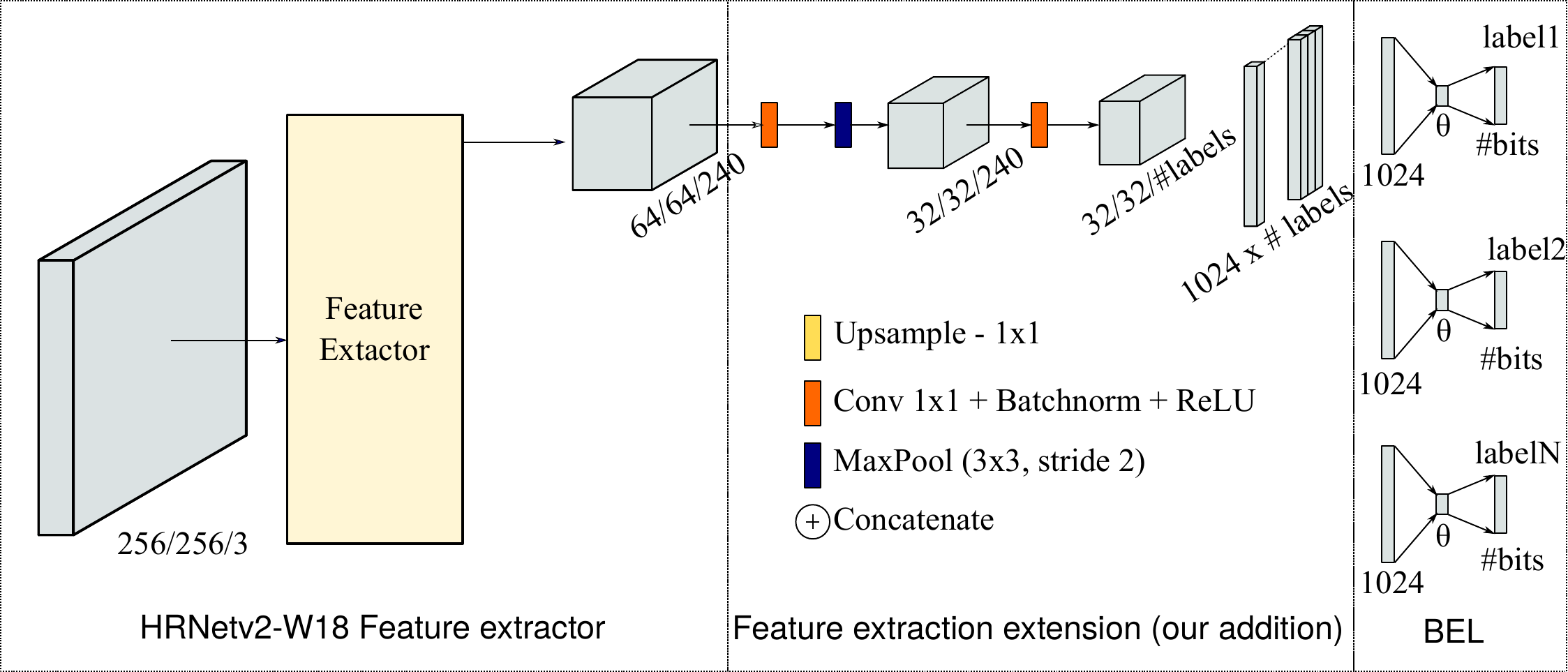}
    \caption{HRNetV2-W18 feature extractor combined with BEL regressor for (x,y) coordinates}
    \label{fig:hrnet}
    \end{figure} 
    \iffalse
\begin{figure}[h]
\centering
    \includegraphics[width=0.9\textwidth]{figures/hrnet_mod6_heatmap.pdf}
    \caption{HRNetV2-W18 feature extractor combined with BEL regressor for heatmaps}
    \label{fig:hrnetc}
\end{figure} 
\fi

We use two runs with different random seeds to decide the learning rate. We consider learning rates $\{0.0003,0.0005,0.0007\}$ and $\theta \in \{10,30\}$.  

Table~\ref{tab:trfdl} provides a summary of all the training parameters. We refer to HRNetV2-W18 evaluated on COFW as \textbf{FLD1}, on 300W as \textbf{FLD2}, on WFLW as \textbf{FLD3}, and on AFLW as \textbf{FLD4}.
\begin{table}[h]
\begin{center}
\setlength\tabcolsep{6pt}
\caption{Training parameters for facial landmark detection for HRNetV2-W18 feature extractor.}
\label{tab:trfdl}
\scriptsize
\begin{tabular}{cC{2.5cm}cC{1cm}C{1.5cm}C{2cm}C{1.5cm}}
    \toprule
Dataset & Optimizer & Epochs & Batch size &  Learning rate (BEL/Direct regression/Multiclass classification) & Learning rate schedule  & Training time (GPU hours) \\  \midrule
COFW & Adam, weight decay=0, momentum = 0 & 60 & 8 & 0.0005/0.0003/ 0.0003 & 1/10 after 30 and 50 Epochs &$\frac{1}{2}$ \\ \hline
300W & Adam, weight decay=0, momentum = 0 & 60 & 8 & 0.0007/0.0003/ 0.0003 & 1/10 after 30 and 50 Epochs & 3 \\ \hline
WFLW & Adam, weight decay=0, momentum = 0 & 60 & 8 & 0.0003/0.0003/ 0.0003 & 1/10 after 30 and 50 Epochs & 5 \\ \hline
AFLW & Adam, weight decay=0, momentum = 0 & 60 & 8 & 0.0005/0.0005/ 0.0003 & 1/10 after 30 and 50 Epochs & 8 \\ 
\bottomrule
\end{tabular}
\end{center}
\end{table}

\paragraph{Related work}
Facial landmark detection is an extensively studied problem used for facial analysis and modeling. 
Common regression approaches for this tasks includes regression using MSE loss~\citep{ssdm,8099876}, cascaded regression~\citep{dsrn,7298989,ccl,fpd}, and coarse-to-fine regression~\citep{fpd,cfss,cfan}. State-of-the-art methods for this task learn heatmaps by regression to find facial landmarks. %~\cite{hrnetface,san,dan,lab}. 
SAN~\citep{san} augments training data using temporal information and GAN-generated faces. 
DVLN~\citep{dvln}, CFSS~\citep{cfss}, LAB~\citep{lab}, DSRN~\citep{dsrn} take advantage of correlations between facial landmarks. % to improve performance. 
DAN~\citep{dan} introduces a progressive refinement approach using predicted landmark heatmaps. % passed across stages of a neural network. 
LAB~\citep{lab} also exploits extra boundary information to improve the accuracy. 
LUVLi~\citep{luvli} proposes a landmark's location, uncertainty, and visibility likelihood-based loss. 
~\cite{binaryheatmap} proposes the use of binary heatmaps with pixel-wise binary cross-entropy loss. 
% SAN~\cite{san} augments training data using temporal information and GAN-generated faces, respectively. DVLN~\cite{dvln}, CFSS~\cite{cfss}, LAB~\cite{lab}, DSRN~\cite{dsrn} take advantage of correlations between facial landmarks to improve performance. DAN~\cite{dan} introduces a progressive refinement approach using predicted landmark heatmaps passed across stages of a neural network. HRNet~\cite{hrnetface} proposes a new CNN architecture to maintain high-resolution representations across the network. 
AWing~\citep{awing} proposes adapted wing loss to improve the accuracy of heatmap regression. 
AnchorFace ~\citep{anchorface} demonstrates that anchoring facial landmarks on templates improves regression performance for large poses. 
HRNet~\citep{hrnetface} proposes a CNN architecture to maintain high-resolution representations across the network, and uses heatmap regression. The target heatmap is generated by assuming a Gaussian distribution around the landmark location. 

We compare BEL with related work in Table~\ref{tab:fldcofw}-~\ref{tab:fldaflw}. 95\% confidence intervals are provided.

\begin{table}[h]
\begin{center}
\setlength\tabcolsep{6pt}
\caption{Facial landmark detection results on COFW dataset (FLD1). The failure rate is measured at the threshold 0.1. $\theta=30$ is used for \BEL. }
\label{tab:fldcofw}
\scriptsize
\begin{tabular}{@{}L{4.1cm}cC{1.5cm}ll}
\toprule
Approach &  Feature Extractor & \#Params/ GFlops & Test NME & FR$_{0.1}$   \\  \midrule
LAB (w B)~\citep{lab}         & Hourglass   & 25.1/19.1 & 3.92 & 0.39   \\ \hline 
AWing~\citep{awing}*         & Hourglass & 25.1/19.1 & 4.94 & -  \\ \hline 
\hline
HRNetV2-W18~\citep{hrnetface} (Heatmap regression) & HRNetV2-W18 & 9.6/4.6 & 3.45 &  0.19 \\   \hline
Direct regression (L2 loss) & HRNetV2-W18 & 10.2/4.7  &  3.96 $\pm$ 0.02 & 0.29  \\ \hline
Direct regression (L1 loss) & HRNetV2-W18 & 10.2/4.7  &  3.60 $\pm$ 0.02 & 0.29 \\ \hline
BEL-HEXJ/GEN-EX/CE          & HRNetV2-W18 & 10.6/4.6  &   \textbf{\underline{3.34}} $\pm$ 0.02  & 0.40 \\ \bottomrule
\end{tabular}
\scriptsize{\\$*$Uses different data augmentation for the training}
\end{center}
\end{table}
\begin{table}[h]
\begin{center}
\setlength\tabcolsep{3.4pt}
\caption{Facial landmark detection results on 300W dataset (FLD2). $\theta=10$ is used for \BEL. }
\label{tab:fld300W}
\scriptsize
\begin{tabular}{@{}L{3.1cm}cC{1.3cm}llll}
\toprule
Approach &  Feature Extractor & \#Params/ GFlops & Test & Common & Challenging & Full  \\  \midrule
DAN~\citep{dan}               & -   & - & -    &  3.19 & 5.24 & 3.59  \\ \hline
LAB (w B)~\citep{lab}         & Hourglass   & 25.1/19.1 & - & 2.98 & 5.19 & 3.49   \\ \hline 
AnchorFace~\citep{anchorface} & ShuffleNet-V2   & - & - & 3.12 & 6.19 & 3.72   \\ \hline 
AWing~\citep{awing}*         & Hourglass & 25.1/19.1 & - & \underline{2.72}& \underline{4.52} & \underline{3.07}  \\ \hline 
LUVLi~\citep{luvli}   & CU-Net & - &  - & 2.76 & 5.16 & 3.23   \\ \hline
\hline
HRNetV2-W18~\citep{hrnetface} (Heatmap regression) & HRNetV2-W18 & 9.6/4.6 &- & \textbf{2.87} & \textbf{5.15} & \textbf{3.32}  \\   \hline
Direct regression (L2 loss) & HRNetV2-W18 & 10.2/4.7  & 4.40 & 3.25 & 5.65 & 3.71 $\pm$ 0.05  \\ \hline
Direct regression (L1 loss) & HRNetV2-W18 & 10.2/4.7  & 4.26 & 3.10 & 5.42 & 3.54 $\pm$ 0.03 \\ \hline
BEL-U/GEN-EX/CE             & HRNetV2-W18 & 11.2/4.6  & \textbf{\underline{4.09}} & {{2.91}} &  5.50  &  {3.40} $\pm$ 0.02   \\ %%\hline
%%BEL-J/GEN-EX/CE**             & HRNetV2-W18 & 11.2/4.6  & {{3.94}} & {{2.87}} &  5.35  &  {3.36} $\pm$ 0.02   \\ 
\bottomrule
%\cmidrule(r){1-4}
\end{tabular}
\scriptsize{\\$*$Uses different data augmentation for the training }
\end{center}
\end{table}

\begin{table}[h]
\begin{center}
\setlength\tabcolsep{2.5pt}
\caption{Facial landmark detection results (NME) on WFLW test (FLD3) and 6 subsets: pose, expression (expr.), illumination (illu.), make-up (mu.), occlusion (occu.) and blur. $\theta=10$ is used for \BEL.  }
\label{tab:fldwflw}
\scriptsize
\begin{tabular}{@{}L{2.6cm}cC{1.3cm}L{1.6cm}cccccc}
\toprule
Approach &  Feature Extractor & \#Params/ GFlops & Test & Pose & Expr. & Illu. & MU & Occu. &Blur  \\  \midrule
LAB (w B)~\citep{lab} & Hourglass   & 25.1/19.1 & 5.27 & 10.24 & 5.51 & 5.23 & 5.15 & 6.79 & 6.32    \\ \hline 
AnchorFace~\citep{anchorface}* & HRNetV2-W18   & -/5.3 & \underline{4.32} & 7.51&4.69 &\underline{4.20} &{4.11} &\underline{4.98} & \underline{4.82} \\ \hline 
AWing~\citep{awing}*         & Hourglass & 25.1/19.1 & 4.36 & \underline{7.38} & \underline{4.58}& 4.32& 4.27& 5.19& 4.96    \\ \hline 
LUVLi~\citep{luvli}   & CU-Net & - &  4.37  &- &- &-  &-  &-  &- \\ \hline
\hline
HRNetV2-W18~\citep{hrnetface} (Heatmap regression) & HRNetV2-W18 & 9.6/4.6 & 4.60 & 7.94 &4.85 &4.55 &4.29 &5.44 & 5.42 \\   \hline
Direct regression (L2 loss) & HRNetV2-W18 & 10.2/4.7  & 5.56 $\pm$ 0.05 & 10.17 & 6.13 & 5.49 & 5.29 & 6.83  &6.52  \\ \hline
Direct regression (L1 loss) & HRNetV2-W18 & 10.2/4.7  & 4.64 $\pm$ 0.03  & 8.13& 4.96& 4.49& 4.45& 5.41&5.25\\ \hline
%0.0501 0.0460 0.0425 0.0738 0.0410 0.0510
%blue    exp     ill    larg   make    occ
%\underline{
BEL-B1JDJ/GEN-EX/CE         & HRNetV2-W18 & 11.7/4.6 &  {\textbf{4.36}} $\pm$ 0.02 &\textbf{{7.53}} &\textbf{4.64} &\textbf{4.28} & \textbf{{4.19}} &\textbf{5.19} &\textbf{5.05}\\  %%\hline
%%BEL-B2JDJ/GEN-EX/CE**         & HRNetV2-W18 & 11.7/4.6 &  {{4.33}} $\pm$ 0.03 &{{7.38}} &{4.60} &{4.25} & {{4.10}} &{5.10} &{5.01}\\ 
\bottomrule
%\cmidrule(r){1-4}
\end{tabular}
\scriptsize{\\$*$Uses different data augmentation for the training}
\end{center}
\end{table}

\begin{table}[h]
\begin{center}
\setlength\tabcolsep{6pt}
\caption{Facial landmark detection results on AFLW dataset (FLD4). $\theta=30$ is used for \BEL. }
\label{tab:fldaflw}
\scriptsize
\begin{tabular}{@{}L{4.1cm}cC{1.5cm}ll}
\toprule
% &     &  &   \multicolumn{1}{c}{COFW} &    \multicolumn{1}{c}{300W} &    \multicolumn{1}{c}{WFLW} &    \multicolumn{1}{c}{AFLW} \\ \cmidrule{4-7}
Approach &  Feature Extractor & \#Params/ GFlops & Full & Frontal  \\  \midrule
LAB (w/o B)~\citep{lab}       & Hourglass   & 25.1/19.1 & 1.85 & 1.62 \\ \hline
AnchorFace~\citep{anchorface} & ShuffleNet-V2   & - &  1.56 &   \\ \hline 
LUVLi~\citep{luvli}   & CU-Net & - & \underline{1.39} & \underline{1.19} \\ \hline \hline
HRNetV2-W18~\citep{hrnetface} (Heatmap regression) & HRNetV2-W18 & 9.6/4.6 & 1.57 & 1.46 \\   \hline 
Direct regression (L2 loss) & HRNetV2-W18 & 10.2/4.7  &  2.10 $\pm$ 0.02 & 1.71   \\ \hline
Direct regression (L1 loss) & HRNetV2-W18 & 10.2/4.7  &  1.51 $\pm$ 0.01 & 1.34  \\ \hline
BEL-B1JDJ/GEN-EX/CE         & HRNetV2-W18 & 10.8/4.6  & \textbf{1.47} $\pm$ 0.00 & \textbf{1.30}      \\ \bottomrule
%\cmidrule(r){1-4}
\end{tabular}
\scriptsize{\\$*$Uses different data augmentation for the training}
\end{center}
\end{table}
\FloatBarrier

\subsection{Age Estimation}
\label{sec:ae}
Age estimation aims to predict the age given an image of a human head. 

\paragraph{Datasets} 
We use the MORPH-II~\citep{morphii} and AFAD~\citep{agecnn} datasets for our evaluation. Cumulative Score (CS) and MAE are used as evaluation metrics. We preprocess the MORPH-II dataset by aligning images first along the average eye position~\citep{mlxtend}, then by re-aligning so that the tip of the nose is in the center of each image. We do not preprocess the AFAD dataset as faces are already centered. Afterwards, face images are resized to $256 \times 256 \times 3$ and randomly cropped to $224 \times 224 \times 3$ for training. For testing, a center crop of $224 \times 224 \times 3$ is taken. 

\underline{MORPH-II: }
This dataset consists of 55,608 face images with age labels between 16 and 70. The dataset is randomly divided into 39,617 training, 4,398 validation, and 11,001 testing images. 

\underline{AFAD: }
This dataset consists of 164,432 Asian facial images and age labels between 15 and 40. The dataset is randomly divided into 118,492 training, 13,166 validation, and 32,763 testing images. 

\paragraph{Evaluation metrics:} 
MAE (\Eqref{eq:mae}) is used as the evaluation metric. 
We report Cumulative Score (CS$\theta$) for some datasets. 
CS$\theta$ is defined as the fraction of test images with absolute error less than $\theta$ years. 

\paragraph{Network architecture and training parameters: }
We evaluate our approach on ResNet-50. 
We perform two runs with different random seeds to determine the learning rate between $[0.00001, 0.0001, 0.001]$ and use a batch size of 64 for all experiments. We use ImageNet pretrained weights to initialize the network. Full training parameters are described in Table~\ref{tab:trage}. We refer to our evaluation on MORPH-II as \textbf{AE1} and AFAD as \textbf{AE2}.

\begin{table}[h]
\begin{center}
    \setlength\tabcolsep{6pt}
    \caption{Training parameters for age estimation using MORPH-II and AFAD dataset }
    \label{tab:trage}
    \scriptsize
    \begin{tabular}{L{3cm}C{1cm}ccC{2cm}C{2cm}}
    \toprule
    Optimizer & Epochs & Batch size &  Learning rate & Learning rate schedule  \\  \midrule
    Adam, weight decay=0, momentum=0 & 50 & 64 & 0.0001 & - 
    \\ \bottomrule
    \end{tabular}
\end{center}
\end{table}

\paragraph{Related work}
Existing approaches for age estimation include ordinal regression~\citep{agecnn, coralcnn}, soft regression~\citep{ssrnet}, and expected value ordinal regression ~\citep{mvloss, dldl}. OR-CNN~\citep{agecnn} proposed the use of ordinal regression via binary classification to predict the label. CORAL-CNN~\citep{coralcnn} refined this approach by enforcing the ordinality of the model output. SSR-Net~\citep{ssrnet} proposed the use of stage-wise soft regression using the softmax of the classification output to refine the predicted label. MV-Loss~\citep{mvloss} extended the soft regression approach by penalizing the output of the model based on the variance of the age distribution, while DLDL~\citep{dldl} proposed to use the KL-divergence between the softmax output and a generated label distribution to train a model. 

We compare BEL with related work in Table~\ref{tab:age} and Table~\ref{tab:agee}. 95\% confidence intervals are provided.

\begin{table}[h]
    {
    \centering
    \setlength\tabcolsep{4pt}
    \caption{Age estimation results on MORPH-II dataset (AE1). $\theta=10$ is used for BEL.}
    \label{tab:age}
    \scriptsize
    \begin{tabular}{L{4.4cm}cR{2cm}L{2cm}L{2cm}}
    \toprule
    Approach & Feature extractor & \#Parameters (M) & MORPH-II (MAE)& MORPH-II (CS$\theta = 5$)\\ \midrule
    OR-CNN~\citep{agecnn} (Ordinal regression by binary classification ) & - & 1.0 & 2.58 & 0.71 \\ \hline
    MV Loss~\citep{mvloss} (Direct regression) & VGG-16 & 138.4 & 2.41 & 0.889 \\ \hline
    DLDL-v2~\citep{dldl} (Ordinal regression with multi-class classification) & ThinAgeNet & 3.7 & \underline{1.96}* & - \\ \hline
    CORAL-CNN~\citep{coralcnn} (Ordinal regression by binary classification) & ResNet34 & 21.3 & 2.49 & - \\ \hline \hline
    Direct Regression (L2 Loss) & ResNet50 & 23.1 & 2.44 $\pm$ 0.01 & 0.903 $\pm$ 0.002 \\ \hline
    BEL-J/BEL-J/BCE & ResNet50 & 23.1 & \textbf{{2.27}} $\pm$ 0.01& \textbf{\underline{0.928}} $\pm$ 0.001\\   \bottomrule 
    \end{tabular}
    }
    \scriptsize{\\$*$Uses different data augmentation for the training}
    %\vspace{-3mm}
\end{table}
\begin{table}[h]
    {
    \centering
    \setlength\tabcolsep{4pt}
    \caption{Age estimation results on AFAD dataset (AE2). $\theta=10$ is used for BEL.}
    \label{tab:agee}
    \scriptsize
    \begin{tabular}{L{4.4cm}cR{2cm}L{2cm}L{2cm}}
    \toprule
    Approach & Feature extractor & \#Parameters (M) & AFAD (MAE)& AFAD (CS$\theta = 5$)\\ \midrule
    OR-CNN~\citep{agecnn} (Ordinal regression by binary classification ) & - & 1.0 & 3.51 & 0.74 \\ \hline
    CORAL-CNN~\citep{coralcnn} (Ordinal regression by binary classification) & ResNet34 & 21.3 & 3.47 & - \\ \hline \hline
    Direct Regression (L2 Loss) & ResNet50 & 23.1 & 3.21 $\pm$ 0.02 & 0.810 $\pm$ 0.02 \\ \hline
   BEL-B1JDJ/GEN-EX/L1 & ResNet50 & 23.1 & \textbf{\underline{3.11}} $\pm$ 0.01& \textbf{\underline{0.823}} $\pm$ 0.001\\ \bottomrule 
    \end{tabular}
    }
    %\vspace{-3mm}
  \end{table}

\subsection{End-to-end Self Driving}
\label{sec:pn}

We evaluate our approach on the NVIDIA PilotNet dataset and PilotNet model for end-to-end autonomous driving~\citep{pilotold}.
In this task, the steering wheel's next angle is predicted from an image of the road. We refer to these experiments as \textbf{PN}. 
MAE (\Eqref{eq:mae}) is used as the evaluation metric. 

\paragraph{Dataset}
We use a driving dataset consisting of 45,500 images taken around Rancho Palos Verdes and San Pedro, California~\citep{sully}. We crop images to $256\times 70\times 3$ then resize them to $200\times 66\times 3$. We randomly vary the brightness of the image between $[0.2\times, 1.5\times]$, randomly flip images, and make random minor perturbations on the steering direction. 
We use $\theta=10$ with $670$ quantization levels for \BEL.
\paragraph{Training parameters}
We perform two runs with different random seeds to determine the learning rate between $[0.00001, 0.0001, 0.001]$ and use a batch size of 64 for all experiments. Full training parameters are described in Table~\ref{tab:tauto}.
\begin{table}[h]
\begin{center}
    \setlength\tabcolsep{6pt}
    \caption{Training parameters for end-to-end autonomous driving using PilotNet. }
    \label{tab:tauto}
    \scriptsize
    \begin{tabular}{L{3cm}C{2cm}ccC{2cm}C{2cm}}
    \toprule
    Optimizer & Epochs & Batch size &  Learning rate & Learning rate schedule  \\  \midrule
    SGD with weight decay=1e-5, momentum=0 & 50 & 64 & 0.1 & 1/10 at 10, 30 epochs
    \\ \bottomrule
    \end{tabular}
\end{center}
\end{table}

\paragraph{Related work}
End-to-end autonomous driving is a novel task that has become increasingly relevant due to the rise of self-driving vehicles. The autonomous driving model's task is to predict the future driving angle based on a forward-facing image from the perspective of the vehicle. PilotNet~\citep{pilotnet} used a small, application-specific network to provide good accuracy within the time constraints of autonomous driving.

We compare BEL with the baseline PilotNet architecture in Table~\ref{tab:pilot}. 95\% confidence intervals are provided.

\begin{table}[h]
    {
    \begin{center}
    \setlength\tabcolsep{4pt}
    \caption{End-to-end autonomous driving results on PilotNet dataset (PN) and architecture~\citep{pilotnet,pilotold}.}
    \label{tab:pilot}
    \scriptsize
    \begin{tabular}{L{5.3cm}ccc}
    \toprule
    Approach & Feature extractor & \#Parameters (M) & MAE \\ \midrule
    PilotNet~\citep{pilotnet} & PilotNet & 1.8 & 4.24 $\pm$ 0.45 \\ \hline
    BEL-J/GEN/CE & PilotNet & 1.8 & \textbf{\underline{3.11}} $\pm$ 0.01 \\ 
    \bottomrule 
    \end{tabular}
    \end{center}
    }
\end{table}

\end{document}